\documentclass[a4paper]{article}
\usepackage[sc]{mathpazo}
\usepackage[T1]{fontenc} 
\usepackage{pdfpages}
\usepackage{microtype}
\usepackage{multicol}
\usepackage{graphicx}
\usepackage{listings}
\usepackage{amssymb,amsfonts,amsmath}
\usepackage[hmarginratio=1:1,top=32mm,columnsep=20pt]{geometry} 
\usepackage{lettrine} 
\usepackage{textcomp}
\usepackage{abstract}
\usepackage{subcaption}
\usepackage{color,listings}
\definecolor{lightgray}{rgb}{.7,.7,.7}
\definecolor{white}{rgb}{1,1,1}
\newsavebox\lstbox

\usepackage{fancyhdr}
\title{\vspace{-12mm}\fontsize{24pt}{25pt}\selectfont\textbf{Sooner than Expected: Hitting the Wall of Complexity in Evolution}}
\author{
\large
\textsc{Thomas Schmickl\,$^{1}$, Payam Zahadat\,$^1$, Heiko Hamann\,$^{2,}$\footnote{Corresponding author: {heiko.hamann@uni-paderborn.de}}} \\[2mm] \\ 
\small $^{1}$Artificial Life Lab of the Department of Zoology, \\ \small Karl-Franzens University Graz, Austria\\
\small $^{2}$Heinz Nixdorf Institute, Department of Computer Science, \\ \small  University of Paderborn, Germany and Department of Computer Science, \\ \small Chemnitz University of Technology, Germany}
\date{\small submitted $25^{th}$ September 2016}

\begin{document}

\maketitle 

\thispagestyle{fancy} 

\begin{abstract}

In evolutionary robotics an encoding of the control software, which
maps sensor data (input) to motor control values (output), is shaped
by stochastic optimization methods to complete a predefined task. This
approach is assumed to be beneficial compared to standard methods of
controller design in those cases where no a-priori model is available
that could help to optimize performance. Also for robots that have to operate
in unpredictable environments, an evolutionary robotics approach is favorable. 
We demonstrate here that such a model-free approach is not a free lunch, as already simple tasks can represent unsolvable barriers for fully open-ended
uninformed evolutionary computation techniques. We propose here the 
``Wankelmut''  task as an objective for an evolutionary approach that starts from 
scratch without pre-shaped controller software or any other informed approach
that would force the behavior to be evolved in a desired way. Our
focal claim is that ``Wankelmut'' represents the simplest set of problems that makes
plain-vanilla evolutionary computation fail. We demonstrate this by
a series of simple standard evolutionary approaches using different fitness
functions and standard artificial neural networks as well as
continuous-time recurrent neural networks. All our tested approaches
failed. We claim that any other evolutionary approach will also fail that
does per-se not favor or enforce modularity and does not freeze or
protect already evolved functionalities. Thus we propose a
hard-to-pass benchmark and make a strong statement for self-complexifying
and generative approaches in evolutionary computation. We anticipate
that defining such a ``simplest task to fail'' is a valuable benchmark
for promoting future development in the field of artificial
intelligence, evolutionary robotics and artificial life.
\end{abstract}
\small{\section*{} \textbf{Keywords: Evolutionary Computation, Complexity, Artificial Neural Networks, CTRNN, Agent-Based Model, Stochastic Optimization}}
\newpage

\section{Introduction}

Robots are used more and more frequently in inherently
unpredictable outdoor environments: Aerial search, 
rescue drones, deep-diving underwater AUVs and even
extra-terrestrial explorative probes. It is impossible to program
autonomous robots for those tasks in a way that one a-priori 
accommodates for all possible events that might occur during such
missions. Thus, on-line and on-board learning, conducted for example by
evolutionary computation and machine learning, becomes a significant
aspect in those
systems \cite{sutton98,bongard13,prokopenko14,eiben15}. Evolutionary
Robotics has become a promising field of research to push forward the
robustness, flexibility, and adaptivity of autonomous robots which
combines the software technologies of machine learning, evolutionary
computation, and sensorimotor control with the physical embodiment of the
robot in its environment.

We claim here that evolutionary robotics operating without a-priori
knowledge can fail easily because it suddenly hits an obscured ``wall
of complexity'' with the current state of the art of unsupervised
learning. While extremely simple (trivial) tasks evolve well with
almost every approach that was tested in literature, already slightly
more difficult tasks make all methods fail that are not a-priori
tailored to the properties of the given task. As
  is well-known, there is no free lunch in
  optimization techniques~\cite{wolpert1997no}. The more specificly an optimizer
  is tailored to a specific problem, the more probable it is that it will fail for other types of problems. Thus, only open-ended, uninformed
  evolutionary computation will allow for generality
  in problem solving as required in long-term autonomous
  operations in unpredictable environments. For example, most studies
in literature that have evolved complex tasks of cooperation and
coordination in robots used pre-structured software
controllers~\cite{marocco_2006_origins,nolfi00} while most studies
that have evolved robotic controllers in an uninformed open-ended way
produced only simple behaviors, such as coupled oscillators that
generate gaits in
robots~\cite{sims94,clune09,Zahadat2010DARS,hamann10b} including
simple reactive gaits~\cite{zahadat2012a}, homing, collision
avoidance, area coverage, collective pushing/pulling, and similar
straight-forward tasks~\cite{FloreanoKeller_2010}.

We hypothesize that one reason for the lack of success of evolving solutions for complex tasks, is the improbable emergence of internal
modularity~\cite{Lipson04principlesof} in software controllers using
open-ended evolution without explicitly enabling the evolution of modules.
In an evolutionary approach it is possible to
push towards modularity by pre-defining a certain topology of
Artificial Neural Networks (ANNs)~\cite{Nolfi96usingemergent}, by allowing to evolve a potentially unlimited number of modules within a pre-defined modular ANN structure~\cite{urzelai98,duro01}, or by
switching between tasks on the time-scale of
generations~\cite{Kashtan2005}, which can then also be improved by
imposing costs for links between
neurons~\cite{Clune2012evolutionaryorigins}. However, if modularity
is not pre-defined and not explicitly encouraged by a designer, then a
modular software controller will not emerge from scratch even if
modularity is directly required by the task. While nature was capable
of evolving highly layered, modularized and complex brain
structures~\cite{Shallice1988} by starting from scratch, evolutionary
computation fails to achieve similar progress within reasonable time.

To support our claim we have searched for the most simple task that 
leads to failure of plain-vanilla uninformed unsupervised evolutionary
computation starting from a 100\% randomized control software without
any interference (guidance) by a designer during evolution and without
any a-priori mechanism or impetus to favor self-modularization. An
easy way to define such a task is to construct it from two simple but
conflicting tasks which are both easily evolvable in isolation for
almost any evolutionary computation and machine learning approach of
today. However, we require that the behavior that solves task~1 is
the inverse of the behavior that solves task~2 (e.g., positive and
negative phototaxis). Hence, once one behavior has been evolved the
other behavior needs to be added together with an action selection
mechanism. 
We have proposed a task that operates in one-dimensional space and as it is shown in the paper, it can be solved by a very simple pseudo-code and also by hand-coded ANNs. Compared to the classification scheme of \cite{braitenberg_vehicles_1984} an agent, that solves the Wankelmut task, would be placed between vehicle~\#4 and vehicle~\#5 concerning the complexity of its behavior. Given that it operates in a one-dimensional space introduces a relation to vehicle~\#1.
We anticipate that searching to define a ``most simple
task to fail'' is a valuable effort  for promoting future
development in the fields of artificial intelligence, evolutionary
robotics, and artificial life.

For simplicity we suggests simple plain-vanilla ANNs as an evolvable
runtime control software in combination with genetic
algorithms~\cite{holland75} or evolution
strategies~\cite{rechenberg73} as an unsupervised adaptation
mechanism. For our benchmark defined here we accept either large fully connected and randomized ANNs
as initialization as well as ANNs implementations that allow
restructuring (adding and removing of nodes and connections) as a
starting point. However, we consider all implementations that have
special implementations to facilitate or favor modular networks as
``inapplicable for our focal research question'' because the main challenge in our benchmark task is to
evolve modularization from scratch as an emergent solution to the task.

\section{Our Benchmark Task}

\begin{figure}[t]
\begin{center}
\includegraphics[width=10cm,angle=90]{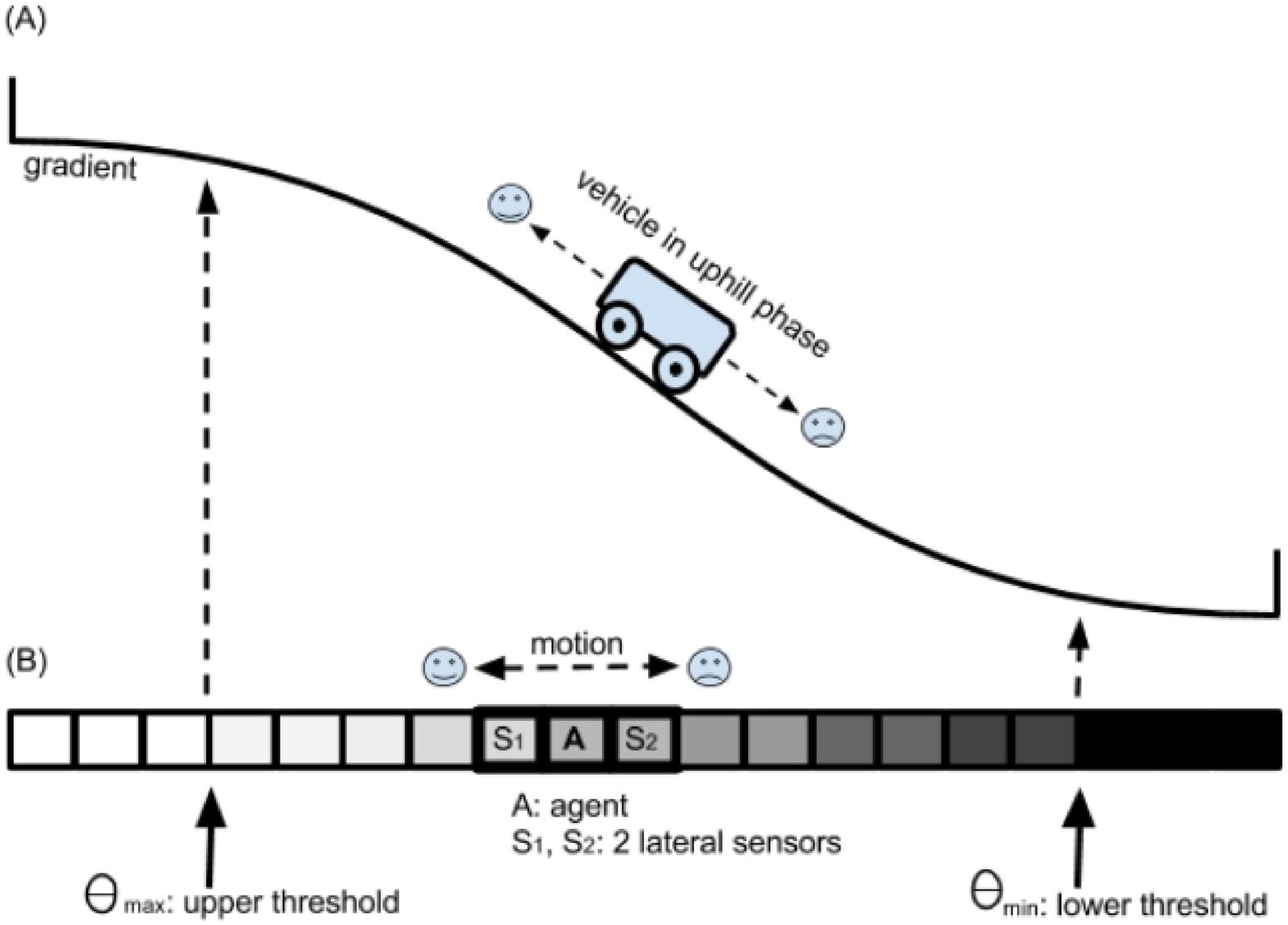}
\end{center}
\caption{ (A) Schematic drawing of the problem posed in our focal
  benchmark task. The evolved software controller should make a
  vehicle cycle between the high and the low region. (B)~The focal
  task is broken down to a 1D cellular simulation with discrete space
  and time to achieve a maximum of simplification without losing the
  difficulty of the benchmark.}\label{fig:task}
\end{figure}

For the sake of simplicity we restrict ourselves to a very simple task that is
hard to evolve in an open-ended, uninformed, and unguided way. As
shown in Fig.~\ref{fig:task}, we assume an agent that moves in an
environment expressing one single quality factor (e.g., height, water or air pressure,
luminance, temperature) and which has to evolve a behavior that first
makes the agent move uphill in this environmental gradient.

In the ``uphill-walk'' phase of the experiment the agent always should
move in the direction of the sensor that reports the higher value of
the focal quality factor. As soon as the agent reaches an area of
sufficiently high quality (above a threshold $\Theta_{max}$) the agent
should switch its behavioral mode and start to seek areas of low
environmental quality. In this ``downhill-walk'' phase the agent should
always move towards the side where its sensor reports the lowest
environmental quality value. After the agent has reached a
sufficiently low quality area (below a threshold $\Theta_{min}$) the
agent should switch back to the ``uphill-walk'' behavior again.

We call the behavior that we aim to evolve ``\textbf{Wankelmut}'',
a term that expresses in German a character trait that always switches
between two different goals as soon as one of the goals was reached.
A "Wankelmut" agent is never satisfied and thus does not decide for one thing
and does not stick with it. It is a variant of ``the grass looks
always greener on the other side'', a commonly found personality
feature found in natural agents (from humans to animals) and, despite
its negative perception in many cultural moral systems, has its
benefits. It keeps the agent going, being explorative, curious and,
never satisfied. That is exactly the desired behavior of an autonomous
probe on a distant planet that needs to be explored.

In summary, the task is to follow an environmental gradient up and
down in an alternating way, hence maximizing the coverage (monitoring,
observation, patrolling) the areas between $\Theta_{min}$ and
$\Theta_{max}$. There are many examples that were evolved by natural
selection~\cite{darwin1859}. In social insects (ants, termites,
wasps, honeybees), foragers have first to go out of the nest and after
they encountered food they switch their behavior and go back to the
nest. After they have unloaded the food to other nest workers they go
outwards to forage again~\cite{Seeley_wisdom}. Often environmental
cues and gradients are involved in the homing and in the foraging
behaviors (sun compass, nest scent, pheromone marks) and are
exploited differently by the workers in the outbound behavioral state
compared to the inbound behavioral state~\cite{camazine01}. Other
biological sources of inspiration are animals following a diurnal
rhythm (day-walkers and night-walkers). In an engineering context the
rhythm might be imposed by energy recharging cycles, water depths, or
aerial heights in transportation tasks by underwater vehicles or
aerial drones.

Figure~\ref{fig:task}B shows that we de-complexified the benchmark
task to a one-dimensional cellular space of $N$~cells in which always
one cell of index~$P(t)$ is occupied by the agent which has state~$S(t)$. The
environmental gradient is produced by the Gauss error function
\begin{equation}
  \operatorname{erf}(x) = \frac{2}{\sqrt\pi}\int_0^x e^{-t^2}\,\mathrm dt,
\end{equation}
whereby the quality of every cell $i$ is modeled as
\begin{equation}
\text{quality}[i] = \operatorname{erf}\left(\frac{4  (i - \frac{N}{2})}{N}\right).
\end{equation}

In every time step $t \in [0, t_{max} ]$, the agent has access to two
lateral sensor readings which are modeled as
\begin{equation}
\label{eq:sensorleft}
s_l(t) = 
\begin{cases}
\text{quality}[P(t)-1] & \text{ if } (P(t)>0)\\ 
\text{quality}[P(t)] & \text{otherwise} \\
\end{cases}
\end{equation}
and
\begin{equation}
\label{eq:sensorright}
s_r(t) = 
\begin{cases}
\text{quality}[P(t)+1] & \text{ if } (P(t)<N-1)\\ 
\text{quality}[P(t)] & \text{otherwise} \\
\end{cases}.
\end{equation}
The agent changes its position based on these sensor readings:
\begin{equation}
P(t+1) = \min(N-1, \max(0, P(t) + f(s_l(t), s_r(t)))),
\label{eq:P}
\end{equation}
whereby the agent's position is restricted by the boundaries of the
simulated world. Its motion ($f$) is restricted to a maximum of 1~step to the left or to the right of the agent's current position.

Initially the agent is in uphill state which means it should move
uphill. If the agent's local $\text{quality}[P(t)] \geq \Theta_{max}$
then the state is changed to downhill and if the agent's local
$\text{quality}[P(t)] \leq \Theta_{min}$ then the agent's state is
changed back to uphill.



\section{Known solutions}

The Wankelmut task is actually very simple to solve, as it is just a
greedy uphill walk in combination with a greedy downhill walk
complemented by a threshold-dependent switch. 

A~simple algorithm, such
as the python-like pseudo code in Fig.~\ref{fig:pseudoCode} solves the task with a few lines of code in the desired reactive and optimal way. We used this code to calculate the ``maximum reachable fitness'' in our quantitative analysis section.
The variable \texttt{state} and the constant \texttt{theta} define the
agent's own state and the thresholds at which it should switch from
one behavior to the other and vice versa. The two variables
\texttt{S\_l} and \texttt{S\_r} hold the current sensor values at the
left and at the right side of the agent and report values between
$-1.0$ and $+1.0$. The function \texttt{set-actuators()} drives the
robot to the left with positive numbers and to the right with negative
numbers used as the only argument. In our study we use this simple
hand-coded solution as a reference and investigate how close the
evolved control software gets to the fitness values achieved by this
simple but optimal solution.

\begin{figure}
\begin{verbatim}
# the agent's initialization is executed at the beginning

def agent_init():
  state=1     # state: +1 is uphill, -1 is downhill state
  theta=0.95  # thresholds are at .05 and .95 (symmetrical setting)

# the agent's action procedure is executed every time step

def agent_act():
  if (S_l*state)>(S_r*state):
    set-actuators(state)
  else:
    set-actuators(state*-1)
  if (mean(S_l,S_r)*state)>theta:
    state=state*-1
\end{verbatim}
\caption{\label{fig:pseudoCode}Python-like pseudo-code of a hand-coded controller
  which produces the desired reactive Wankelmut behavior in an optimal way.}
\end{figure}

It is noteworthy that this reactive solution of the
Wankelmut task would operate in gradients of any shape and size in an
optimal way as long as gradients are strictly monotone between the two
points $\Theta_{min}$ and $\Theta_{max}$. While we used a sigmoid-type
non-linear gradient by using the Gauss error function, our hand-coded
solution, presented as pseudo-code in Fig.~\ref{fig:pseudoCode}, 
works optimally also in linear gradients of any size and steepness.



\begin{figure}[h!]
\vspace*{2cm}
\begin{minipage}[b]{.49\linewidth}
\includegraphics[width=1.0\textwidth]{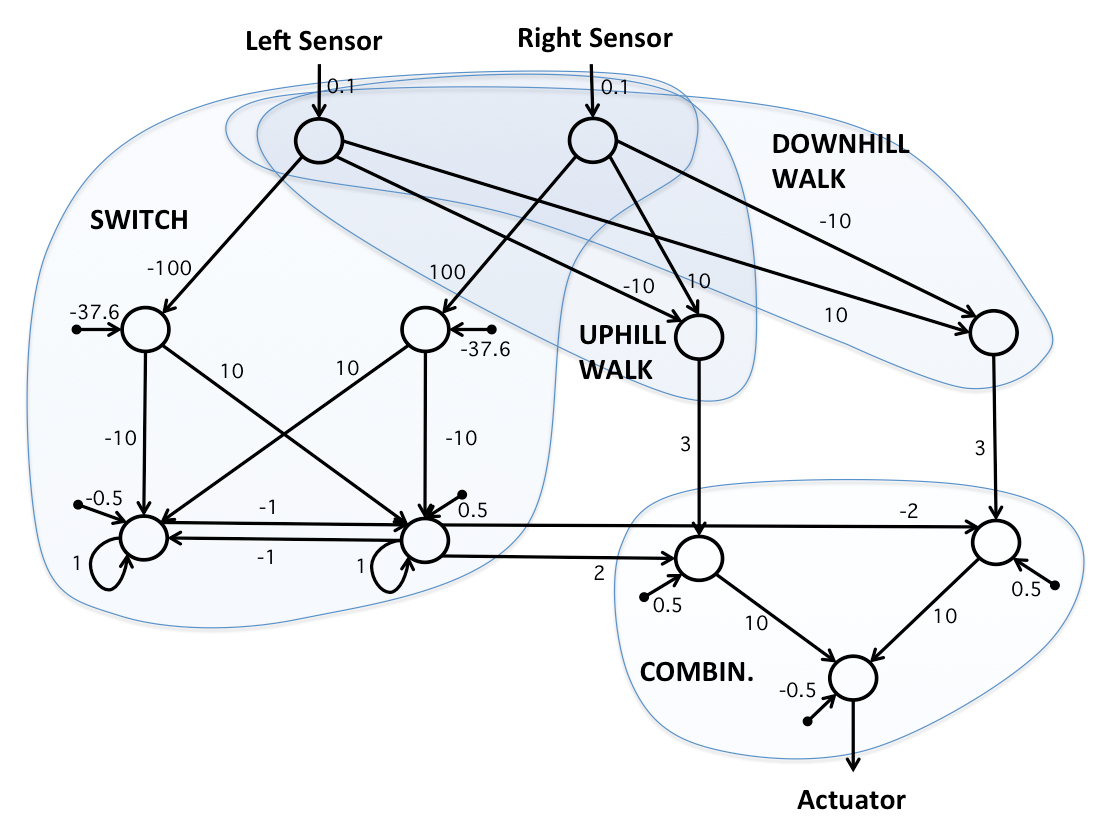}
\subcaption{Hand-coded ANN (first example)}\label{fig:handcoded:noxor}
\end{minipage}%
\hspace*{.05\linewidth}%
\begin{minipage}[b]{.49\linewidth}
\includegraphics[width=0.87\textwidth]{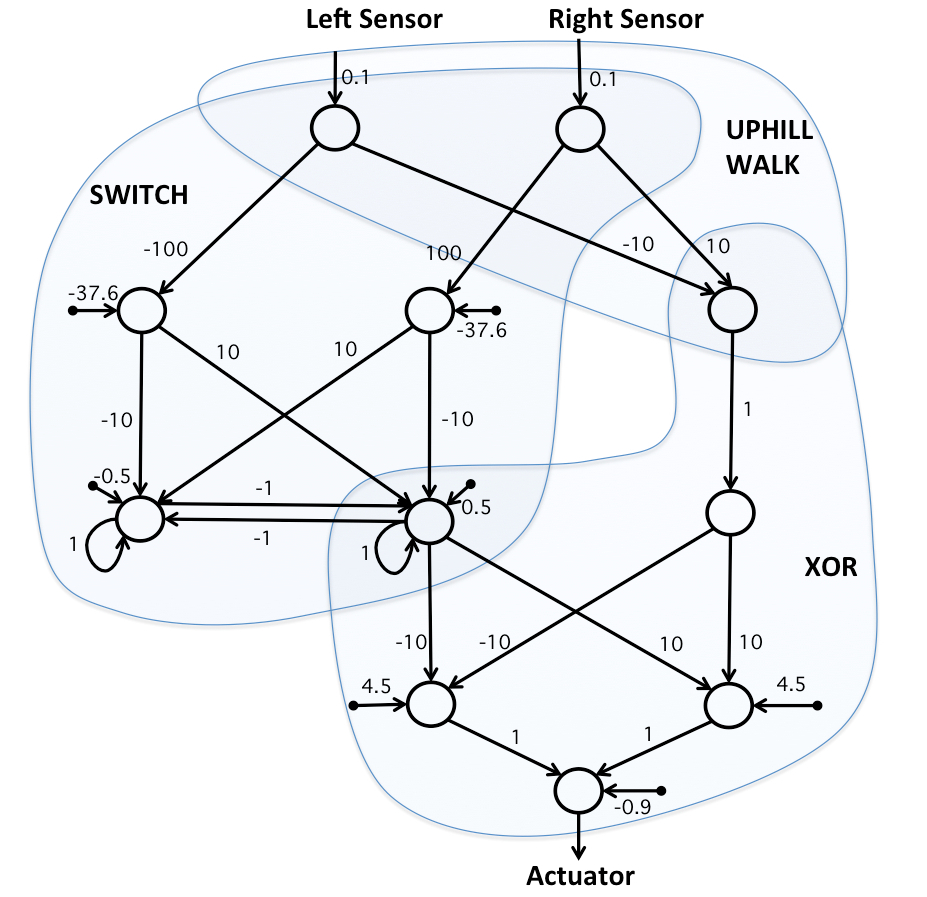}
\subcaption{Hand-coded ANN (second example)}\label{fig:handcoded:withxor}
\end{minipage}
\caption{Schematic drawing of two example hand-coded ANN solutions.}\label{fig:handcoded}
\end{figure}
\setcounter{subfigure}{0}

\section{Evolvable Agent Controllers}

\paragraph*{Hand-Coded Artificial Neural Network} 
In order to make sure that the topology of our neural networks that are evolved is sufficient for solving the problem, we  designed hand-coded neural networks. For the activation function, we used a sigmoid function: $1/(1+\exp(-20x))-0.5$.
Fig.~\ref{fig:handcoded} shows two example hand-coded neural networks. 
In Fig.~\ref{fig:handcoded:noxor}, an uphill and a downhill walk subnetworks and a switch subnetwork are designed. The switch keeps the current state of the controller and when the inputs pass the thresholds, it switches to the other state. Finally in the last subnetwork, the information from the switch is used to choose between uphill and downhill walk.
Fig.~\ref{fig:handcoded:withxor}, has a slightly different design where the output of the switch and an uphill walk subnetworks are combined by using a logical xor subnetwork. The number of the nodes in both networks are the same, however, the number of weights used in the second example is lower.
The topology of the neural network to be evolved is chosen in a way that it covers the second hand-coded network.
The number of the nodes chosen for the CTRNN network to be evolved is chosen so that it covers both networks.

The behaviors of an agent controlled by both networks are the same. The behaviors in two different environments are demonstrated in Fig.~\ref{fig:evolvehandcoded:hct_st1}~,~\ref{fig:evolvehandcoded:hct_st2}. Fig.~\ref{fig:evolvehandcoded:hct_st1}, shows the behavior when the quality of the environment is increasing from left to right. Fig.~\ref{fig:evolvehandcoded:hct_st2}, shows the behavior when the quality of the environment is decreasing (The quality of the environments are represented in gray-scales). 


\subsection{Simple Artificial Neural Network}
\label{sec:ann}
Our simple approach made use of recurrent artificial neural
networks. 
The used activation function is a sigmoid function:
$1/(1+\exp(-20x))-0.5$. We had 11~neurons with two in the input layer,
three in a first hidden layer, another three in a second hidden layer, two in the third hidden layer,
and one in the output layer. Each neuron in the second hidden
layer had a link to itself (loop) and an input link from each of the neighboring nodes in addition to the links from the nodes in the previous layer.
Weights were randomly initialized with random uniform distribution
from the interval~$[-0.5,0.5]$.


\subsection{Continuous-Time Recurrent Neural Networks}
In a second approach we used Continuous-Time Recurrent Neural Networks
(CTRNNs) which are Hopfield continuous networks with unrestricted weights
matrix inspired by biological neurons
~\cite{FunahashiN93}. A~neuron~$i$ in the network was of the following
general form:
\begin{equation}
\label{eq:ctrnn}
\tau_i\dot{y_i} = -y_i+\sum_{j=1}^{N}{w_{ji}\sigma(g_j(y_j+\theta_j))}+I_i
\end{equation}
where $y_i$ was the state of the $i$th neuron, $\tau_i$ was the neuron's
time constant, $w_{ji}$ was the weight of the connection from the $j$th
to $i$th neuron, $\theta_i$ was a bias term, $g_i$ was a gain term,
$I_i$ was an external input, and $\sigma(x)=1/(1+\exp(-x))$ was the
standard logistic output function.

The weights were randomly initialized. By considering the study of the
parameter space structure of CTRNN by \cite{Beer2006}, the values of
the $\theta$s were set based on the weights in a way that the richest
possible dynamics were achieved.  We used 11~nodes where two
nodes received the sensor inputs and one node was used as the output
node determining the direction of the movement (right/left).

\subsection{How we expect that evolution to solve the Wankelmut task}

Although the task might seem to be solved in a straight forward way, a
number of different strategies can be taken by an optimal
controller. Also notice that we tested our agents in two types of
environments in the following: One environment with the maximum at the
left hand side and another environment with the maximum at the right
hand side.

An intuitive solution is a controller that can go uphill or downhill
along the gradient with a 1-bit internal memory for determining the
current required movement strategy (uphill or downhill). The internal
memory switches its state when the sensor values reach the extremes
(defined thresholds). The initial state of the memory should indicate
the uphill movement. Hence, a controller requires only this 1-bit
internal memory. The comparison between the instant values of the
sensors then determines the actual direction of the movement in each
step. We would consider such a solution as a ``correct'' solution, as
it basically resembles the pseudo-code given in
Fig.~\ref{fig:pseudoCode}. We assume that our chosen topology of the
ANNs allow in principle to evolve the required behavior based on one
internal binary state variable. Our simple ANN was a recurrent network
and had two hidden layers with three nodes each. These two hidden
layers should have provided enough options to evolve an independent
(modular) uphill and downhill controller in combination with a 1-bit
memory and an action selection mechanism. Similarly, for the CTRNN
approach we have used nine neurons and any topology between them is
allowed.

Another solution to solve the task can be done in the following
way: The controller starts by deciding about the initial direction of
the agent's movement to the right or left depending on the initial sensor
value. In the following it continues a blind movement (i.e., not considering
the directional information of the sensor inputs) until extreme sensor
values are perceived and then switches the direction of the agent. Also
here an internal 1-bit state variable is required to keep the direction of the 
agent's movement.

There is even another solution possible. As our environment does not change in
size, the controller does not even need to consider the sensor values
at any time except the first time step: In this strategy, the robot has to initially
classify the environment and then the remaining task can be completed
correctly by choosing one of two ``pre-programmed'' trajectories.
In this solution, the sensor information is only used at the first time step but instead, a more complicated memory (more than 1-bit) is needed to maintain the oscillatory movement between the two extremes.
This is not a maximally reactive solution meaning that it partially replaces reactivity to sensor inputs with other mechanisms; i.e., using extra memory and pre-programmed behaviors.
Such a solution would not work in the way we expect it to work in other 
environments (e.g., changed size of the gradient). 
Such a controller might achieve the optimal fitness in our evolutionary runs, but a post-hoc
test in a slightly different environment would identify that it generates
sub-optimal behavior. Thus, it does not represent a valid solution
for the Wankelmut task. A post-hoc test that detects such a solution could be done by changing
the size or the steepness of the environmental gradient or by using a
Gaussian function (bell curve) instead of the Gauss error function
(erf) with randomized starting positions: In this case the agent should
oscillate only in the left or in the right half of the environment, depending on its starting 
position, to implement a true reactive solution to the Wankelmut task.

Other solution strategies can also concern the switching condition, for example
the switching may occur based on the extreme values of the sensors
(defined thresholds), the difference between the two sensor values, or
the fact that at the boundaries of the arena, the sensor values may
not change even if the agent attempts to keep moving in the same
direction, as we do not allow the agent to leave the arena.


Here, we are interested in controllers which are maximally reactive,
meaning that they base their behaviors on the reacting to sensor
values instead of scheduling pre-programmed trajectories.  The usage
of memory in such controllers is minimized since memory is replaced by
reactivity to sensors wherever it is possible. In our case, a valid
controller is expected to need a sort of 1-bit memory (that is not
replaceable by reactivity to sensors) to keep track of the direction
of its movement. Solutions that use extra memory for scheduling their
pre-programmed trajectories are considered invalid.

We are aware that the ``creativity'' of evolutionary computation cuts
both ways. It can surprise by ``cheap tricks'' that maximize fitness
without producing the desired agent behavior. Post-hoc, such a result
is usually considered to be an attribute of bad fitness function
design~\cite{nelson_2009_fitness}. Therefore, we define several
fitness functions and have tested two software control
techniques (ANN and CTRNN) using each fitness function
in 30~evolutionary runs per setting as it is described in the following
quantitative analysis section.

\section{Quantitative analysis}
\subsection{Fitness Evaluation}
In the experiments reported here, we have used different fitness regimes based on the number of the correct switches and the positioning of the agents.
In order to define the different fitness regimes, three types of rewards are considered:
\begin{itemize}
\item \textbf{Rewarding for switch}: rewarding $+1$ point for every correct switch: $R_{switch} = n$, where $n$ is the number of correct switches over the whole period of the experiment.
\item \textbf{Cumulative rewarding for positions}: rewarding based on the state and the environmental quality of the positions of the agent accumulated over the whole period of the experiment:\\ $R_{cum} = \sum_{t=0}^{T}quality[P(t)]*state$. That means that an agent in uphill mode ($state = 1$) is rewarded the current environmental quality of its position
and an agent in downhill mode ($state = -1$) is rewarded the current quality multiplied by -1. In consequence,
agents in uphill mode are rewarded positively for locating themselves in high-quality regions and agents in
downhill model for locating themselves in low-quality regions.
\item \textbf{Instant rewarding for final position}: rewarding based on the state and the environmental quality of the final position of the agent: $R_{ins} = quality[P(T)]*state$, where $T$ is the period of experiment. In this case, the reward is given for the final position of the agent. The exact value of the reward depends on the final state of the agent, but in any case a higher reward is given if the agent is closer to the next correct switch. That means, if the agent is in the uphill mode, a higher reward is given if it is higher up the hill and if it is in the downhill mode, a higher reward is given if it is more down the hill.
\end{itemize}

By giving different weights to the three types of the rewards, the fitness function is defined as follows:
\begin{equation}
F(w_{n}, w_{cum}, w_{ins})=w_{switch} \cdot R_{switch}+w_{cum}\cdot R_{cum}+w_{ins} \cdot R_{ins}
\end{equation}
where $w_{switch}$, $w_{cum}$, and $w_{ins}$ are parameters of the fitness function representing the weights for every type of the rewards.

In this study, the following fitness regimes are investigated: 
\begin{enumerate}
\item \textbf{Switch:} $F(1,0,0)$, rewarding only for correct switches.
\item \textbf{Cumulative} $ F(0,1,0)$, only cumulative rewarding for positions.
\item \textbf{Instant+Switch}: $F(100,0,0.01)$, rewarding for correct switches as well as instant rewarding for final position.
\item \textbf{Cumulative+Switch}:  $F(100,0.01,0)$, rewarding for correct switches as well as cumulative rewarding for positions over the whole period.
\end{enumerate}

\subsection{Evolving Solutions in Various Scenarios}

In the following, we were using a number of evolutionary setups (scenarios)
and investigated the resulted controllers achieved by evolution in each
setup. Two different controller types, ANN and CTRNN were used. In
all the scenarios, every controller ran for 250~time-steps unless
otherwise stated. Results were pooled from 30~independent runs for
each scenario.

We used a simple genetic algorithm~\cite{holland75} with proportionate
selection based on fitness and elitism of one. The population size was set to~150 and we
ran for 300~generations. A genome for the ANN consisted of 41 genes each encoding a connection weight between the nodes. The weights are randomly initialized between $(-0.5,0.5)$. 
A genome in the CTRNN consisted of 121 genes for the weights of the connections, 11 genes for the values of $\theta$ and 11 genes for the $\tau$ values of every node. The weights and $\tau$ values are randomly initialized in the range of $(-15,15)$ and $(0.9,5.9)$ respectively. The values of $\theta$s are set based on the weights such that the richest possible dynamics is achieved, as described in~\cite{Beer2006}.
We did not use a recombination operator.
In ANN, the weights are mutated with a rate of $0.3$ where a random value in $(-0.4,0.4)$ is added to the weight.
In CTRNN the mutation rate if $0.1$ for weights and a random $\theta$ and $\tau$ are mutated at a time. The values are changed in the range of $(-0.4,0.4)$ for weights and $\theta$, and in the range of $(-0.1,0.1)$ for $\tau$. Table~\ref{table:ec} summarizes the evolutionary parameters.

\begin{table}[ht]\small
\centering
  \caption{Parameters used for the evolutionary algorithm
	\label{table:ec}}
\begin{center}
\begin{tabular}{|lc|lc|}
  \hline
  parameter & ANN & parameter & CTRNN\\
  \hline
  population size & 150 & population size & 150\\
  number of genes & 41 &  number of genes (weights) & 121\\
   & &  number of genes ($\theta$) & 11\\
   & &  number of genes ($\tau$) & 11\\
   init. range (weight) & $(-0.5,0.5)$ &  init. range (weight) & $(-0.5,0.5)$ \\
   & & init. range ($\tau$) & $(0.9,5.9)$ \\
   & & init. range ($\theta$) & as in~\cite{Beer2006} \\
   mutation rate & 0.3 & mutation rate (weights) & 0.1 \\
   &  & mutation rate ($\theta$, $\tau$) & a random one at a time \\
   \hline
\end{tabular}
\end{center}
\end{table}

\subsubsection{Single Environment vs. Double Environment, Mean fitness versus Minimum Fitness}

\begin{figure}[h!]
\vspace*{2cm}
\begin{minipage}[b]{.45\linewidth}
\includegraphics[width=1.0\textwidth]{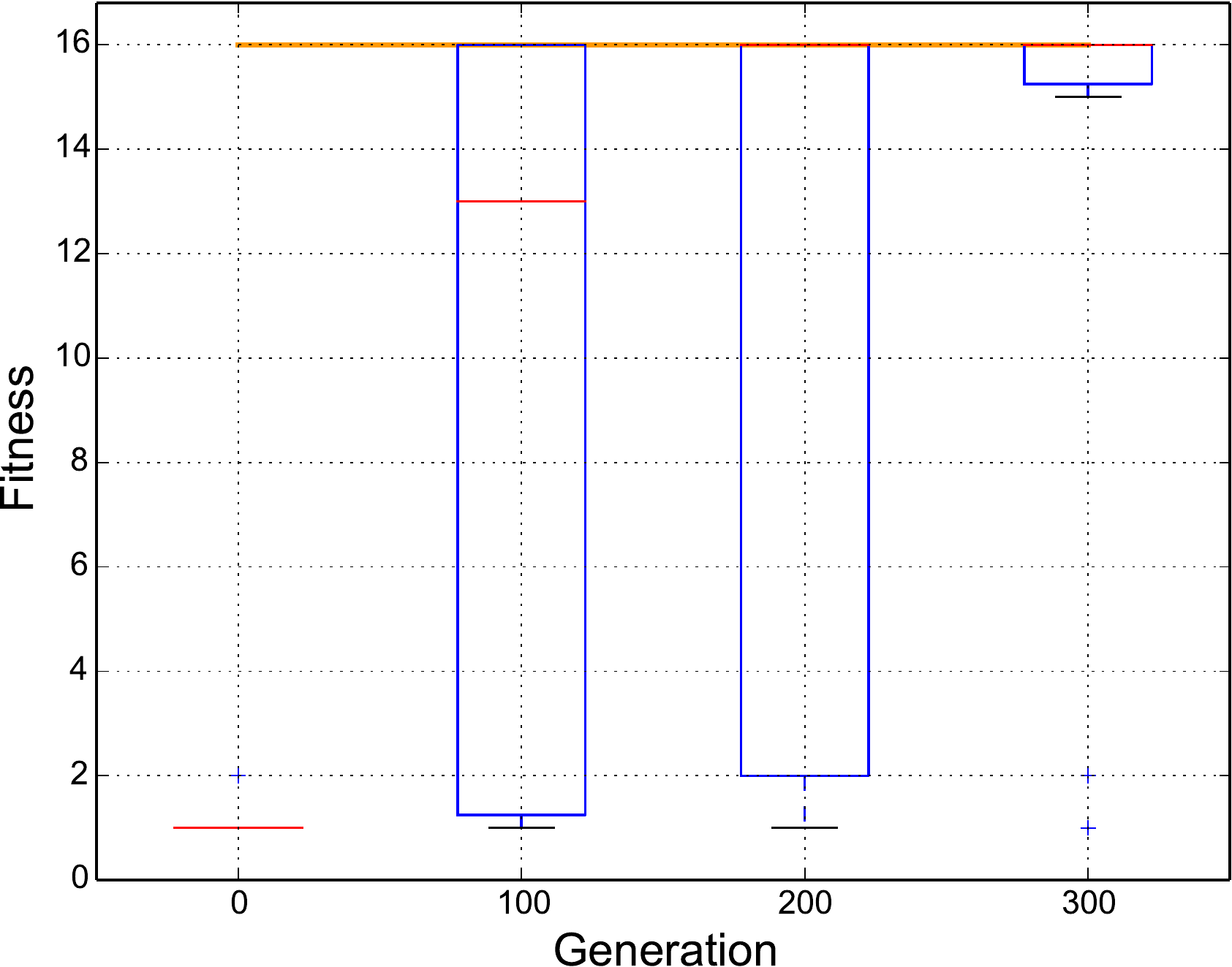}
\subcaption{ANN, fitness}\label{fig:results:single-sw:a}
\end{minipage}%
\hspace*{.05\linewidth}%
\begin{minipage}[b]{.45\linewidth}
\includegraphics[width=1.0\textwidth]{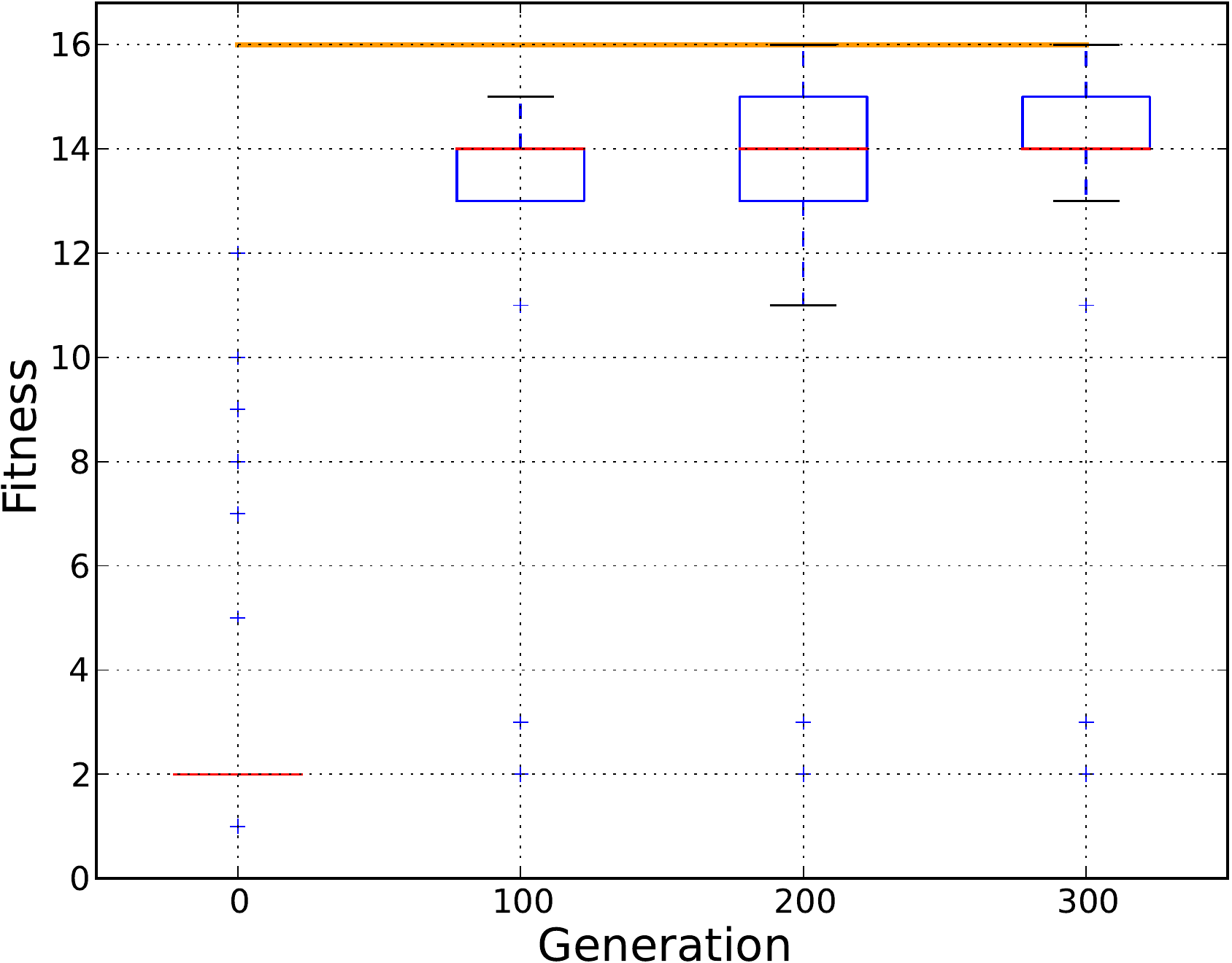}
\subcaption{CTRNN, fitness}\label{fig:results:single-sw:b}
\end{minipage}
\begin{minipage}[b]{.24\linewidth}
\includegraphics[width=1.0\textwidth]{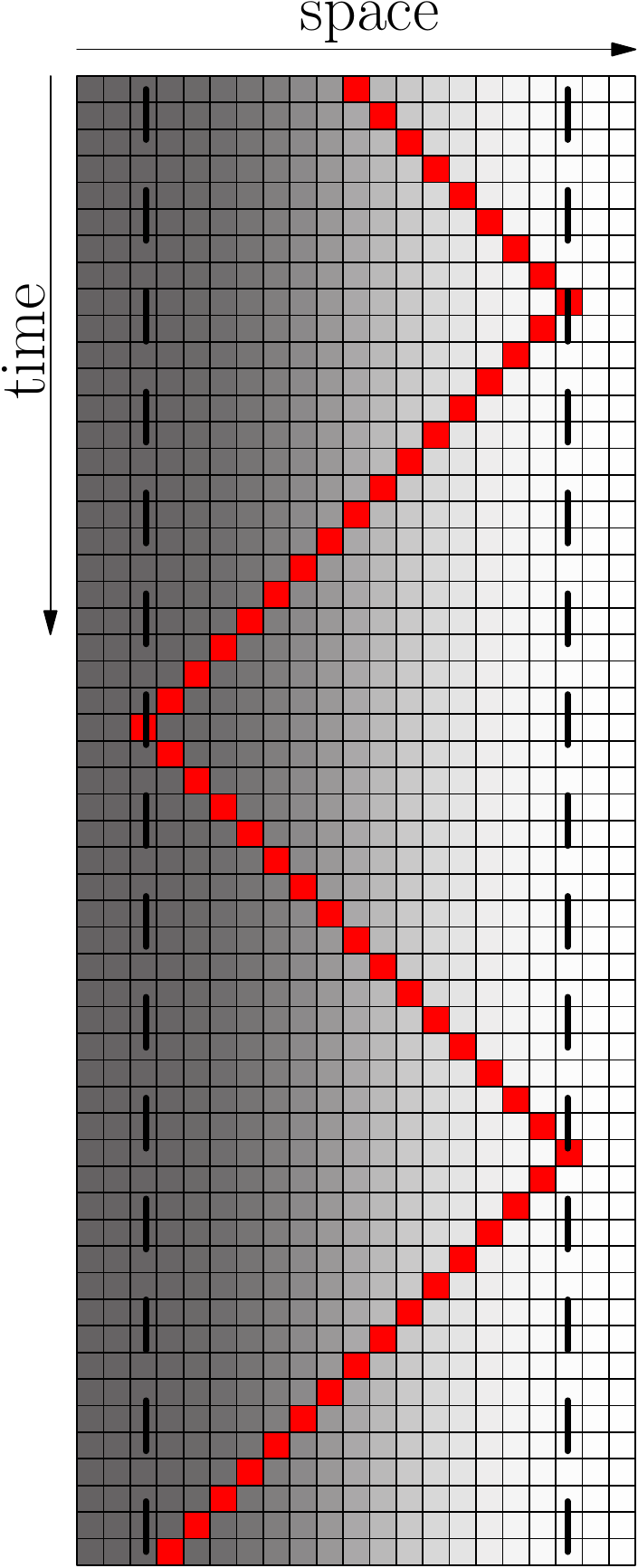}
\subcaption{ANN, environment used in evolution}\label{fig:results:single-sw:c}
\end{minipage}
\begin{minipage}[b]{.24\linewidth}
\includegraphics[width=1.0\textwidth]{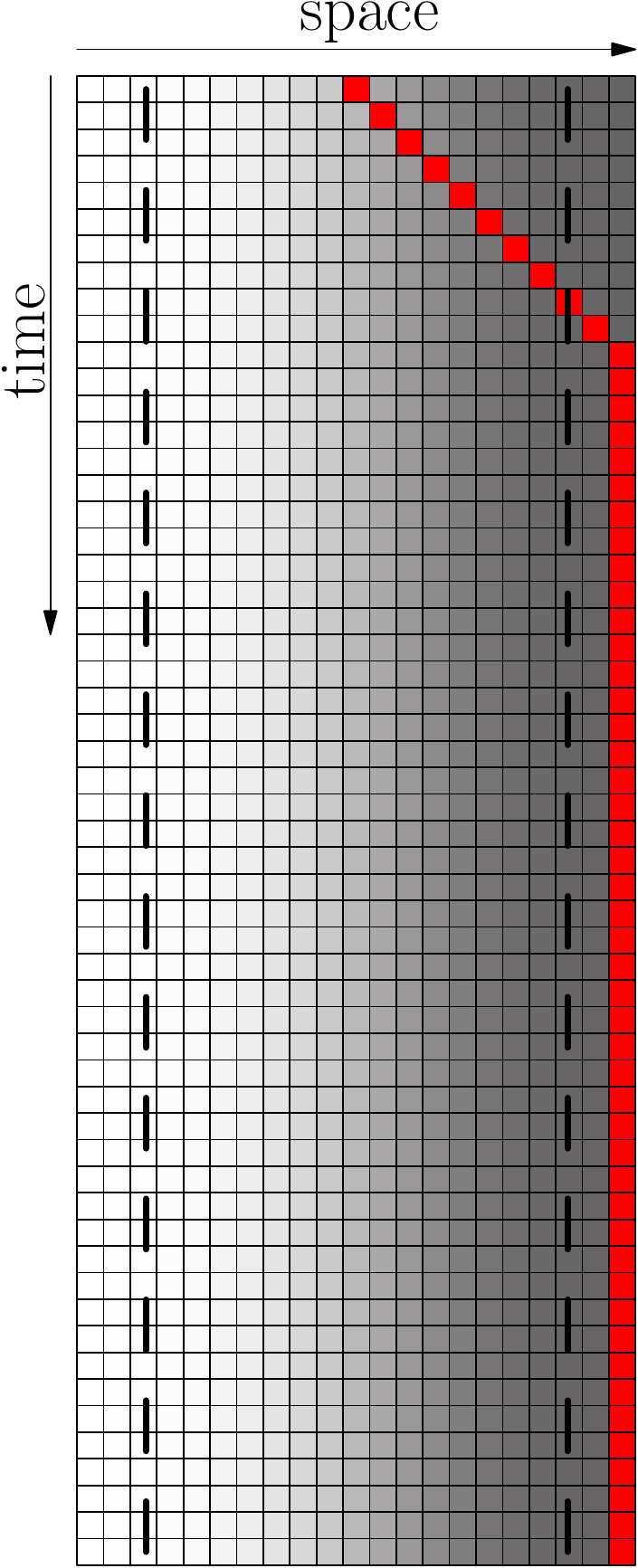}
\subcaption{ANN, environment used in post-hoc eval.}\label{fig:results:single-sw:d}
\end{minipage}
\begin{minipage}[b]{.24\linewidth}
\includegraphics[width=1.0\textwidth]{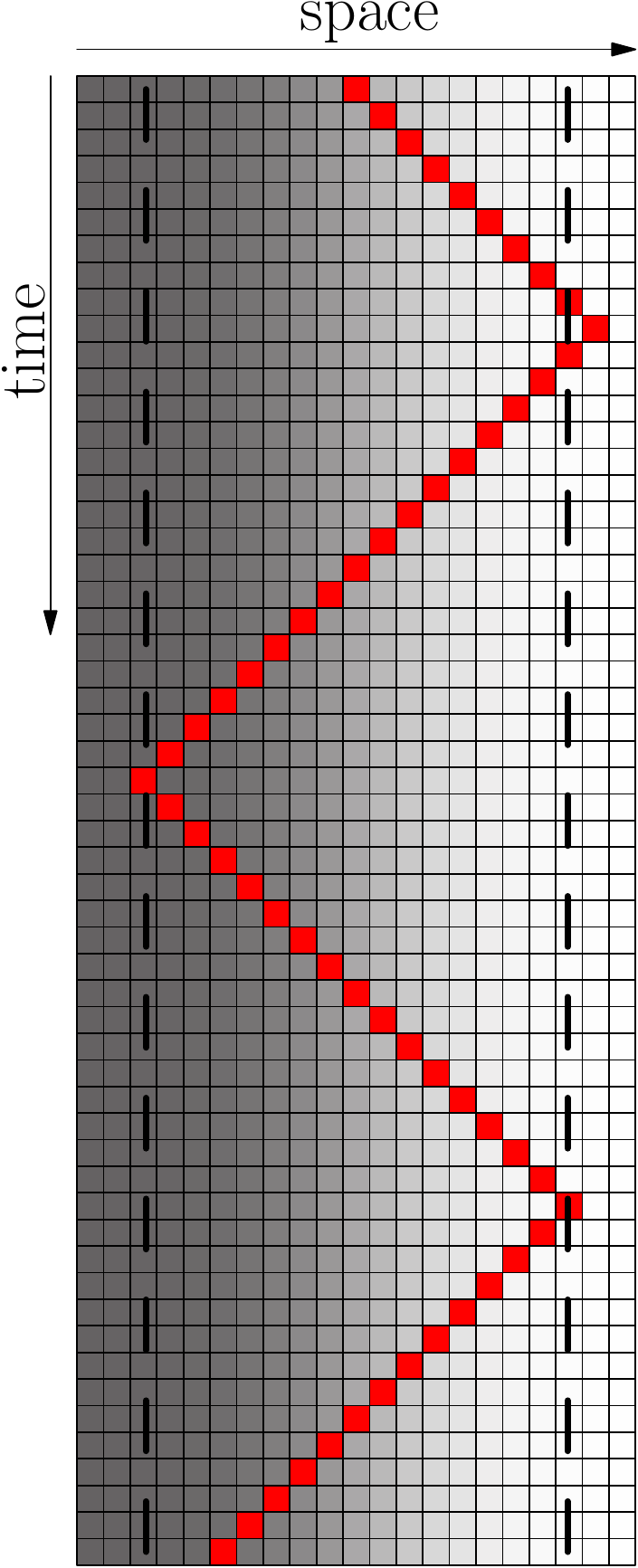}
\subcaption{CTRNN, environment used in evolution}\label{fig:results:single-sw:e}
\end{minipage}
\begin{minipage}[b]{.24\linewidth}
\includegraphics[width=1.0\textwidth]{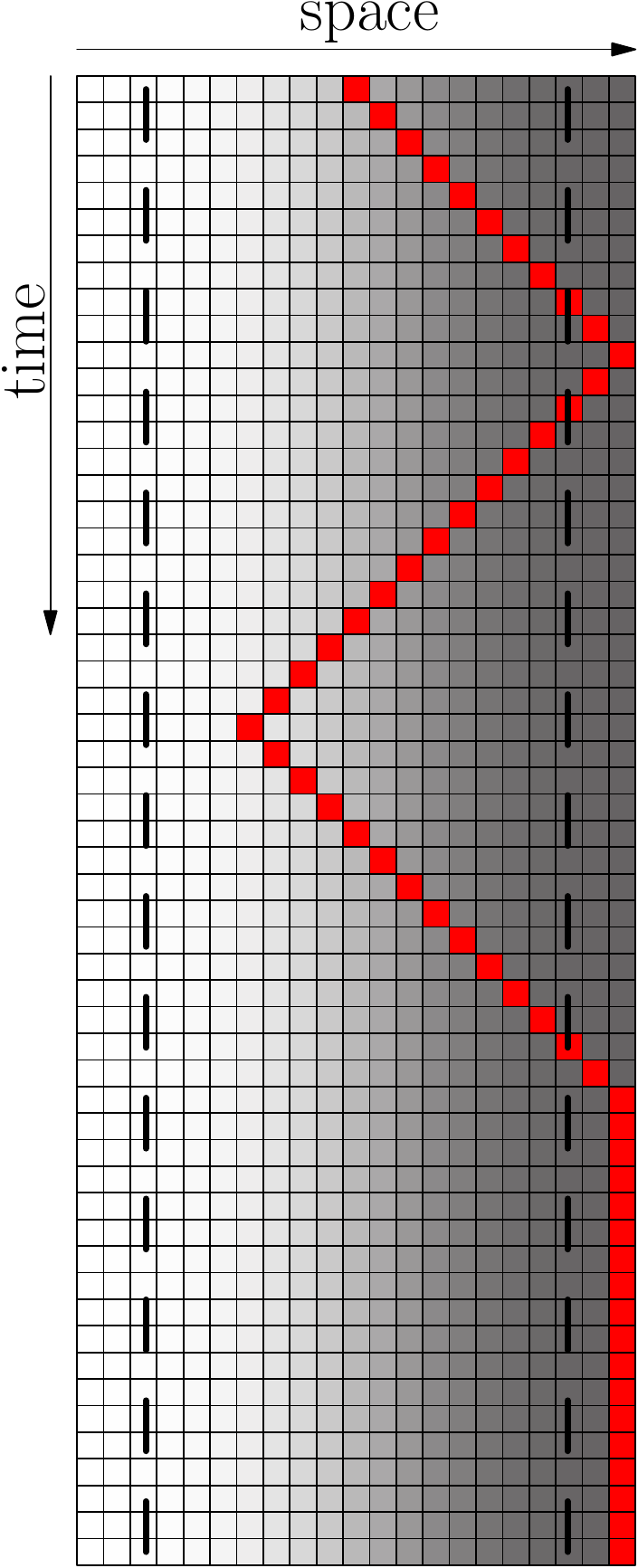}
\subcaption{CTRNN, environment used in post-hoc eval.}\label{fig:results:single-sw:f}
\end{minipage}
\caption{Single Environment, \emph{Switch} $\text{fitness} =
  F(1,0,0)$, Top row: fitness dynamics of ANNs (left) and CTRNNs
  (right). The horizontal orange line indicates the maximum fitness
  achievable by a reactive Wankelmut controller, as it is achieved by
  the simple hand-coded controller. Bottom row: trajectory of the best
  performing genome of each controller type in the environment used in
  evolution versus a run in a flipped environment as a post-hoc
  evaluation (only the initial 56 time steps of 250 time steps are shown).}\label{fig:results:single-sw}
\end{figure}

At first we evaluated the performance of evolution in a fixed
environment, like it is depicted in Fig.~\ref{fig:task}.  We found
that evolution could quickly converge to a sort of pre-programmed
trajectory (see Fig.~\ref{fig:results:single-sw:c} and
Fig.~\ref{fig:results:single-sw:e}) that achieved very high fitness (see
Fig.~\ref{fig:results:single-sw:a} and
Fig.~\ref{fig:results:single-sw:b}) but was non-reactive as it was not
considering sensor inputs in the wanted way: A post-hoc test of the best evolved genomes
in a flipped environment, where the gradient pointed to the other side,
was failed by both evolved controller types, clearly indicating that no
reactive Wankelmut behavior has evolved (see
Fig.~\ref{fig:results:single-sw:d} and
Fig.~\ref{fig:results:single-sw:f}). The trajectories in both post-hoc runs 
show some difference to the runs in the environment that was used
in evolution, this indicates that some sensor-input was affecting the behavior
but not in the desired way: The agents still started off in the wrong direction
and also the oscillations ceased after some time.

As this approach was not yielding the wanted
reactive Wankelmut behavior, we decided to evaluate each individual
genome twice: One time the gradient pointed uphill to the left and one
time it pointed uphill to the right side. The fitness function in all the three
scenarios used the \emph{Switch} fitness regime which is described as $\text{fitness} =
F(1,0,0)$.

\begin{figure}[h!]
\vspace*{1.5cm}
\begin{minipage}[b]{.45\linewidth}
\includegraphics[width=1.0\textwidth]{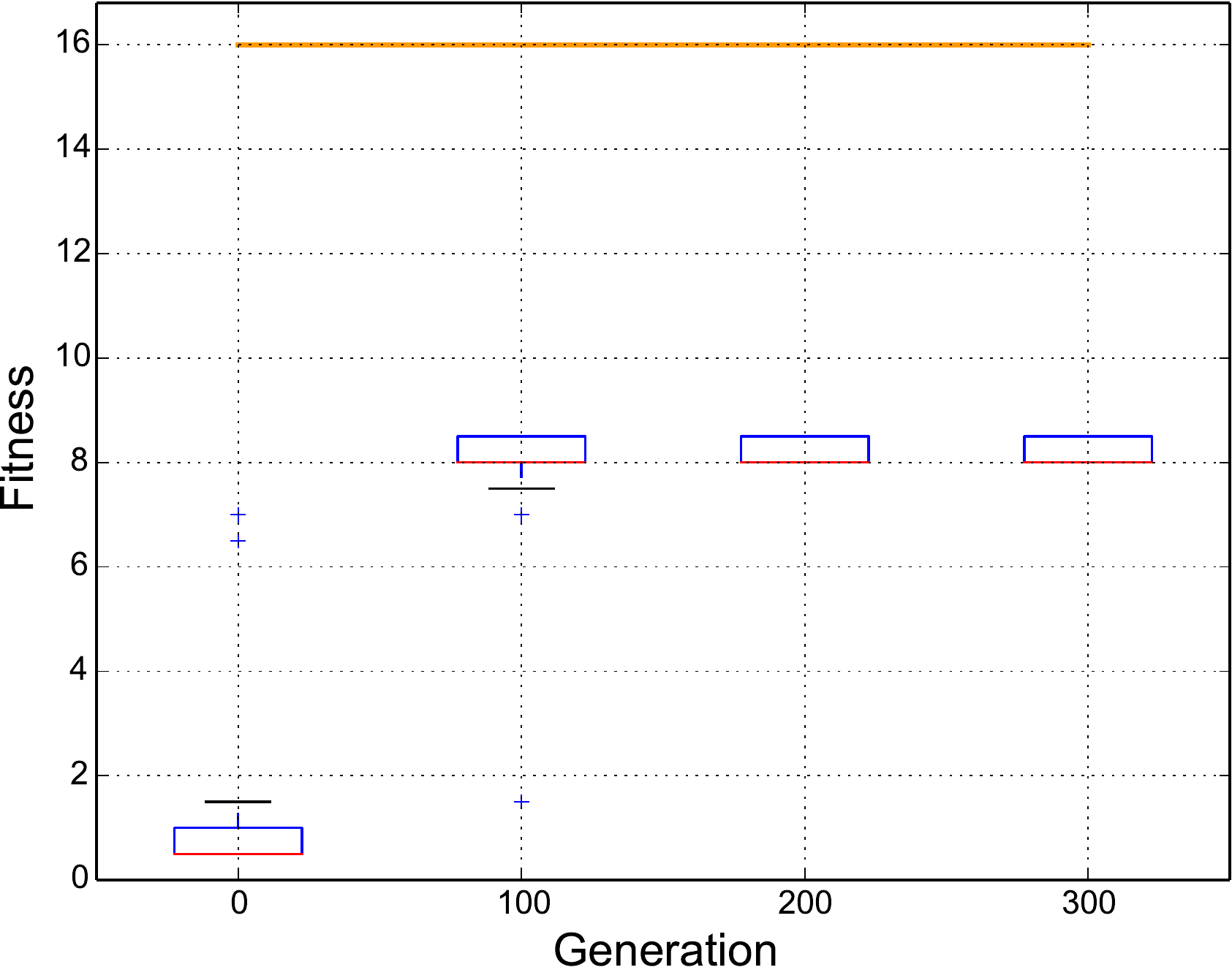}
\subcaption{ANN, fitness}\label{fig:results:mean-sw:a}
\end{minipage}%
\hspace*{.05\linewidth}%
\begin{minipage}[b]{.45\linewidth}
\includegraphics[width=1.0\textwidth]{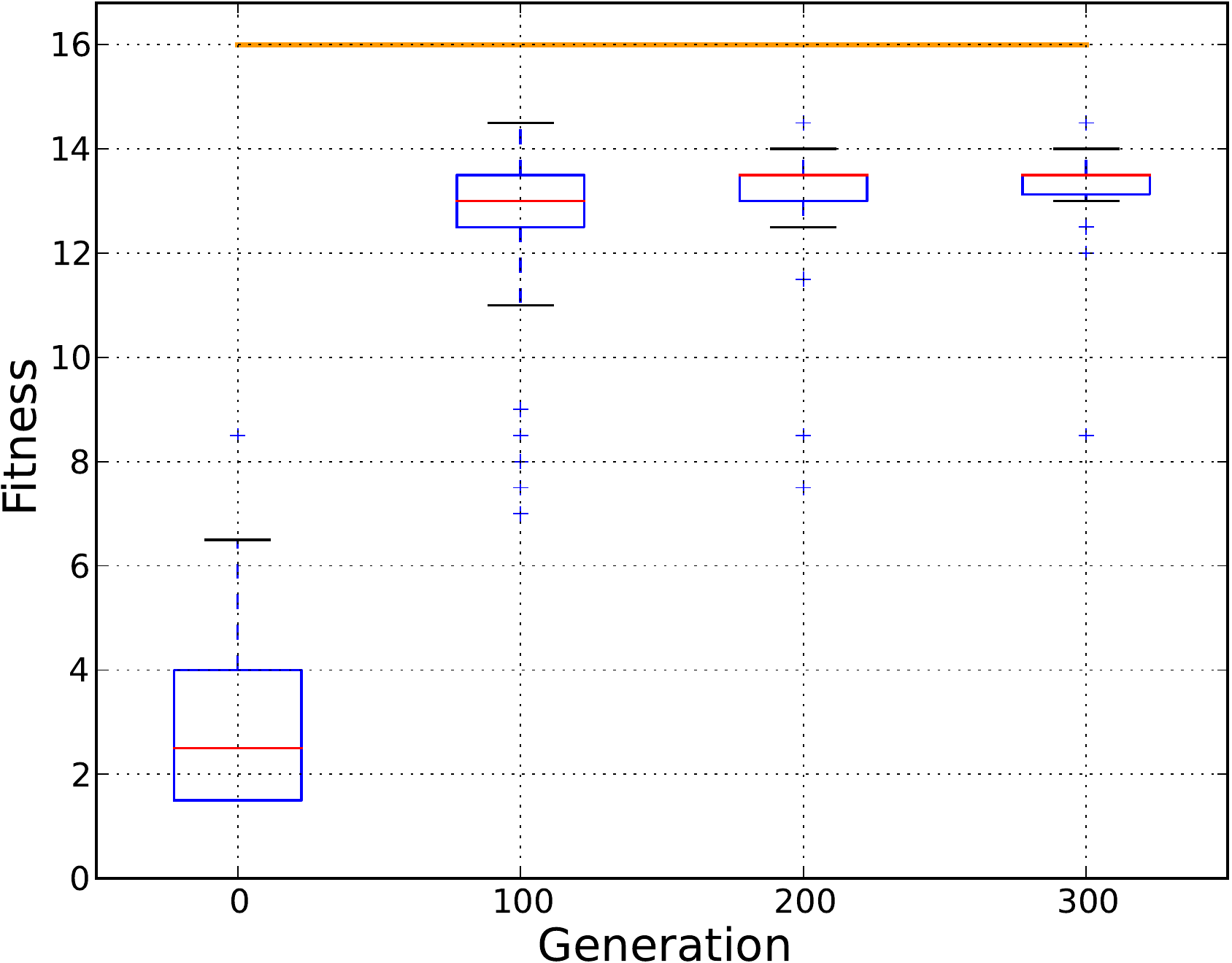}
\subcaption{CTRNN, fitness}\label{fig:results:mean-sw:b}
\end{minipage}
\begin{minipage}[b]{.24\linewidth}
\includegraphics[width=1.0\textwidth]{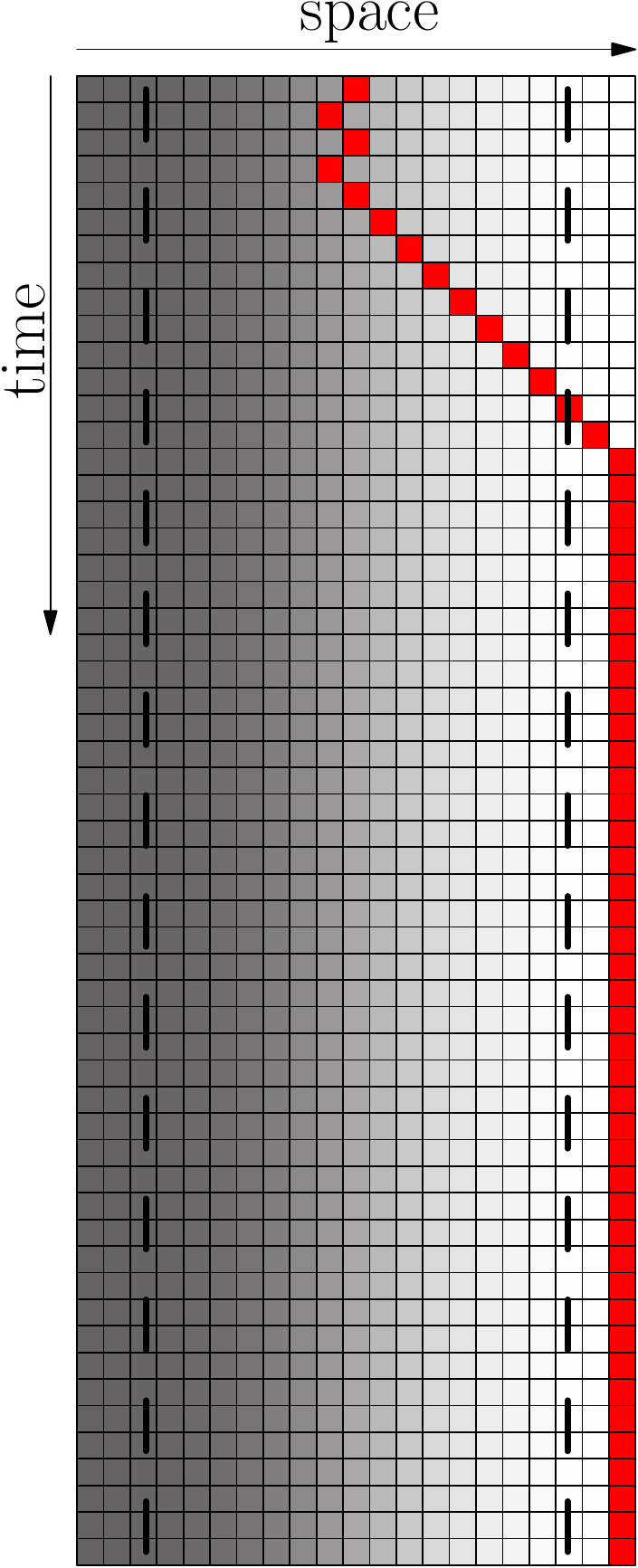}
\subcaption{ANN, environmental setting~1}\label{fig:results:mean-sw:c}
\end{minipage}
\begin{minipage}[b]{.24\linewidth}
\includegraphics[width=1.0\textwidth]{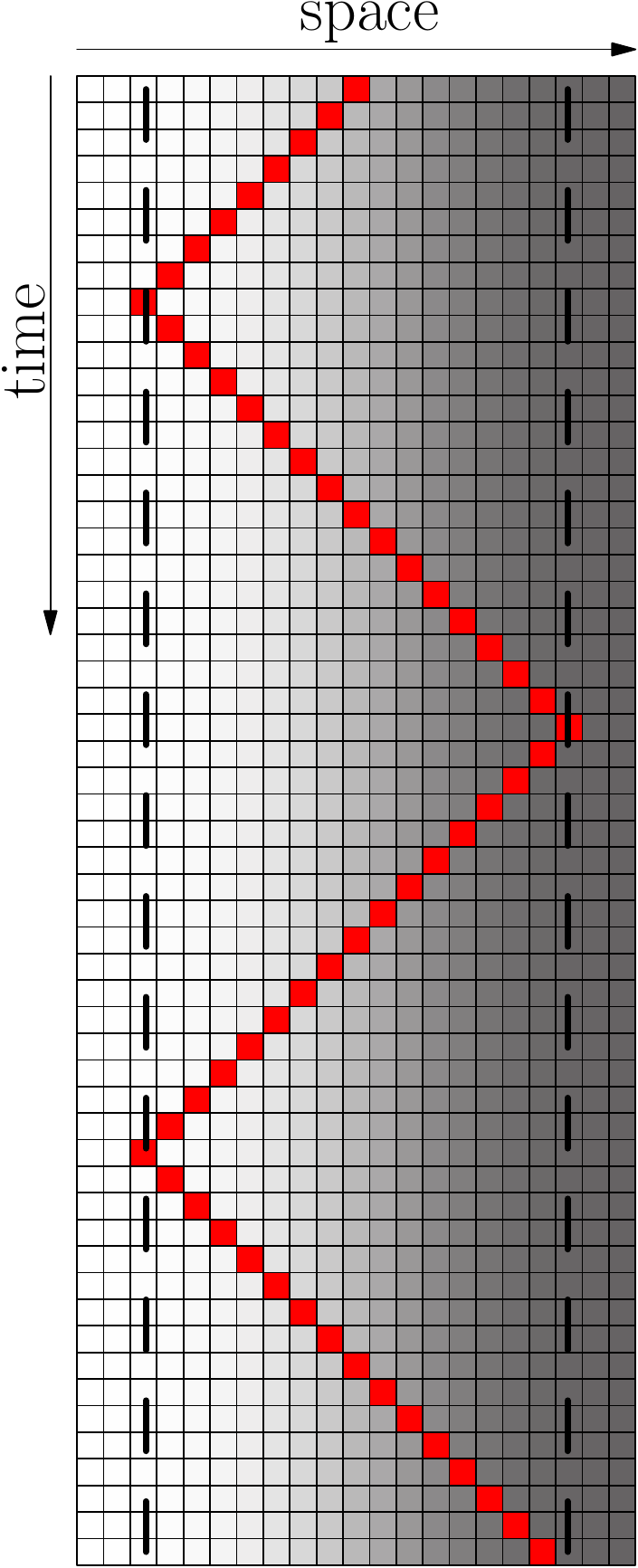}
\subcaption{ANN, environmental setting~2}\label{fig:results:mean-sw:d}
\end{minipage}
\begin{minipage}[b]{.24\linewidth}
\includegraphics[width=1.0\textwidth]{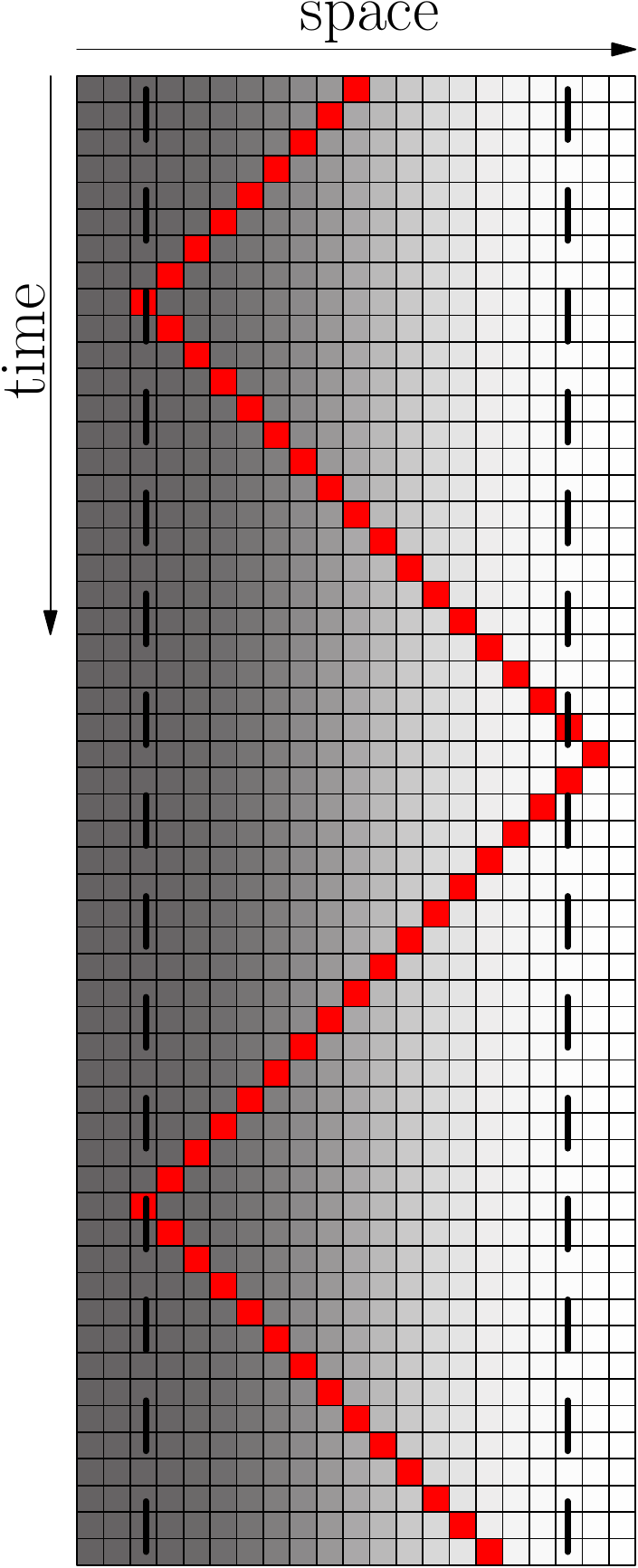}
\subcaption{CTRNN, environmental setting~1}\label{fig:results:mean-sw:e}
\end{minipage}
\begin{minipage}[b]{.24\linewidth}
\includegraphics[width=1.0\textwidth]{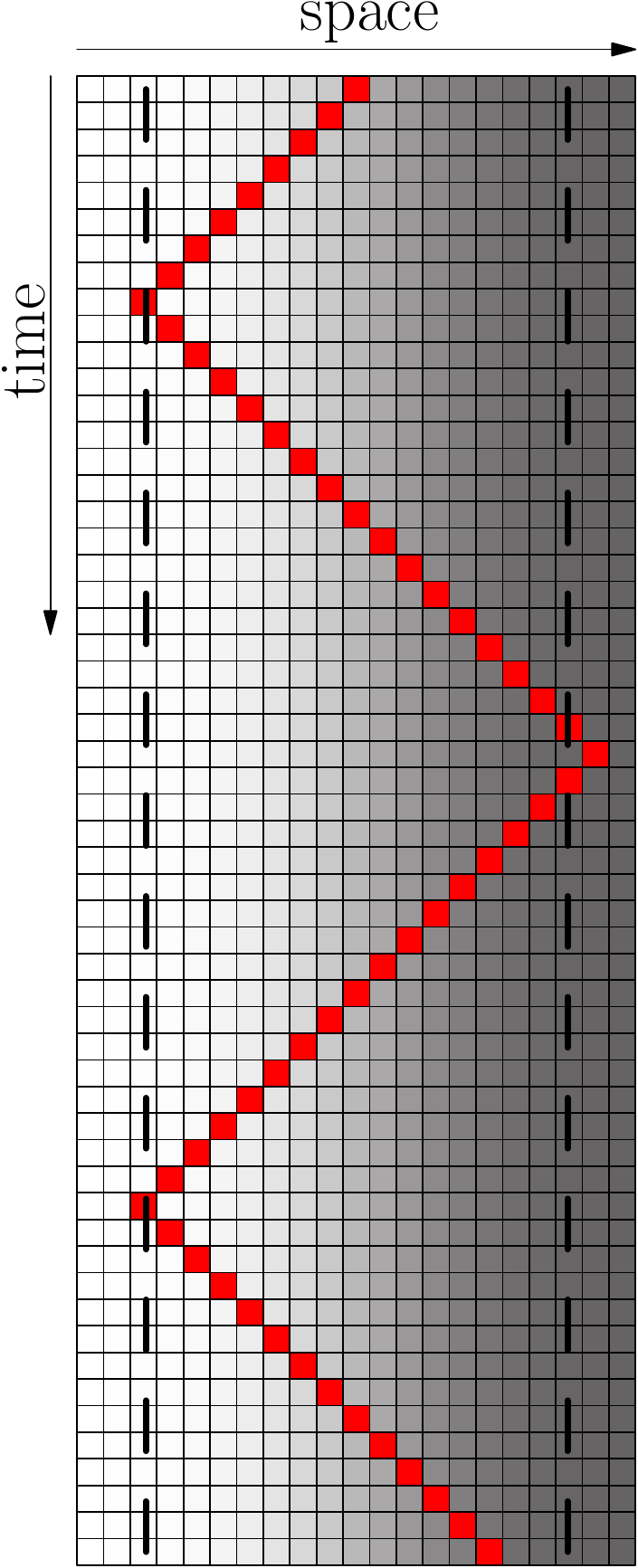}
\subcaption{CTRNN, environmental setting~2}\label{fig:results:mean-sw:f}
\end{minipage}
\caption{Evolutionary results of the double environment setting,
  fitness regime \emph{Switch}: $\text{fitness} = F(1,0,0)$. The {\bf
    arithmetic mean} of both evaluations was used as fitness value for
  a genome. Top row: fitness dynamics of ANNs (left) and CTRNNs
  (right). The horizontal orange line indicates the maximum fitness
  achievable by a reactive Wankelmut controller, as it is achieved by
  the simple hand-coded controller. Bottom row: trajectory of the best
  performing genome of both controller types in both environments used
  in evolution (only the initial 56 time steps of 250 time steps are shown).}\label{fig:results:mean-sw}
\end{figure}

\begin{figure}[h!]
\vspace*{1.5cm}
\begin{minipage}[b]{.45\linewidth}
\includegraphics[width=1.0\textwidth]{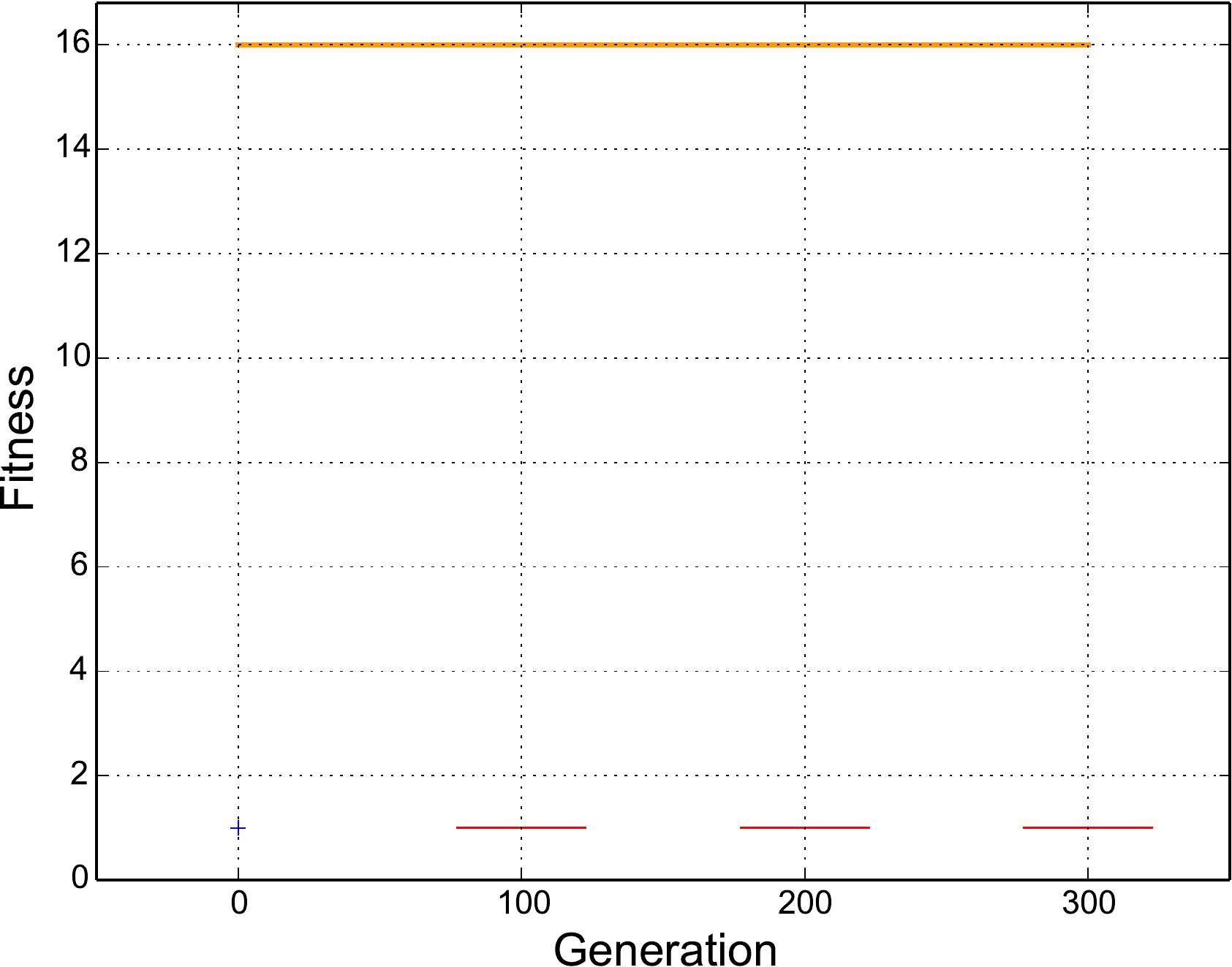}
\subcaption{ANN, fitness}\label{fig:results:min-sw:a}
\end{minipage}%
\hspace*{.05\linewidth}%
\begin{minipage}[b]{.45\linewidth}
\includegraphics[width=1.0\textwidth]{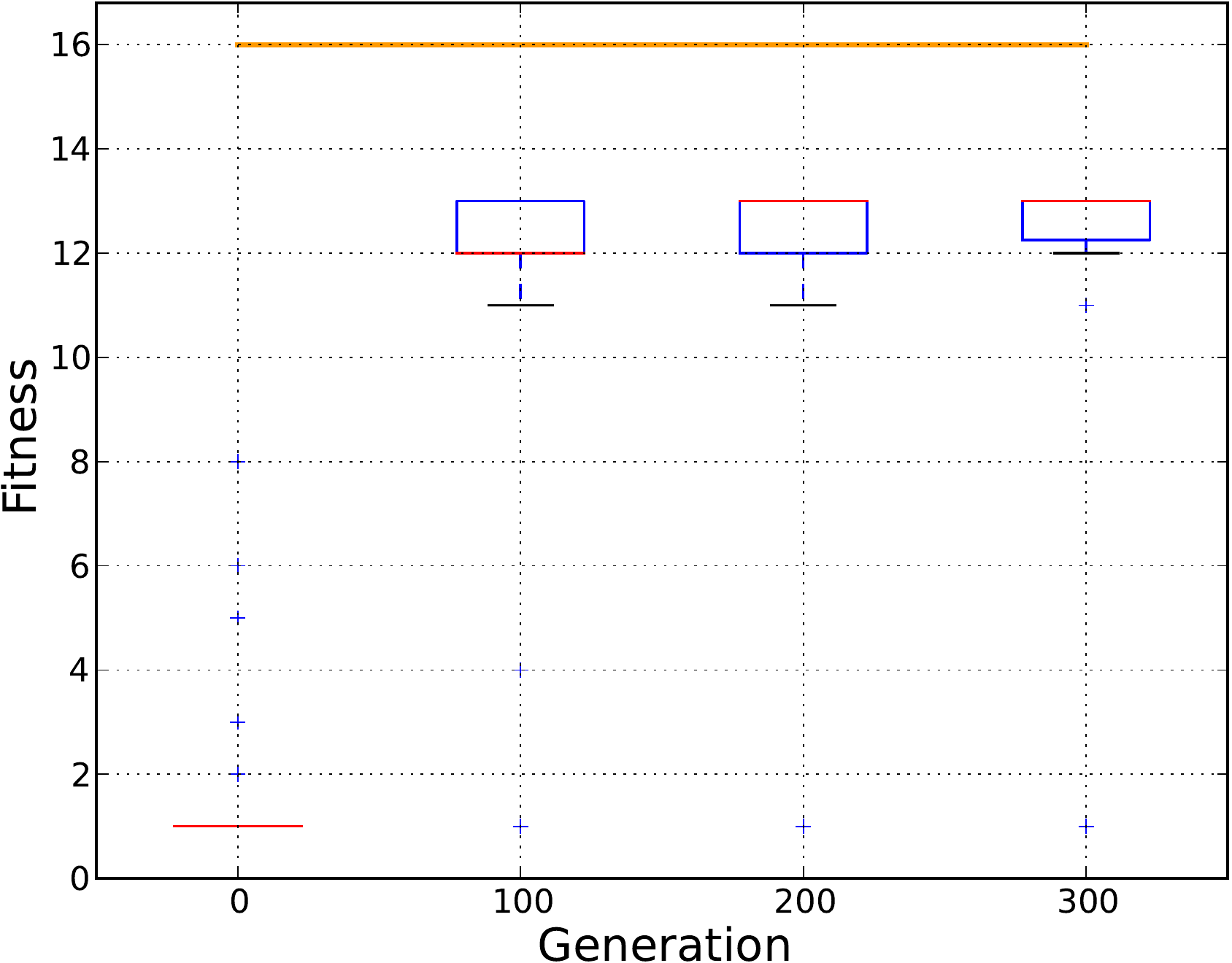}
\subcaption{CTRNN, fitness}\label{fig:results:min-sw:b}
\end{minipage}
\begin{minipage}[b]{.24\linewidth}
\includegraphics[width=1.0\textwidth]{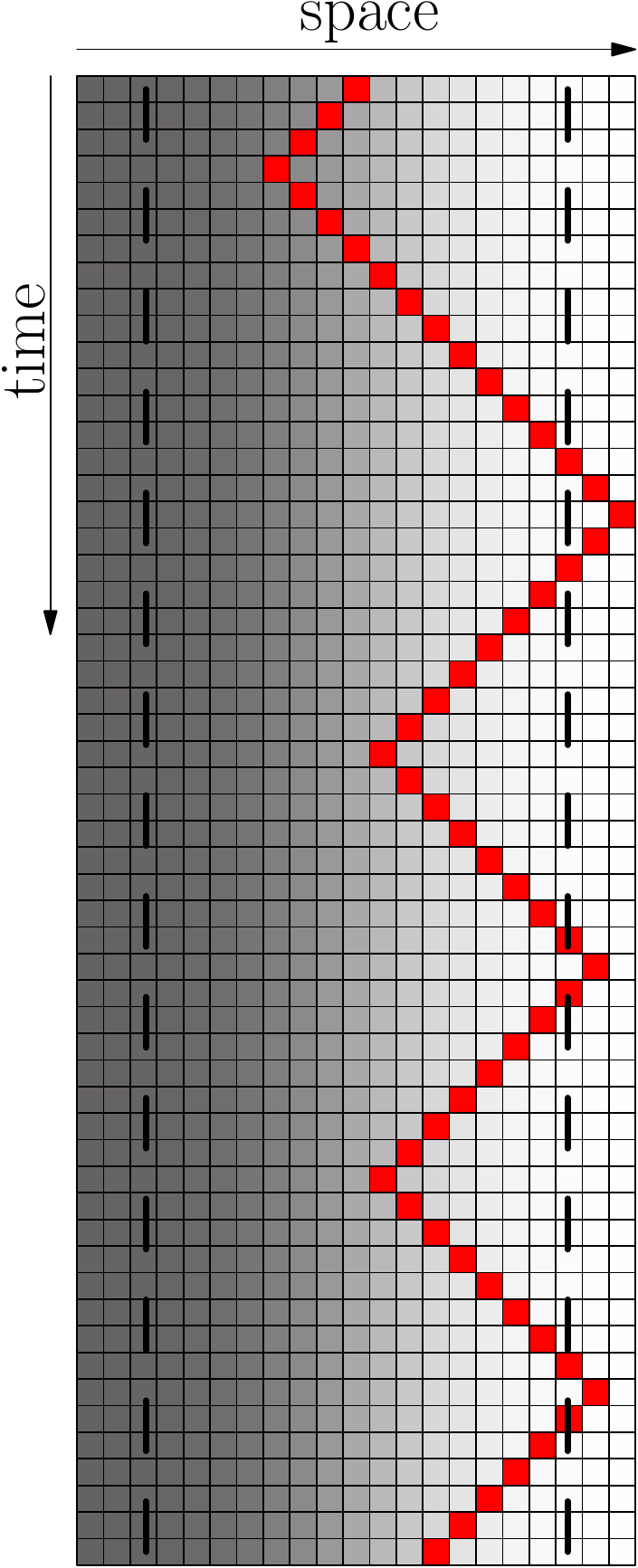}
\subcaption{ANN, environmental setting~1}\label{fig:results:min-sw:c}
\end{minipage}
\begin{minipage}[b]{.24\linewidth}
\includegraphics[width=1.0\textwidth]{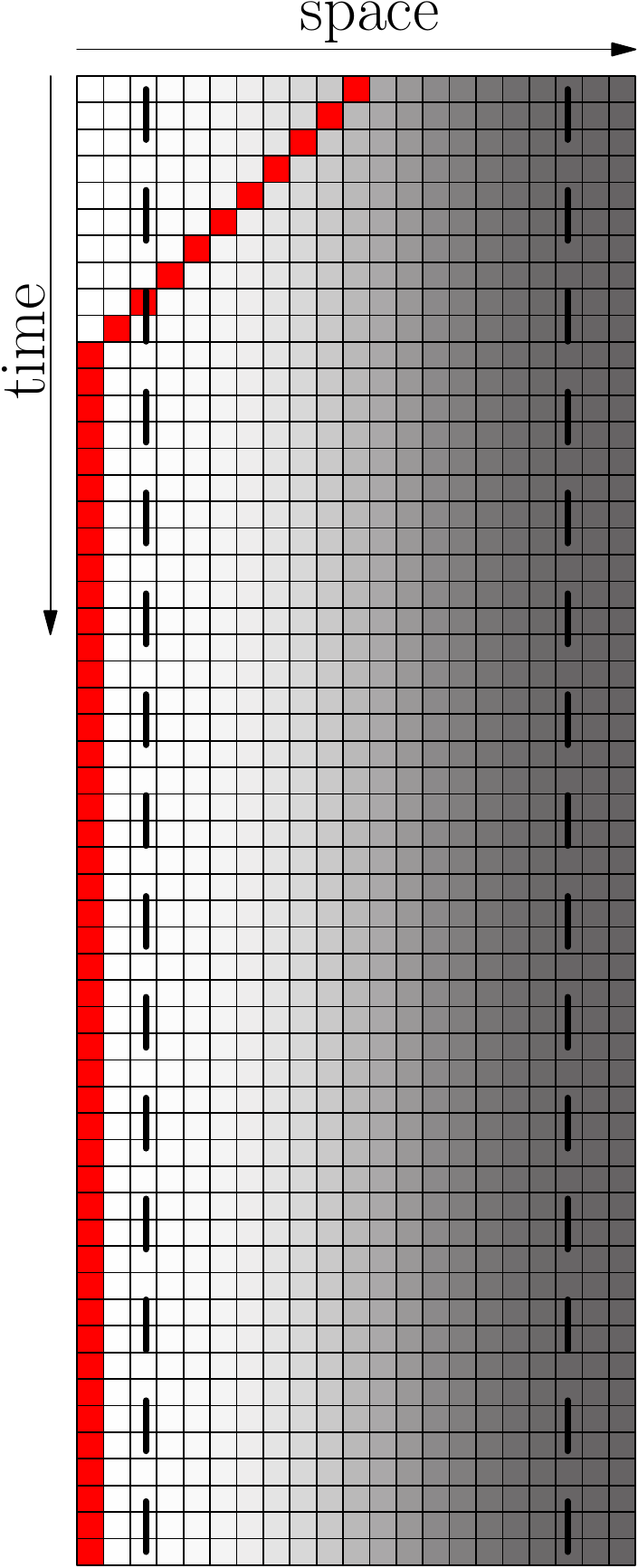}
\subcaption{ANN, environmental setting~2}\label{fig:results:min-sw:d}
\end{minipage}
\begin{minipage}[b]{.24\linewidth}
\includegraphics[width=1.0\textwidth]{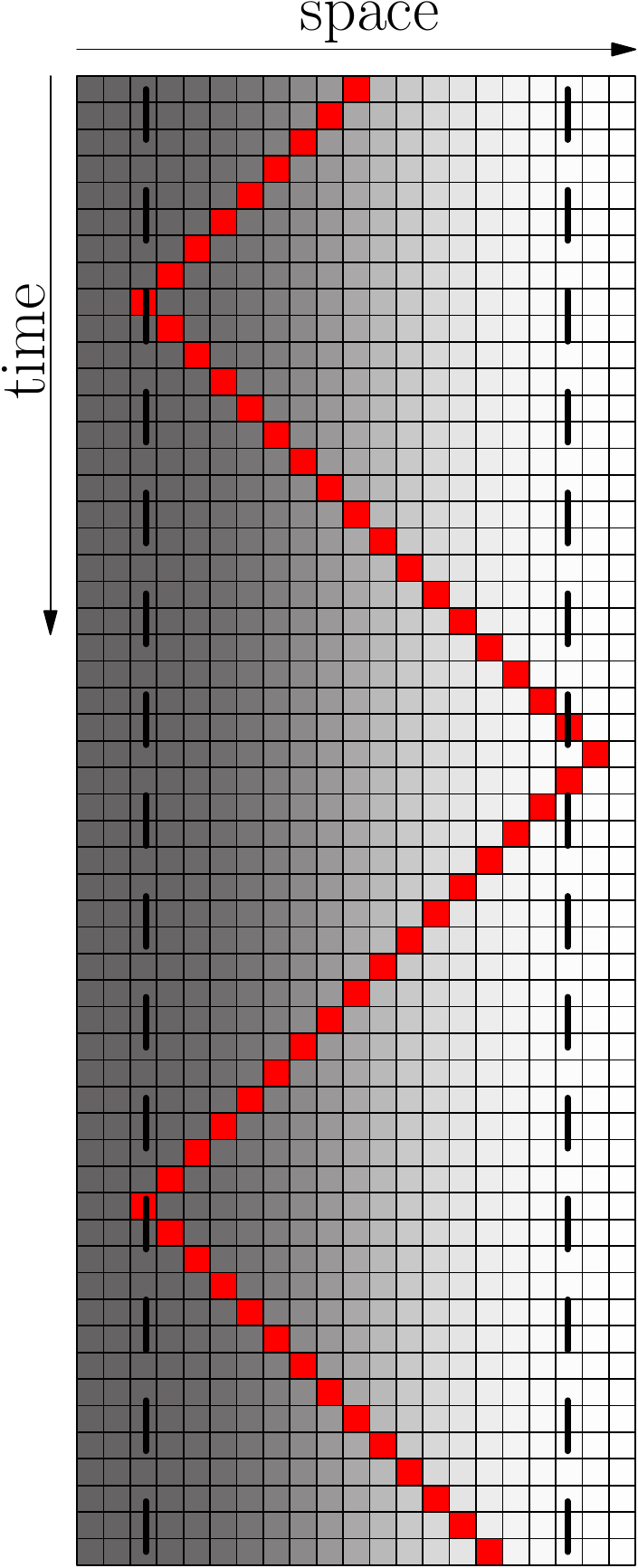}
\subcaption{CTRNN, environmental setting~1}\label{fig:results:min-sw:e}
\end{minipage}
\begin{minipage}[b]{.24\linewidth}
\includegraphics[width=1.0\textwidth]{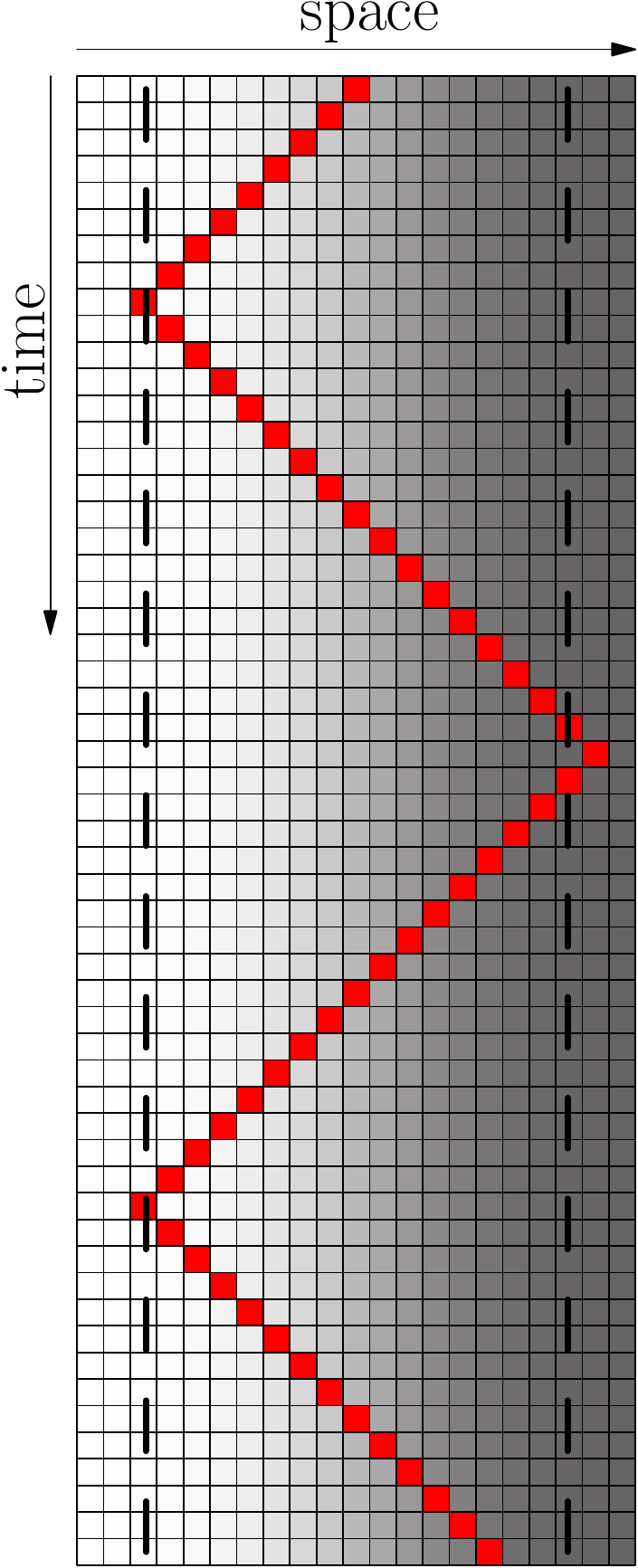}
\subcaption{CTRNN, environmental setting~2}\label{fig:results:min-sw:f}
\end{minipage}
\caption{Evolutionary results of the double environment setting,
  fitness regime \emph{Switch}: $\text{fitness} = F(1,0,0)$. The {\bf
    minimum} of both evaluations was used as fitness value for a
  genome. Top row: fitness dynamics of ANNs (left) and CTRNNs
  (right). The horizontal orange line indicates the maximum fitness
  achievable by a reactive Wankelmut controller, as it is achieved by
  the simple hand-coded controller. Bottom row: trajectory of the best
  performing genome of both controller types in both environments used
  in evolution (only the initial 56 time steps of 250 time steps are shown).}\label{fig:results:min-sw}
\end{figure}

We evaluated two methods of associating a fitness value to each genome
from these two evaluations: First, we calculated the arithmetic mean value of
both runs and, as a second variant, we took the minimum value of both
runs. In consequence, a good genome had to perform well in both
environments, in the ``minimum''-fitness function the selection was
even harsher then in the ``mean'' variant. We found that the
``minimum'' variant of selection showed the higher selection pressure
on evolving reactive control (compare Fig.~\ref{fig:results:mean-sw}
and Fig.~\ref{fig:results:min-sw}): With the minimum-fitness regime
the controllers had it harder to gather high fitness values, as they had
to perform well in both environments, preventing evolution from ``just 
considering on one of the two environments". We hoped that this
would support the evolution of reactive control. As Fig.~\ref{fig:results:min-sw:c}
to Fig.~\ref{fig:results:min-sw:f} show, it still did not
produce the desired reactive Wankelmut behavior in the long run.
Based on those findings, we used the double-environment evaluation
with a minimum-of-both-run fitness function for all further quantitative
analysis in this study, in order to maximize the selection pressure towards
reactivity of the evolved controller.

\subsubsection{\emph{Cumulative} fitness regime}

In another attempt to help the evolutionary algorithm to develop good
reactive controllers we decided to reward genomes purely with
cumulative fitness. The idea behind this was that every single step in
the correct direction would be paying off in evolution. We assumed that, on the one
hand, bootstrapping problems are minimized this way and, on the other
hand, early reactive turns into the correct directions are rewarded
even if they do not reach the switching threshold places initially.
In this scenario, the fitness was computed by the \emph{Cumulative}
fitness regime described as $\text{fitness} = F(0,1,0)$

\begin{figure}[h!]
\vspace*{1.5cm}
\begin{minipage}[b]{.45\linewidth}
\includegraphics[width=1.0\textwidth]{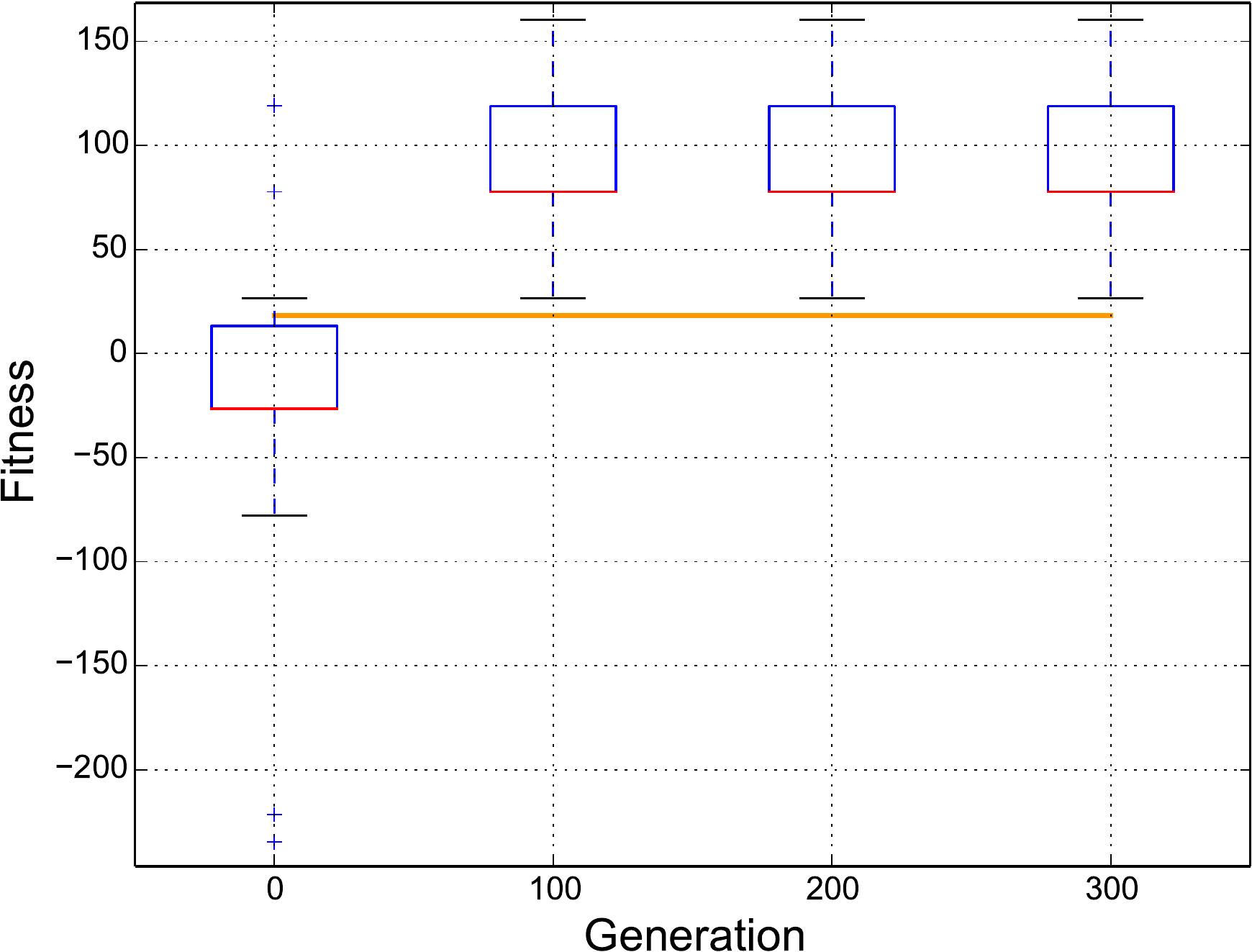}
\subcaption{ANN, fitness}\label{fig:results:cum:a}
\end{minipage}%
\hspace*{.05\linewidth}%
\begin{minipage}[b]{.45\linewidth}
\includegraphics[width=1.0\textwidth]{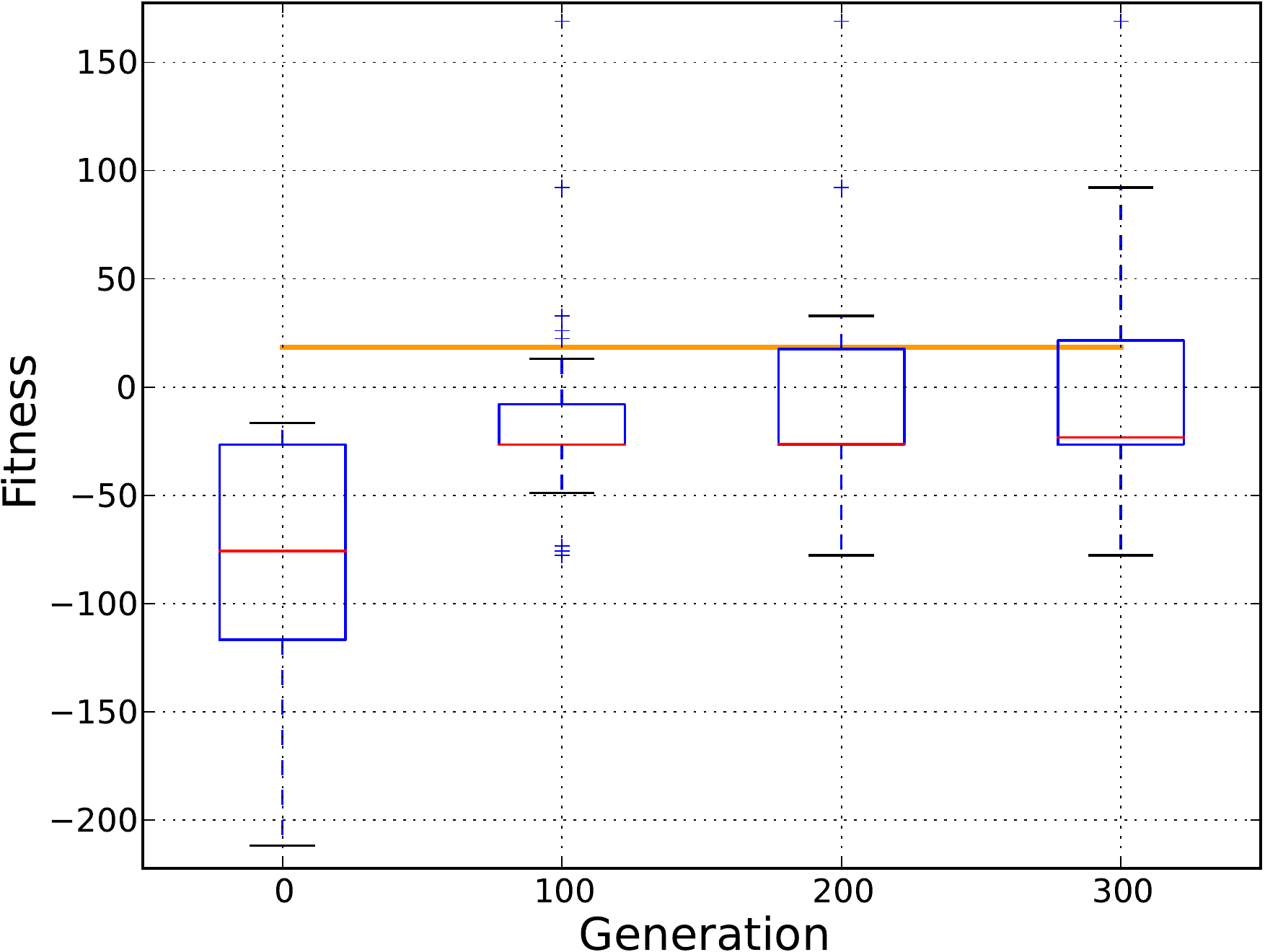}
\subcaption{CTRNN, fitness}\label{fig:results:cum:b}
\end{minipage}
\begin{minipage}[b]{.24\linewidth}
\includegraphics[width=1.0\textwidth]{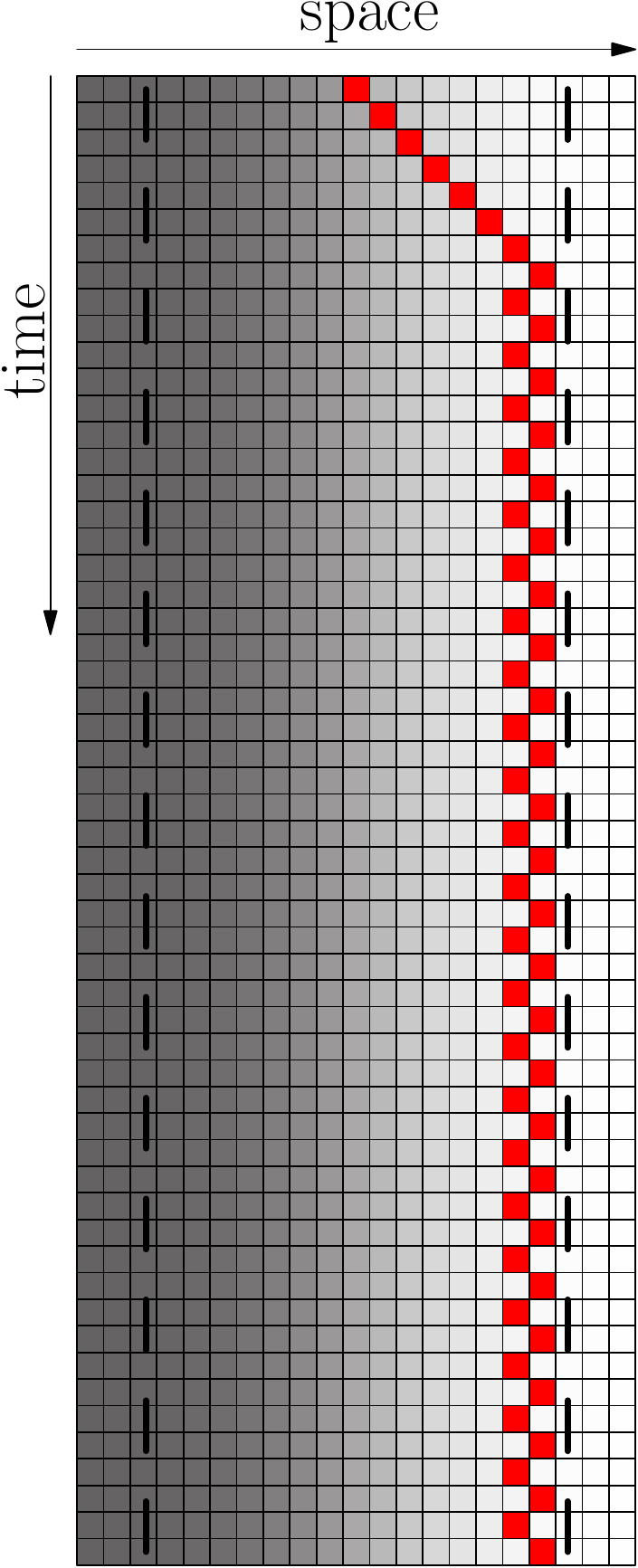}
\subcaption{ANN, environmental setting~1}\label{fig:results:cum:c}
\end{minipage}
\begin{minipage}[b]{.24\linewidth}
\includegraphics[width=1.0\textwidth]{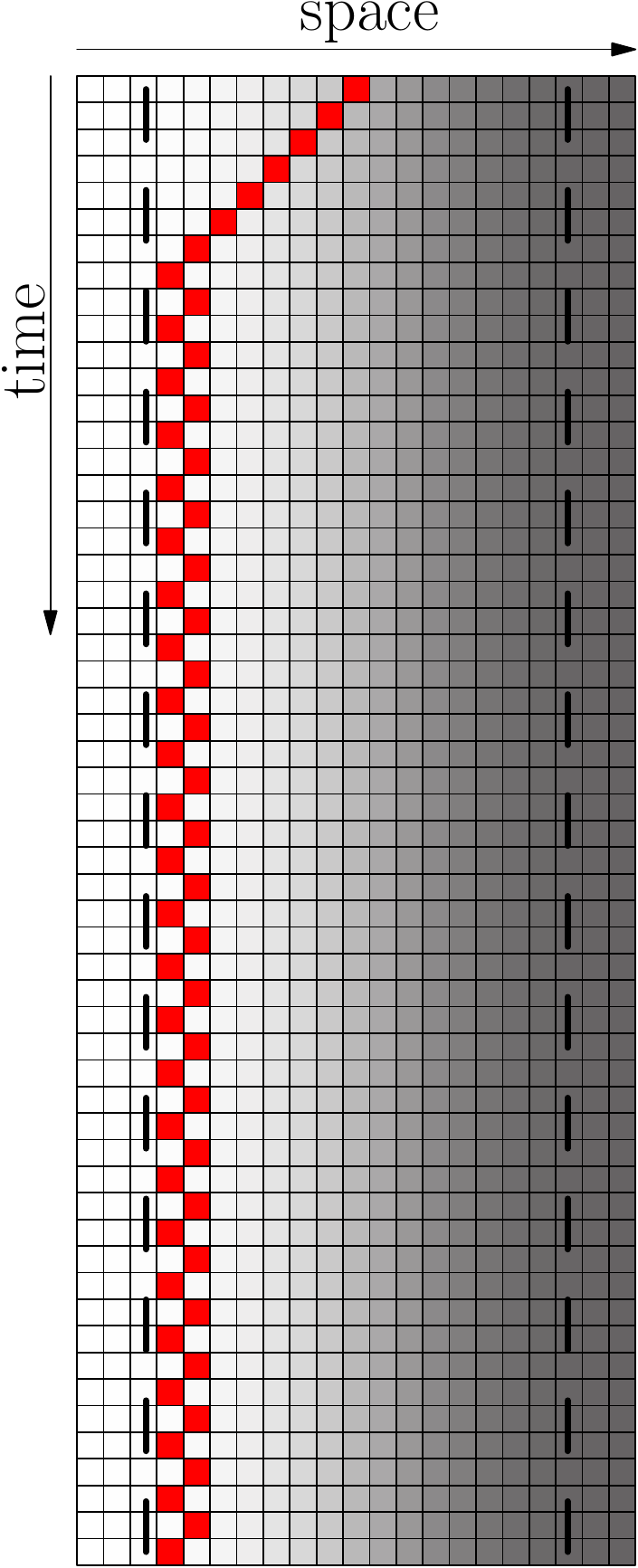}
\subcaption{ANN, environmental setting~2}\label{fig:results:cum:d}
\end{minipage}
\begin{minipage}[b]{.24\linewidth}
\includegraphics[width=1.0\textwidth]{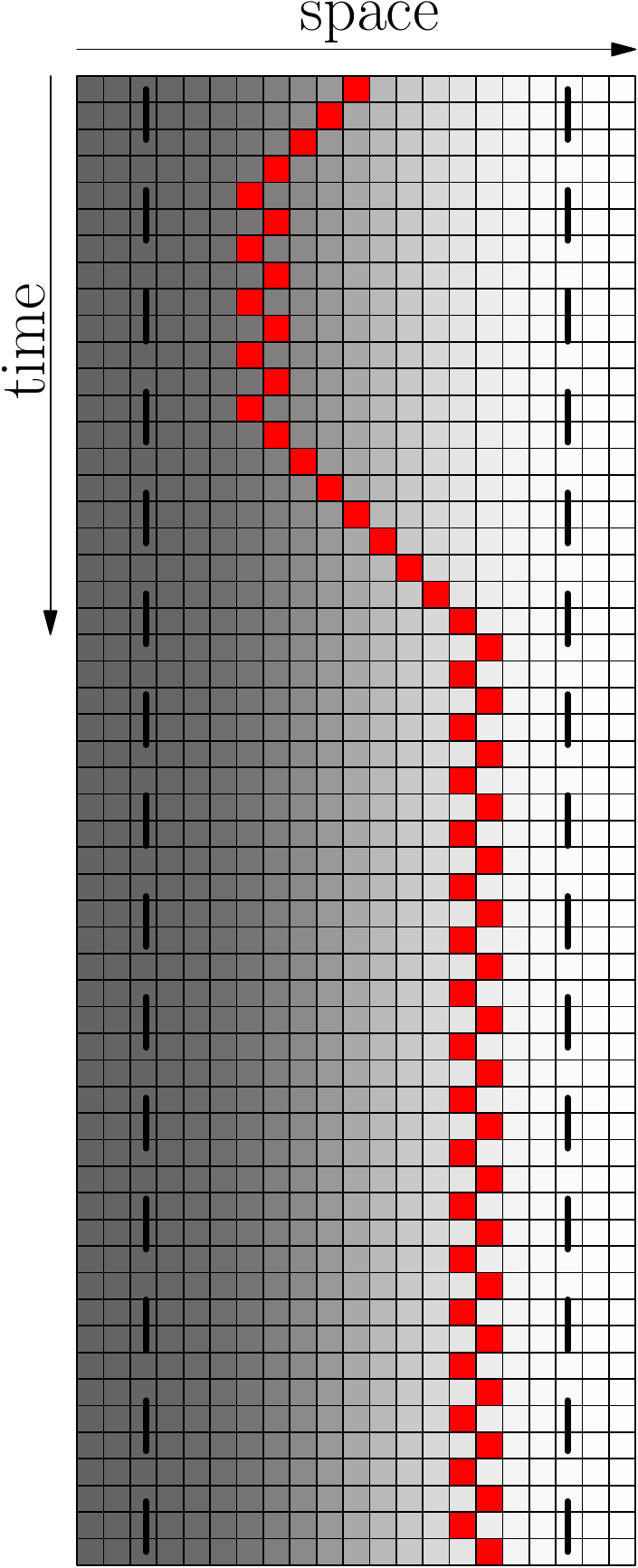}
\subcaption{CTRNN, environmental setting~1}\label{fig:results:cum:e}
\end{minipage}
\begin{minipage}[b]{.24\linewidth}
\includegraphics[width=1.0\textwidth]{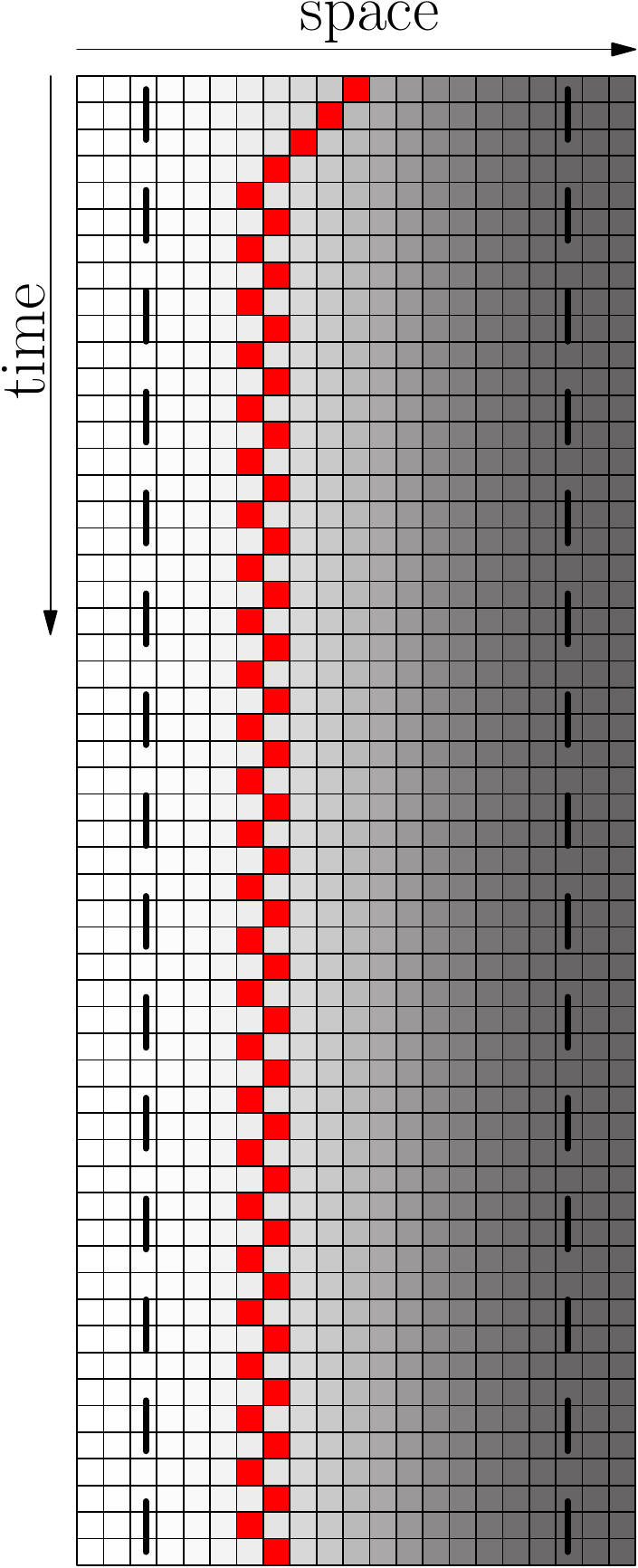}
\subcaption{CTRNN, environmental setting~2}\label{fig:results:cum:f}
\end{minipage}
\caption{Evolutionary results of the \emph{Cummulative} fitness
  regime: $\text{fitness} = F(0,1,0)$, the minimum of both evaluations
  was used as fitness value for each genome. Top row: fitness dynamics of
  ANNs (left) and CTRNNs (right). The horizontal orange line indicates
  the maximum fitness achievable by a reactive Wankelmut controller,
  as it is achieved by the simple hand-coded controller. Bottom row:
  trajectory of the best performing genome of both controller types in
  both environments used in evolution (only the initial 56 time steps of 250 time steps are shown).}\label{fig:results:cum}
\end{figure}

We found that the idea of cumulative fitness for every step was
exceptionally bad. To our surprise both controller types managed to
even outperform significantly the hand-coded ``optimal'' controller
(see Fig.~\ref{fig:results:cum:a} and
Fig.~\ref{fig:results:cum:b}). Looking at the trajectories revealed
that the evolutionary algorithm had found a ``cheap trick'' to
maximize this purely cumulative fitness function without evolving the
desired task: The controllers approached the threshold areas, thus gained
the maximum reward without having to switch the behavior.  They kept
this place for the rest of the run. This cheap trick was found for
both controllers, ANNs (see Fig.~\ref{fig:results:cum:c} and
Fig.~\ref{fig:results:cum:d}) and also by CTRNNs (see
Fig.~\ref{fig:results:cum:e} and Fig.~\ref{fig:results:cum:f}) in a less effective way. 
In both cases they actively avoided to behave like the desired
reactive Wankelmut controller by avoiding to cross the
threshold places.

\subsubsection{\emph{Instant+Switch} fitness regime}

In order to prevent the ``cheap trick'' deadlock for evolution we
discovered in the previous section, we again revised our fitness
function to the \emph{Instant+Switch} reward setting: We again rewarded
for every correct switch the agent performed and, instead of a cumulative
reward every time step, we rewarded the final (one instant) position of
the agent. This way, bootstrapping problems could also be mediated as
again movement in the right direction without reaching the threshold
place for switching the behavior would be rewarded but it did not pay
out to keep the position just before triggering the switch.The fitness function 
used here can be described as $\text{fitness} = F(100,0,0.01)$.

\begin{figure}[h!]
\vspace*{1.5cm}
\begin{minipage}[b]{.45\linewidth}
\includegraphics[width=1.0\textwidth]{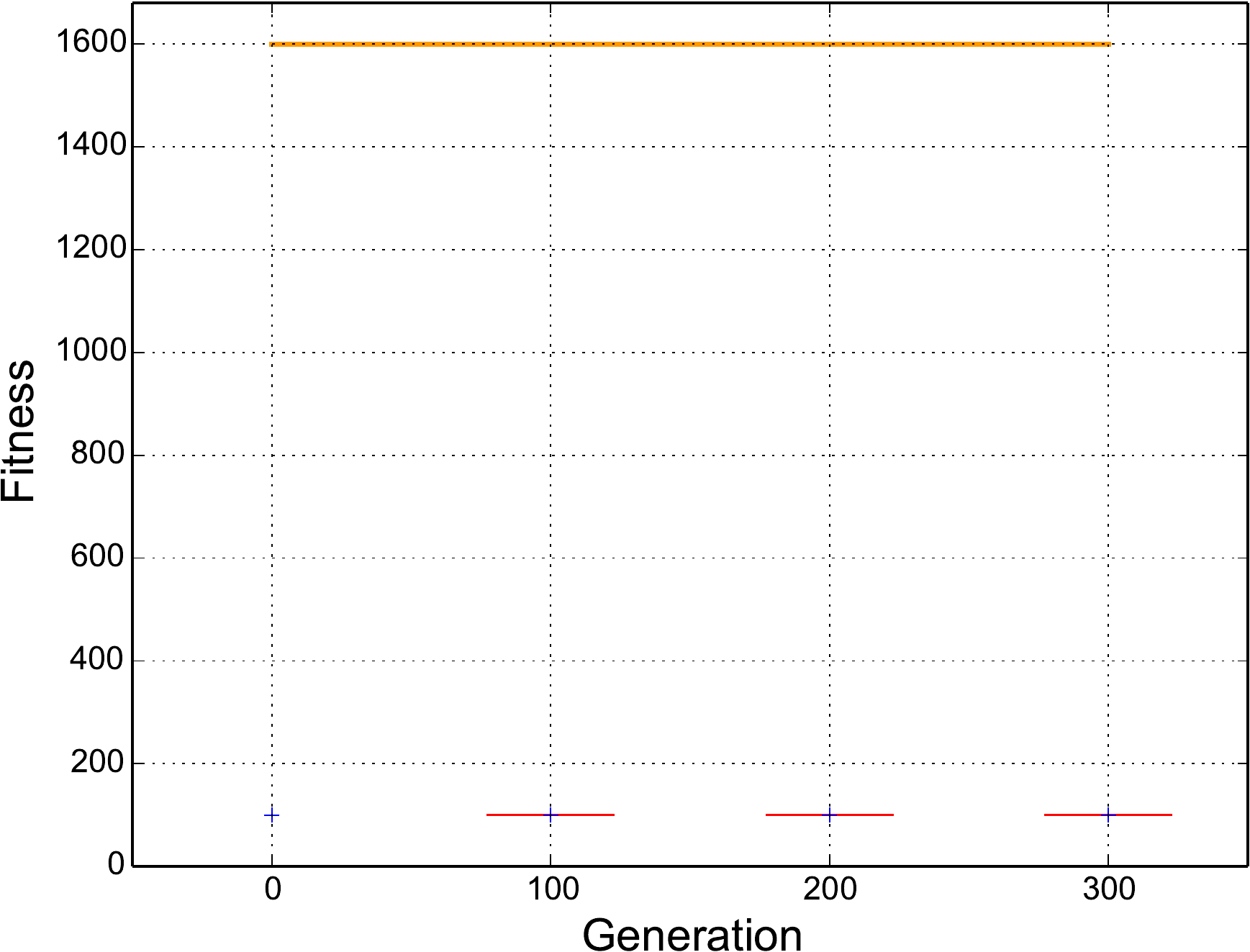}
\subcaption{ANN, fitness}\label{fig:results:sw_ins:a}
\end{minipage}%
\hspace*{.05\linewidth}%
\begin{minipage}[b]{.45\linewidth}
\includegraphics[width=1.0\textwidth]{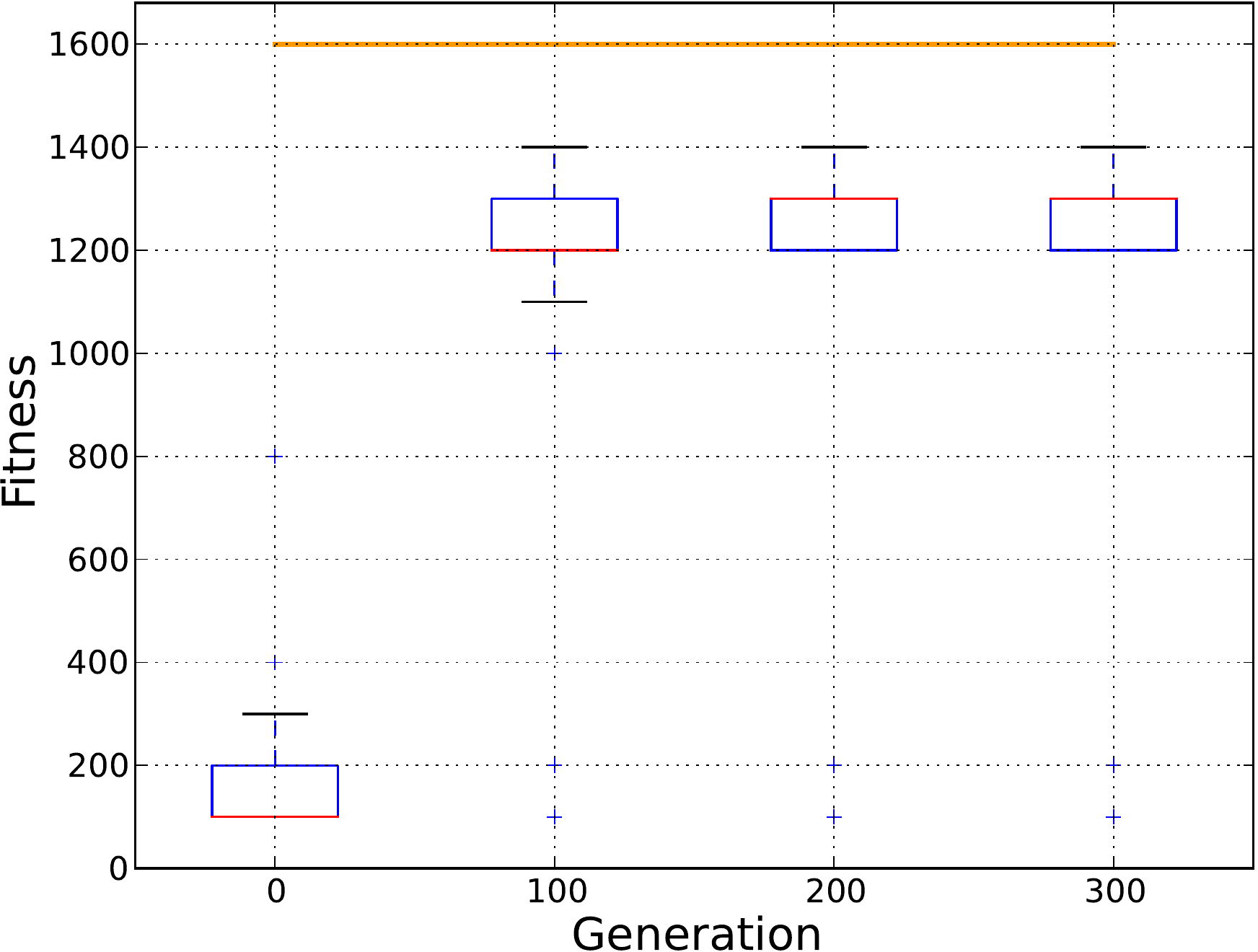}
\subcaption{CTRNN, fitness}\label{fig:results:sw_ins:b}
\end{minipage}
\begin{minipage}[b]{.24\linewidth}
\includegraphics[width=1.0\textwidth]{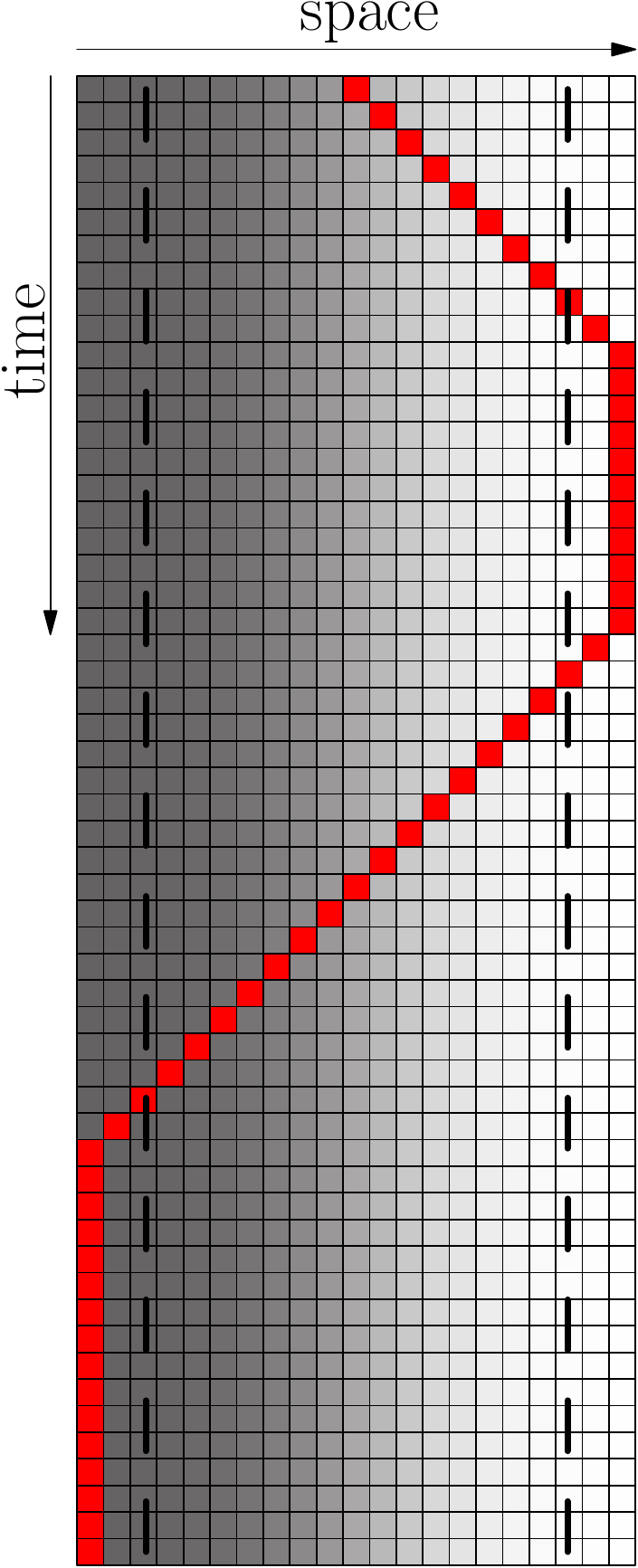}
\subcaption{ANN, environmental setting~1}\label{fig:results:sw_ins:c}
\end{minipage}
\begin{minipage}[b]{.24\linewidth}
\includegraphics[width=1.0\textwidth]{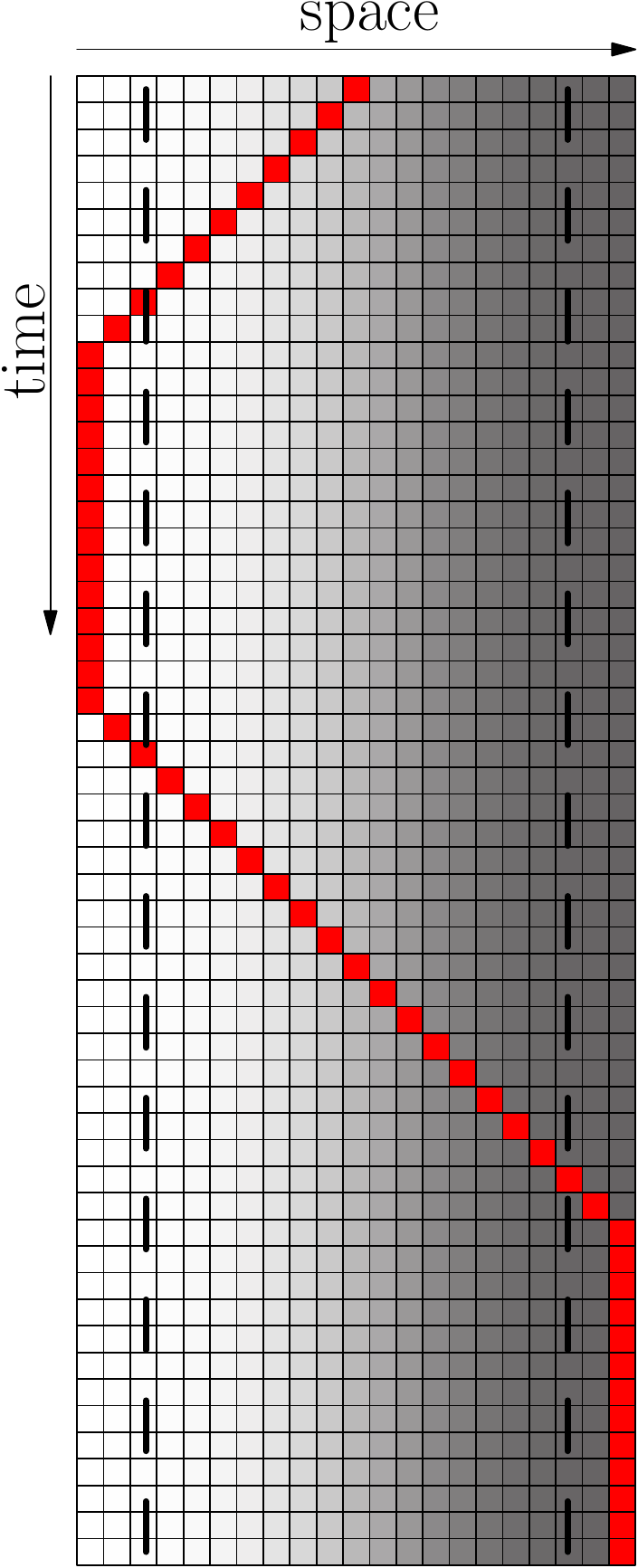}
\subcaption{ANN, environmental setting~2}\label{fig:results:sw_ins:d}
\end{minipage}
\begin{minipage}[b]{.24\linewidth}
\includegraphics[width=1.0\textwidth]{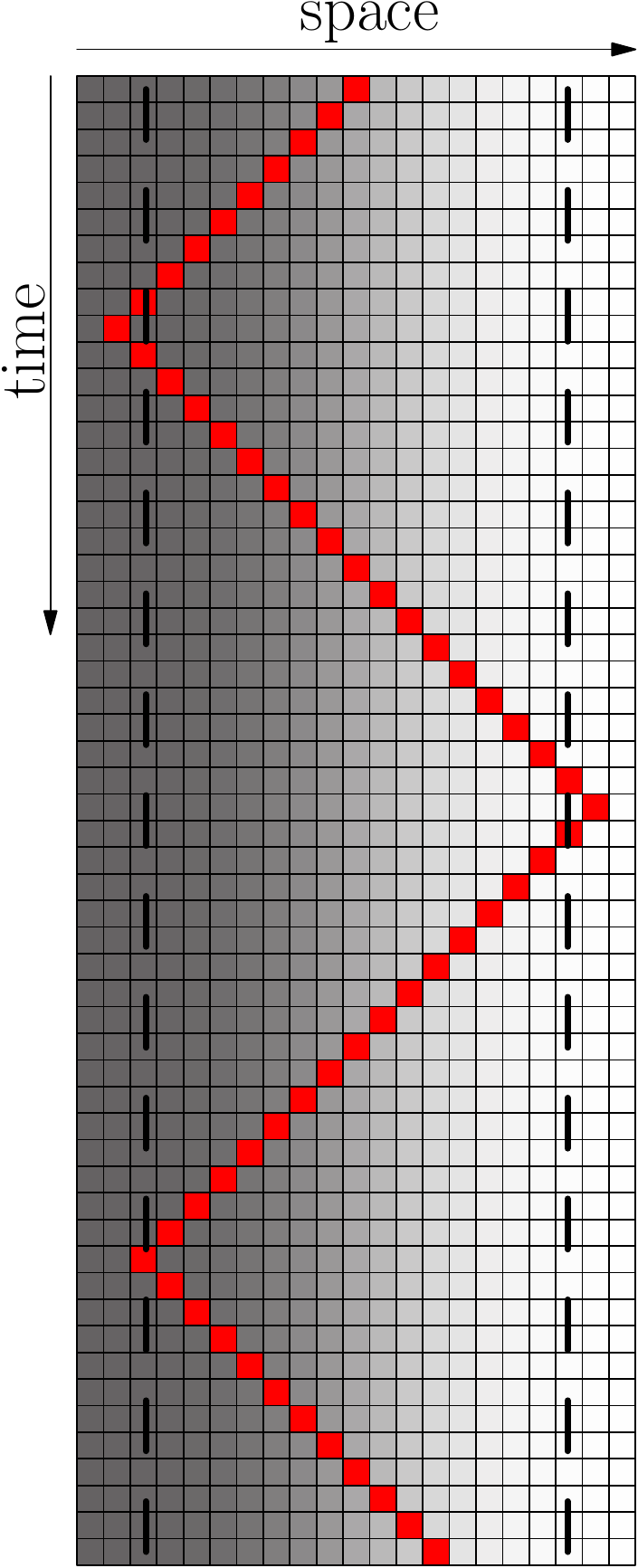}
\subcaption{CTRNN, environmental setting~1}\label{fig:results:sw_ins:e}
\end{minipage}
\begin{minipage}[b]{.24\linewidth}
\includegraphics[width=1.0\textwidth]{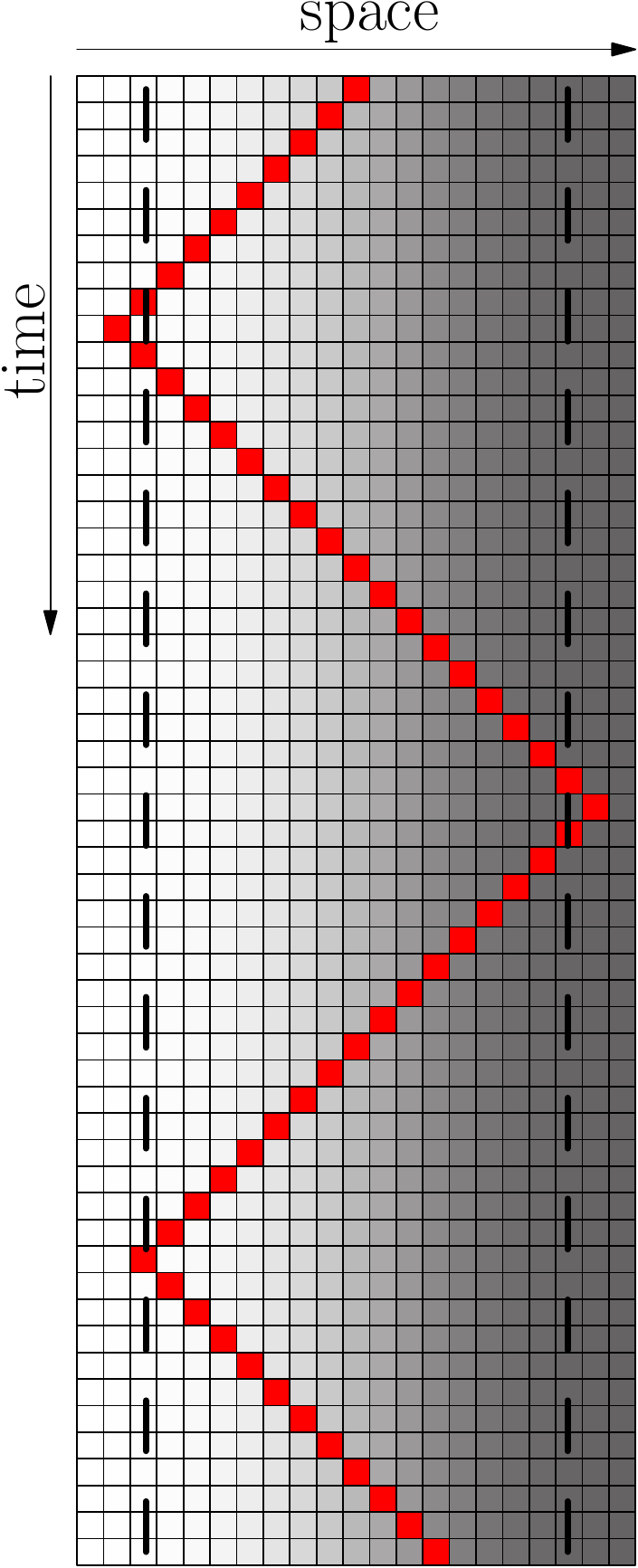}
\subcaption{CTRNN, environmental setting~2}\label{fig:results:sw_ins:f}
\end{minipage}
\caption{Evolutionary results of the \emph{Instant+Switch} fitness
  regime: $\text{fitness} = F(100,0,0.01)$, the minimum of both
  evaluations was used as fitness value for each genome. Top row: fitness
  dynamics of ANNs (left) and CTRNNs (right). The horizontal orange
  line indicates the maximum fitness achievable by a reactive
  Wankelmut controller, as it is achieved by the simple hand-coded
  controller. Bottom row: trajectory of the best performing genome of
  both controller types in both environments used in
  evolution (only the initial 56 time steps of 250 time steps are shown).}\label{fig:results:sw_ins}
\end{figure}

The highest fitness values in this setting were achieved by a CTRNNs (see
Fig.~\ref{fig:results:sw_ins:b}). However this was again achieved
 by just quickly zig-zagging across the world without reacting to
the environmental situation: The agents started off into the
wrong direction (see Fig.~\ref{fig:results:sw_ins:e} and
Fig.~\ref{fig:results:sw_ins:f}) in one environmental setting. 
Evolved ANNs did not at all develop highly rewarded behaviors in this 
setting (see Fig.~\ref{fig:results:sw_ins:a}). However the best ANN genome 
evolved in principle parts of the desired behaviors for the first period of time:
They started off in both environments into the correct direction and executed
the behavioral switch one time but never closed the cycle back again 
(see Fig.~\ref{fig:results:sw_ins:c} and
Fig.~\ref{fig:results:sw_ins:d}). Overall, the desired reactive
Wankelmut behavior was never fully evolved in neither one of the controller
types.

\subsubsection{\emph{Cumulative+Switch} fitness regime}

In a final attempt to achieve the desired behavior we also tested the
\emph{Cumulative+Switch} fitness regime where the controller was
rewarded cumulatively for positioning of the agent over time as well
as significant rewards for the correct switches were given.  The fitness 
function can be described as $\text{fitness} = F(100,0.01,0)$.

\begin{figure}[h!]
\vspace*{1.5cm}
\begin{minipage}[b]{.45\linewidth}
\includegraphics[width=1.0\textwidth]{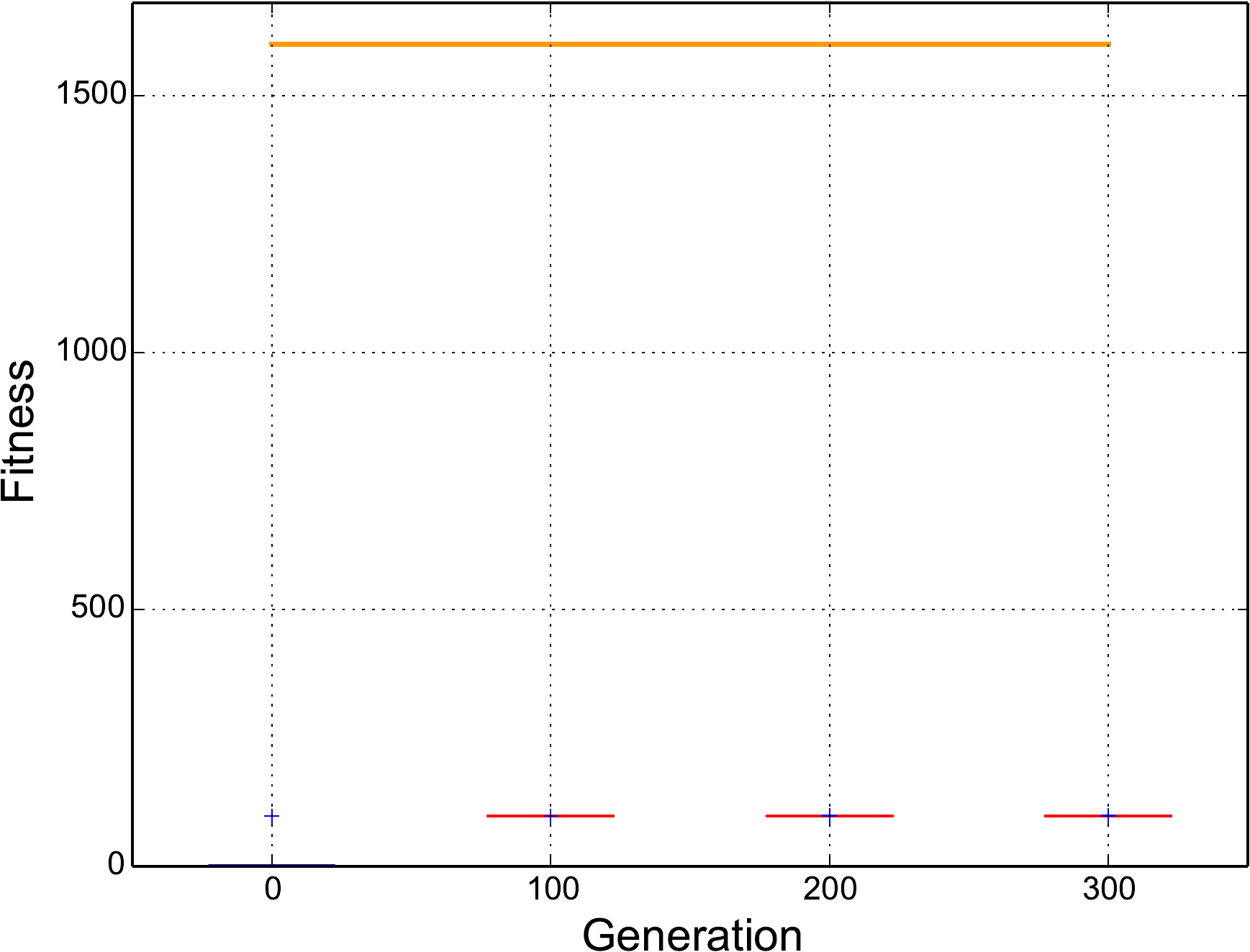}
\subcaption{ANN, fitness}\label{fig:results:sw_cum:a}
\end{minipage}%
\hspace*{.05\linewidth}%
\begin{minipage}[b]{.45\linewidth}
\includegraphics[width=1.0\textwidth]{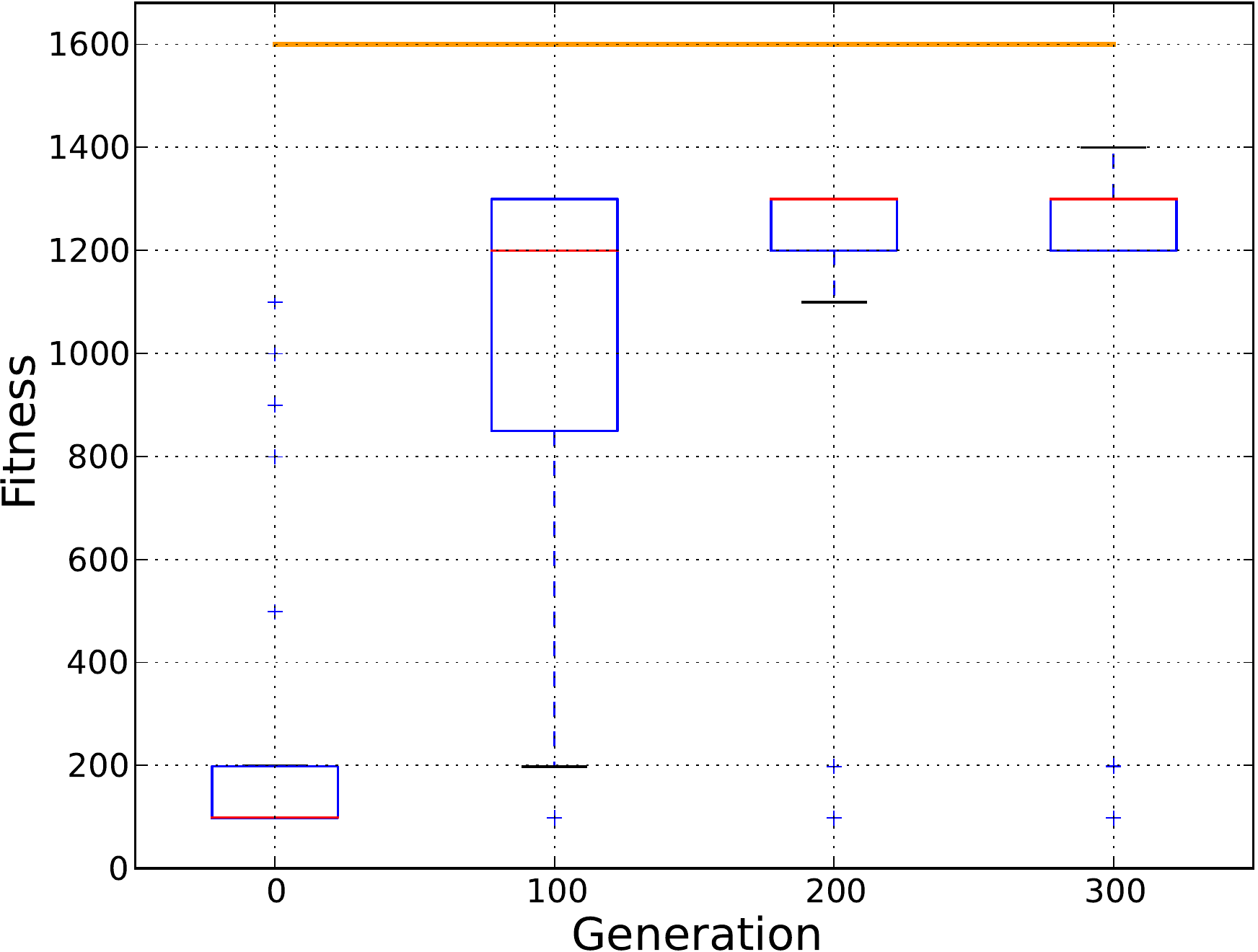}
\subcaption{CTRNN, fitness}\label{fig:results:sw_cum:b}
\end{minipage}
\begin{minipage}[b]{.24\linewidth}
\includegraphics[width=1.0\textwidth]{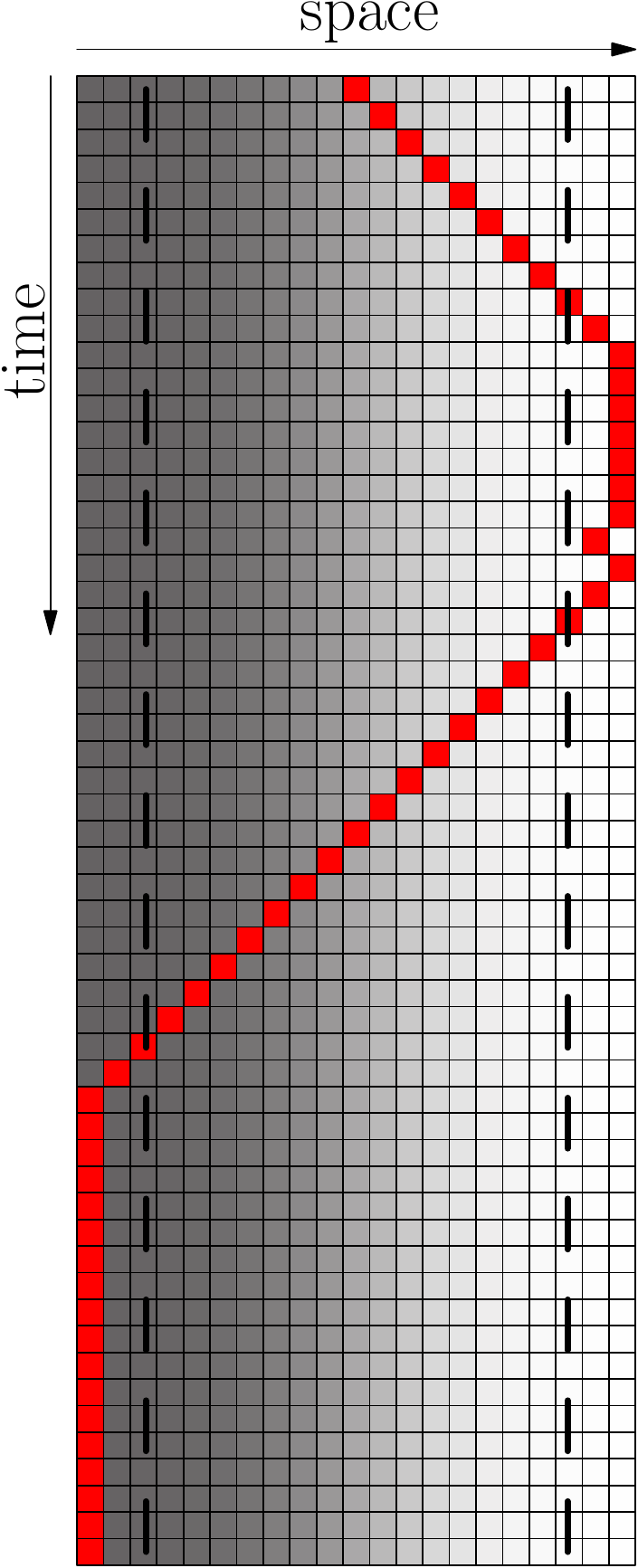}
\subcaption{ANN, environmental setting~1}\label{fig:results:sw_cum:c}
\end{minipage}
\begin{minipage}[b]{.24\linewidth}
\includegraphics[width=1.0\textwidth]{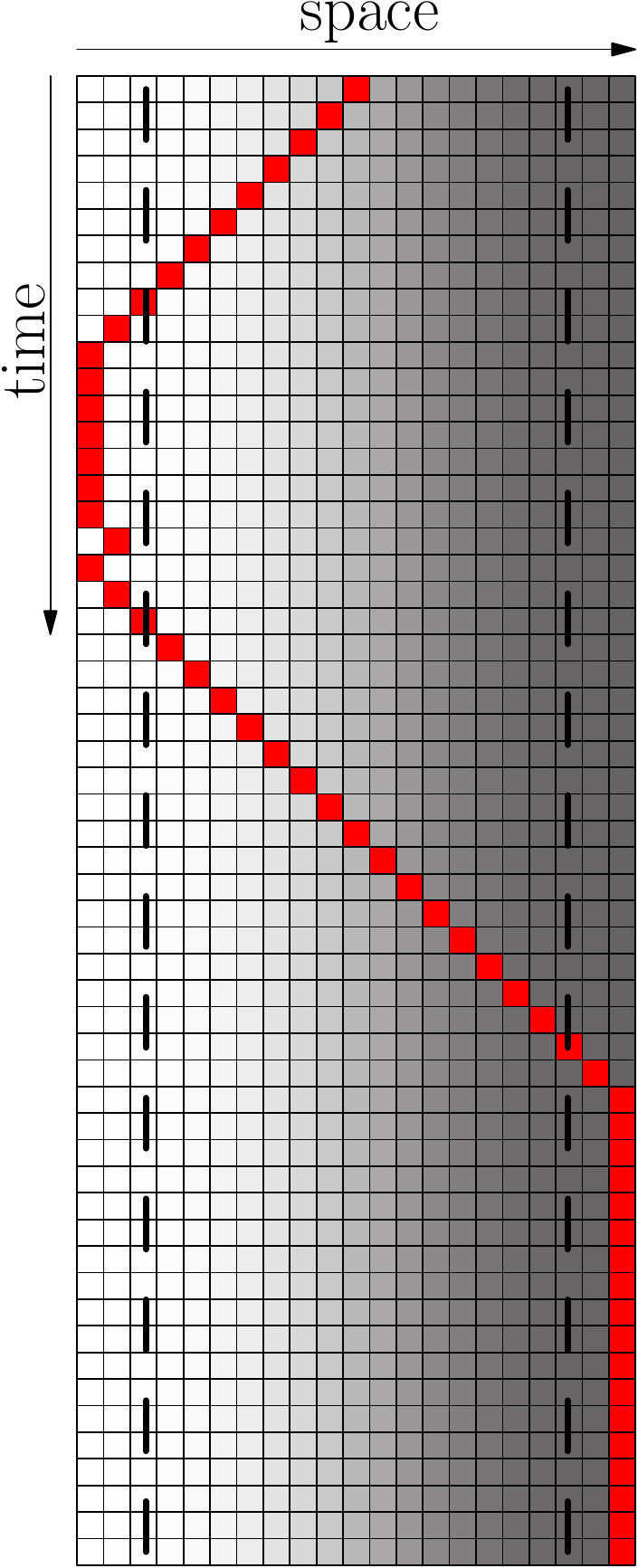}
\subcaption{ANN, environmental setting~2}\label{fig:results:sw_cum:d}
\end{minipage}
\begin{minipage}[b]{.24\linewidth}
\includegraphics[width=1.0\textwidth]{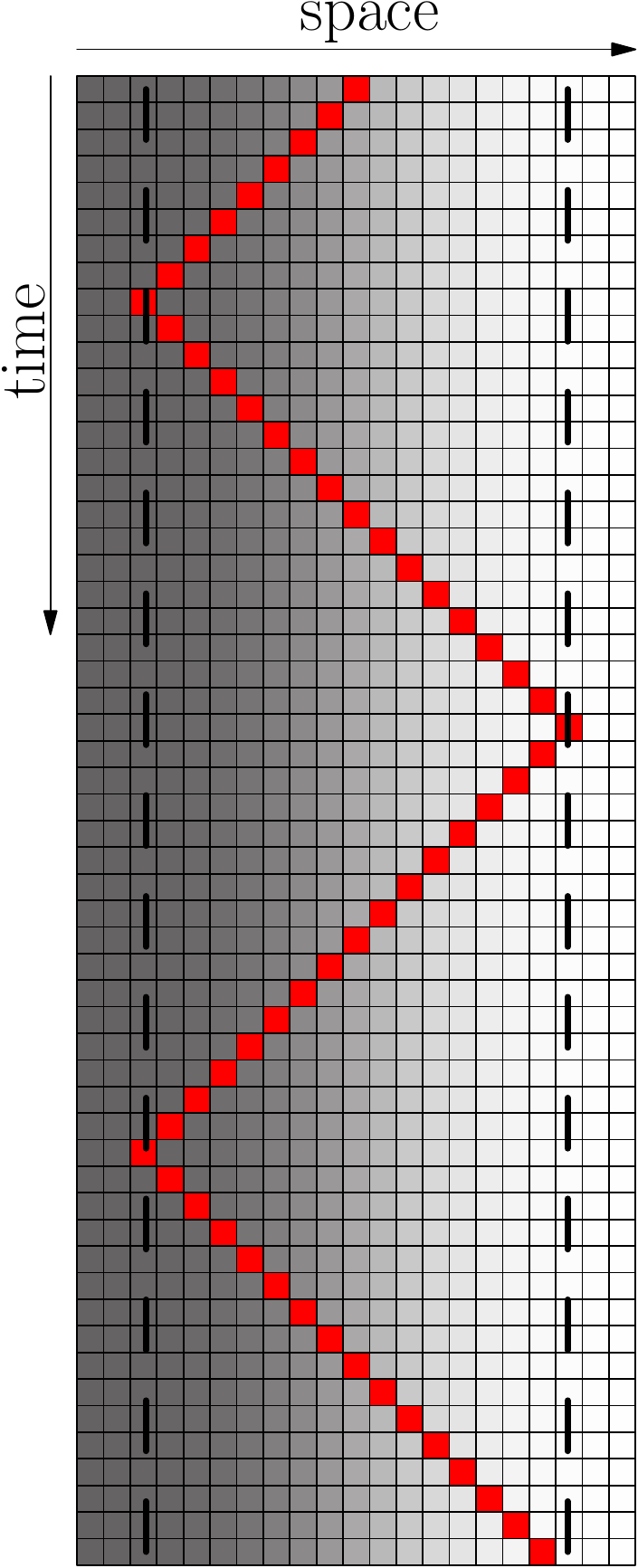}
\subcaption{CTRNN, environmental setting~1}\label{fig:results:sw_cum:e}
\end{minipage}
\begin{minipage}[b]{.24\linewidth}
\includegraphics[width=1.0\textwidth]{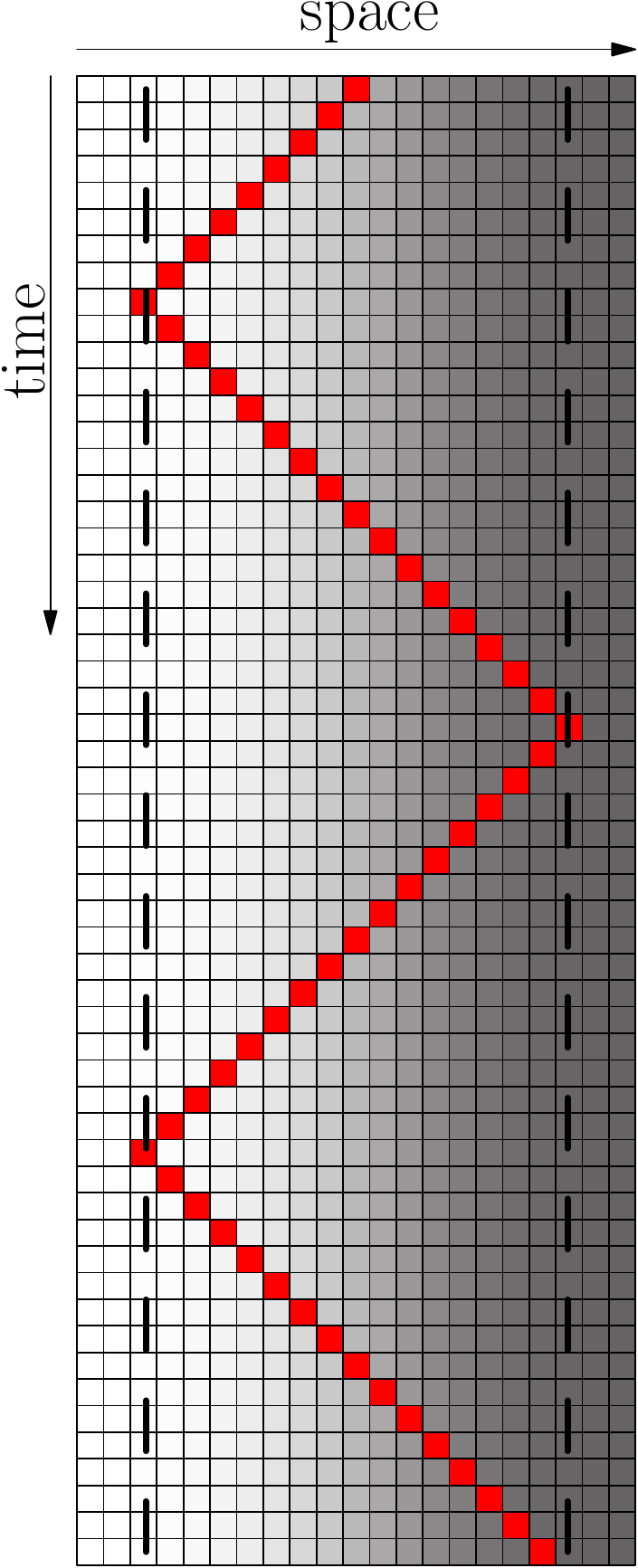}
\subcaption{CTRNN, environmental setting~2}\label{fig:results:sw_cum:f}
\end{minipage}
\caption{Evolutionary results of the \emph{Cummulative+Switch} fitness
  regime: $\text{fitness} = F(100,0.01,0)$, the minimum of both
  evaluations was used as fitness value for each genome. Top row: fitness
  dynamics of ANNs (left) and CTRNNs (right). The horizontal orange
  line indicates the maximum fitness achievable by a reactive
  Wankelmut controller, as it is achieved by the simple hand-coded
  controller. Bottom row: trajectory of the best performing genome of
  both controller types in both environments used in
  evolution (only the initial 56 time steps of 250 time steps are shown).}\label{fig:results:sw_cum}
\end{figure}

Introducing the additional rewarding for correct switches prevented
the evolutionary algorithm from falling into the ``cheap trick'' local
optimum it did when it was only rewarded cumulatively per
step. The results were almost similar to the ones obtained from the
previous \emph{Instant+Switch} regime: CTRNNs did not evolve anything 
close to the desired reactive Wankelmut behavior (see Fig.~\ref{fig:results:sw_cum})
and ANNs evolved again the first switching of behaviors but failed
to evolve the switch back. (see
Fig.~\ref{fig:results:sw_cum:c} and Fig.~\ref{fig:results:sw_cum:d}). 
CTRNNs achieved again some significant reward by oscillating in both environments 
in a non-reactive oscillatory behavior (see Fig.~\ref{fig:results:sw_cum:e}
and Fig.~\ref{fig:results:sw_cum:f}), like they have evolved already in most of
the other evolution regimes.

\begin{figure}[h!]
\vspace*{1.5cm}
\begin{minipage}[b]{.45\linewidth}
\includegraphics[width=1.0\textwidth]{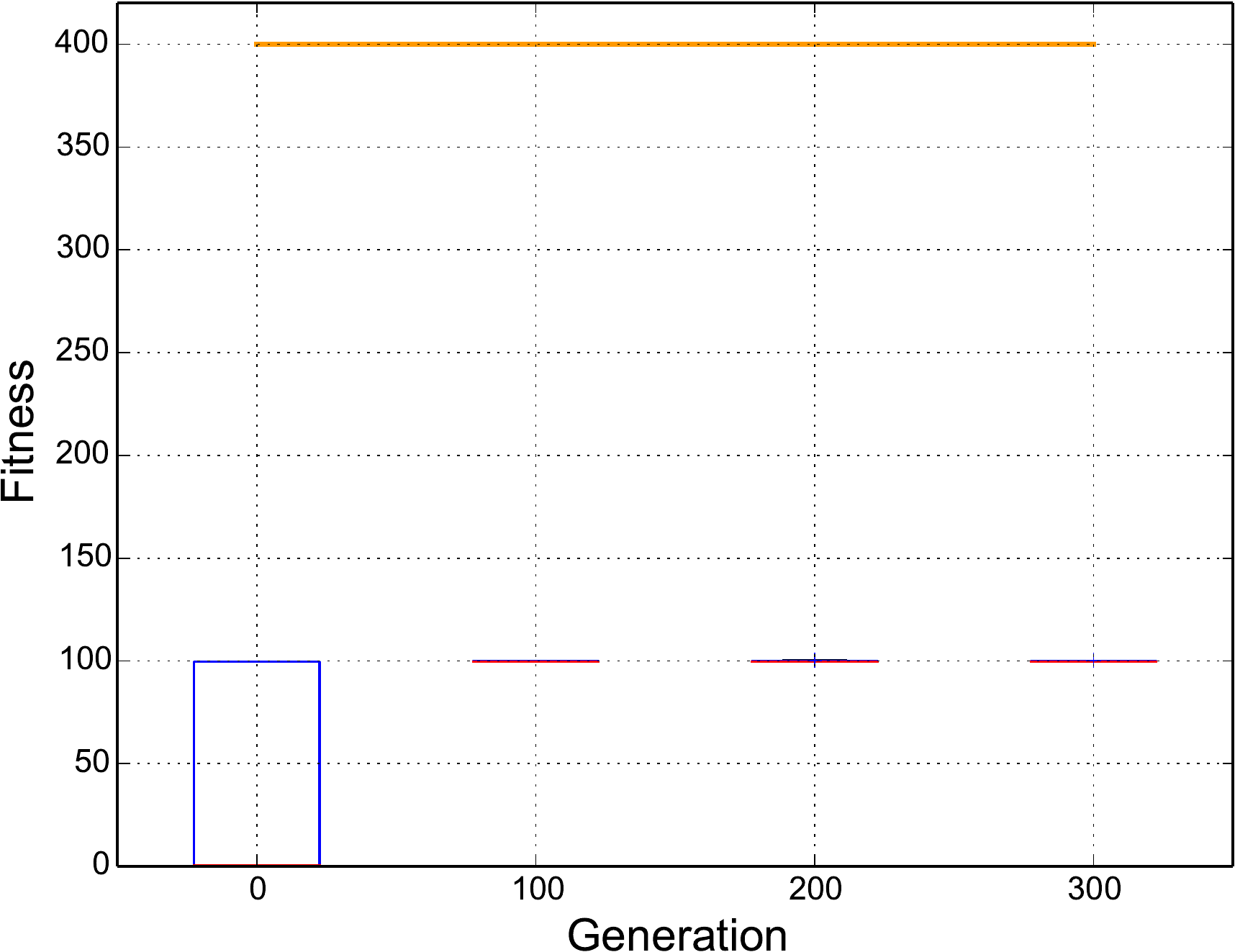}
\subcaption{ANN, fitness}\label{fig:results:sw_cum_56:a}
\end{minipage}%
\hspace*{.05\linewidth}%
\begin{minipage}[b]{.45\linewidth}
\includegraphics[width=1.0\textwidth]{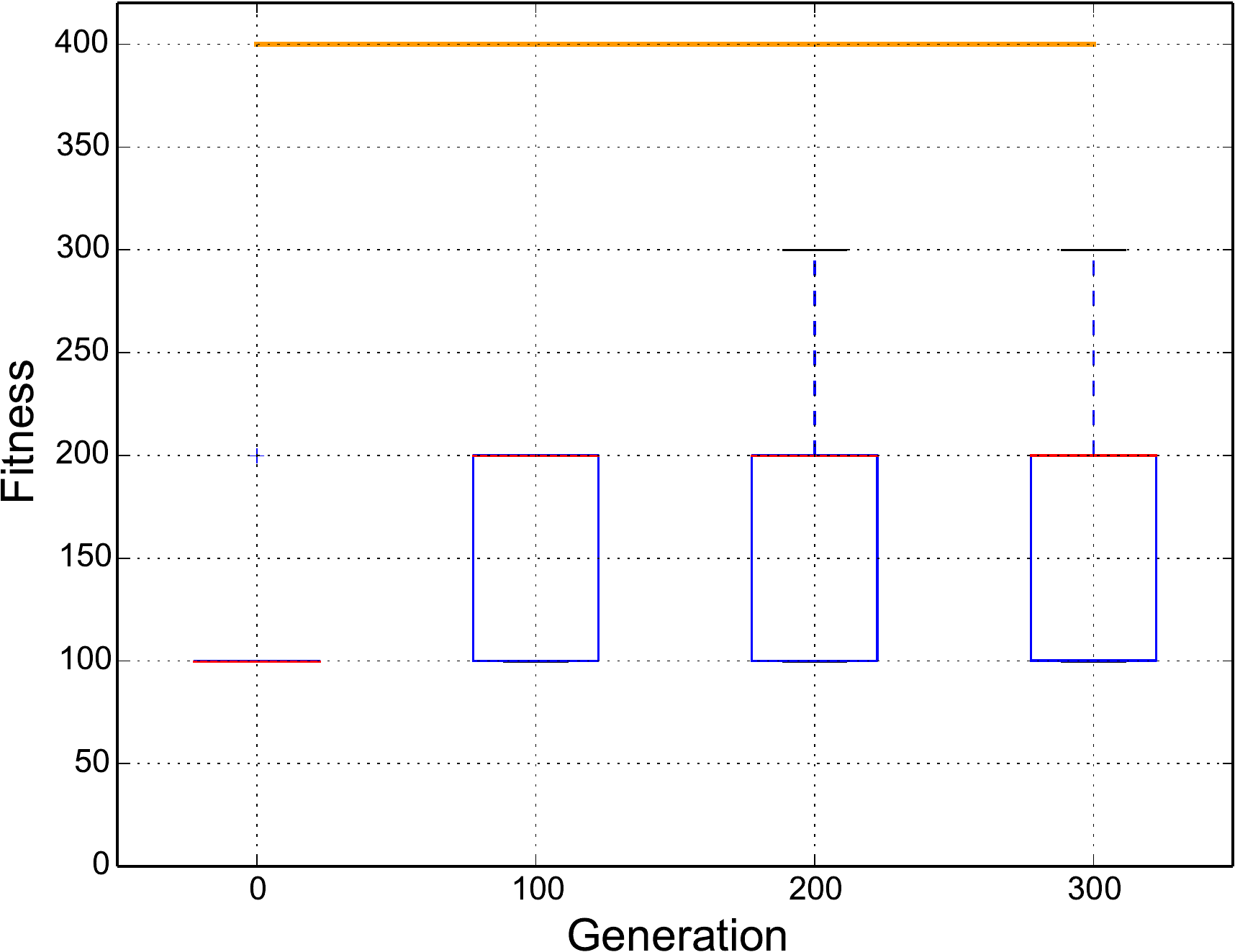}
\subcaption{CTRNN, fitness}\label{fig:results:sw_cum_56:b}
\end{minipage}
\begin{minipage}[b]{.24\linewidth}
\includegraphics[width=1.0\textwidth]{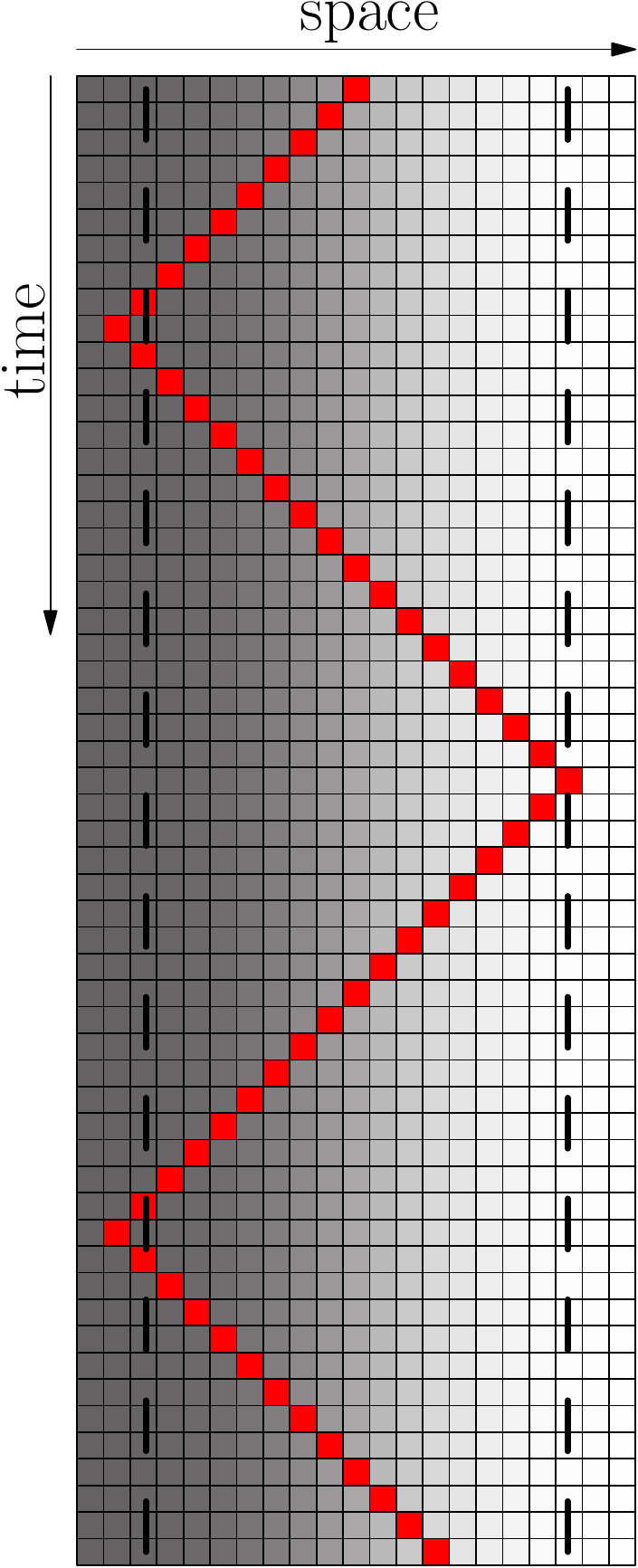}
\subcaption{ANN, environmental setting~1}\label{fig:results:sw_cum_56:c}
\end{minipage}
\begin{minipage}[b]{.24\linewidth}
\includegraphics[width=1.0\textwidth]{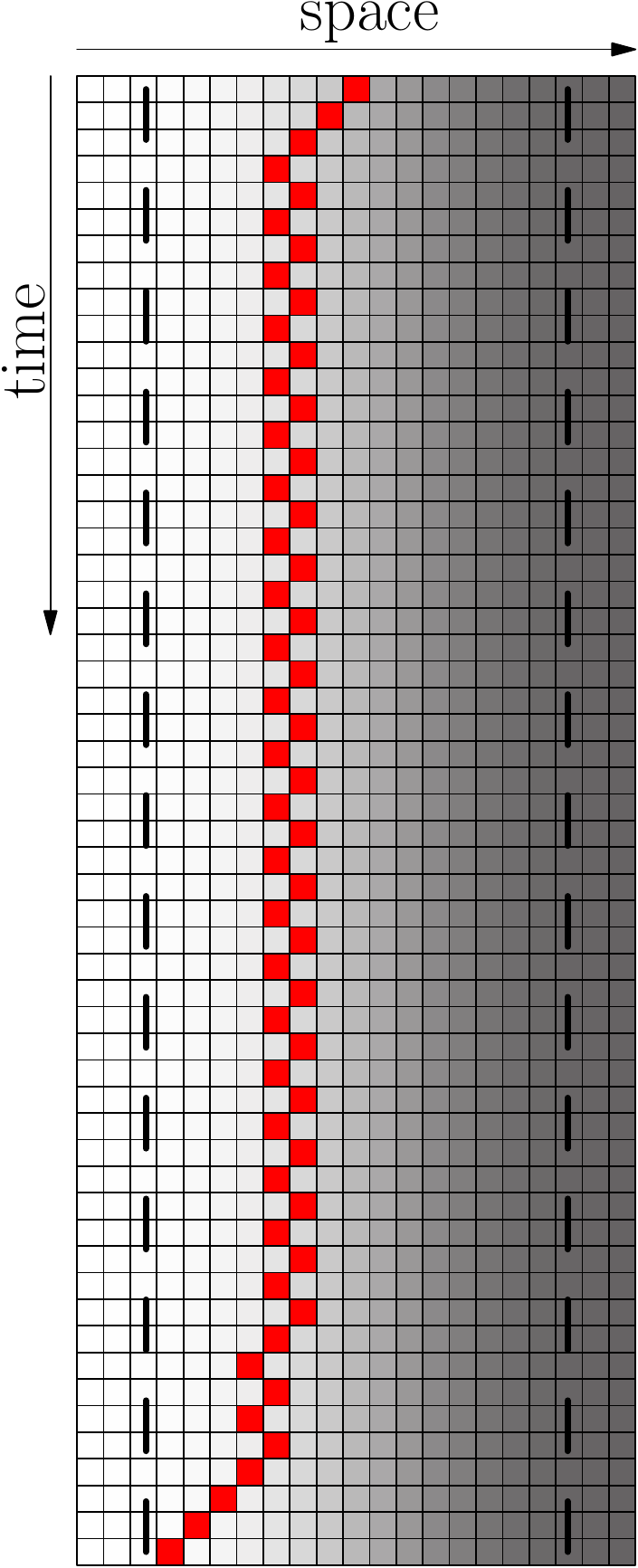}
\subcaption{ANN, environmental setting~2}\label{fig:results:sw_cum_56:d}
\end{minipage}
\begin{minipage}[b]{.24\linewidth}
\includegraphics[width=1.0\textwidth]{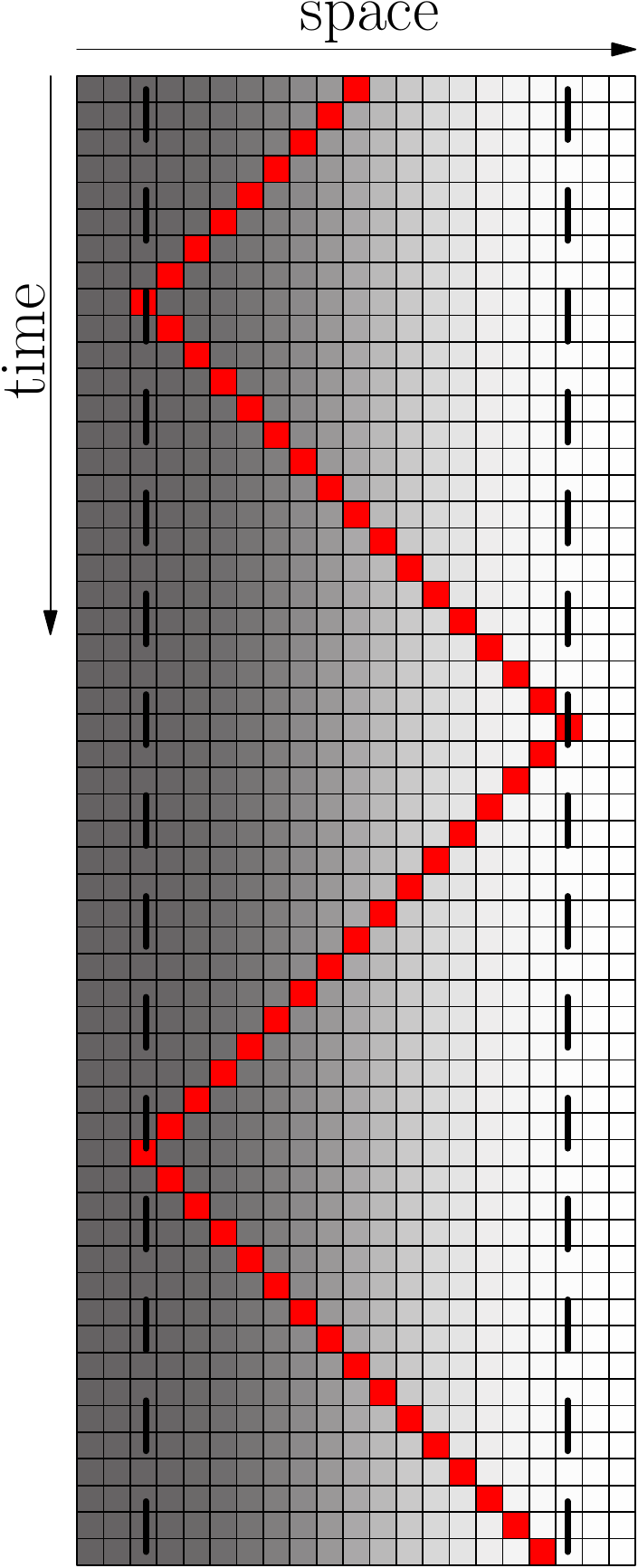}
\subcaption{CTRNN, environmental setting~1}\label{fig:results:sw_cum_56:e}
\end{minipage}
\begin{minipage}[b]{.24\linewidth}
\includegraphics[width=1.0\textwidth]{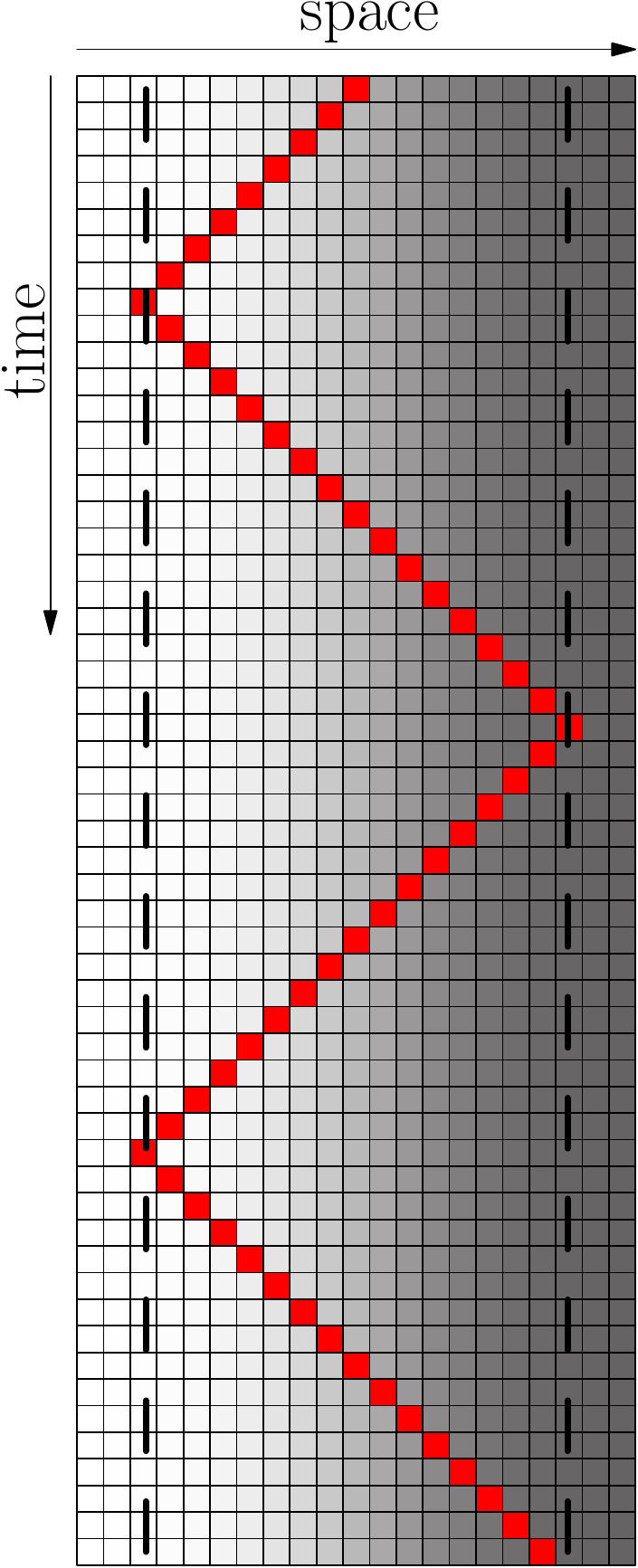}
\subcaption{CTRNN, environmental setting~2}\label{fig:results:sw_cum_56:f}
\end{minipage}
\caption{Evolutionary results of short evaluations with only 56~time
  steps, the \emph{Cummulative+Switch} fitness regime: $\text{fitness}
  = F(100,0.01,0)$, the minimum of both evaluations was used as
  fitness value for each genome. Top row: fitness dynamics of ANNs (left)
  and CTRNNs (right). The horizontal orange line indicates the maximum
  fitness achievable by a reactive Wankelmut controller, as it is
  achieved by the simple hand-coded controller. Bottom row: trajectory
  of the best performing genome of both controller types in both
  environments used in evolution (all 56 time steps are shown).}\label{fig:results:sw_cum_56}
\end{figure}

In another attempt to enforce reactivity of the evolved controller and
considering the evolved solutions shown above, we tried to push the
evolution towards more reactivity of the controllers. Thus, the
evolutionary experiment was repeated with only 56 time-steps per
evaluation (see Fig.~\ref{fig:results:sw_cum_56}). Again, CTRNNs
evolved a higher fitness (Fig.~\ref{fig:results:sw_cum_56:b}) than
ANNs (Fig.~\ref{fig:results:sw_cum_56:a}). However, this was still
achieved with non-reactive fast oscillations (see
Fig.~\ref{fig:results:sw_cum_56:e} and
Fig.~\ref{fig:results:sw_cum_56:f}). The ANN evolved into a behavior
that oscillated in one environment although starting in the wrong
direction initially (see Fig.~\ref{fig:results:sw_cum_56:c}. 
In the second environment, the best agent progressed very slowly towards the 
threshold place and no switch back was observed there
Fig.~\ref{fig:results:sw_cum_56:d}). 

The resulting controllers were
also tested in a post-evaluation to check for reactive behaviors. This
was done with a different environment that had maximum quality in the
middle and two minima, one at the left end and one at the right
end. Agents were tested in two evaluations with one initially
positioning the robot at the left end and the second one with
initially positioning the robot at the right end. While some
controllers managed to operate reasonably in one of the two evaluations
(oscillating between the initial position and the middle place), none of
them did so for both starting places. Hence, we conclude that also in this
experiment we did not see the correct controller that solves the
Wankelmut task in a reactive way.

\subsection{Evolving a Hand-coded Network}
\begin{figure}[h!]
\vspace*{2cm}
\begin{minipage}[b]{.49\linewidth}
\includegraphics[width=0.9\textwidth]{handcoded}
\subcaption{Schematic drawing of the hand-coded ANN}\label{fig:evolvehandcoded:base}
\end{minipage}%
\hspace*{.05\linewidth}%
\begin{minipage}[b]{.18\linewidth}
\includegraphics[width=1.0\textwidth]{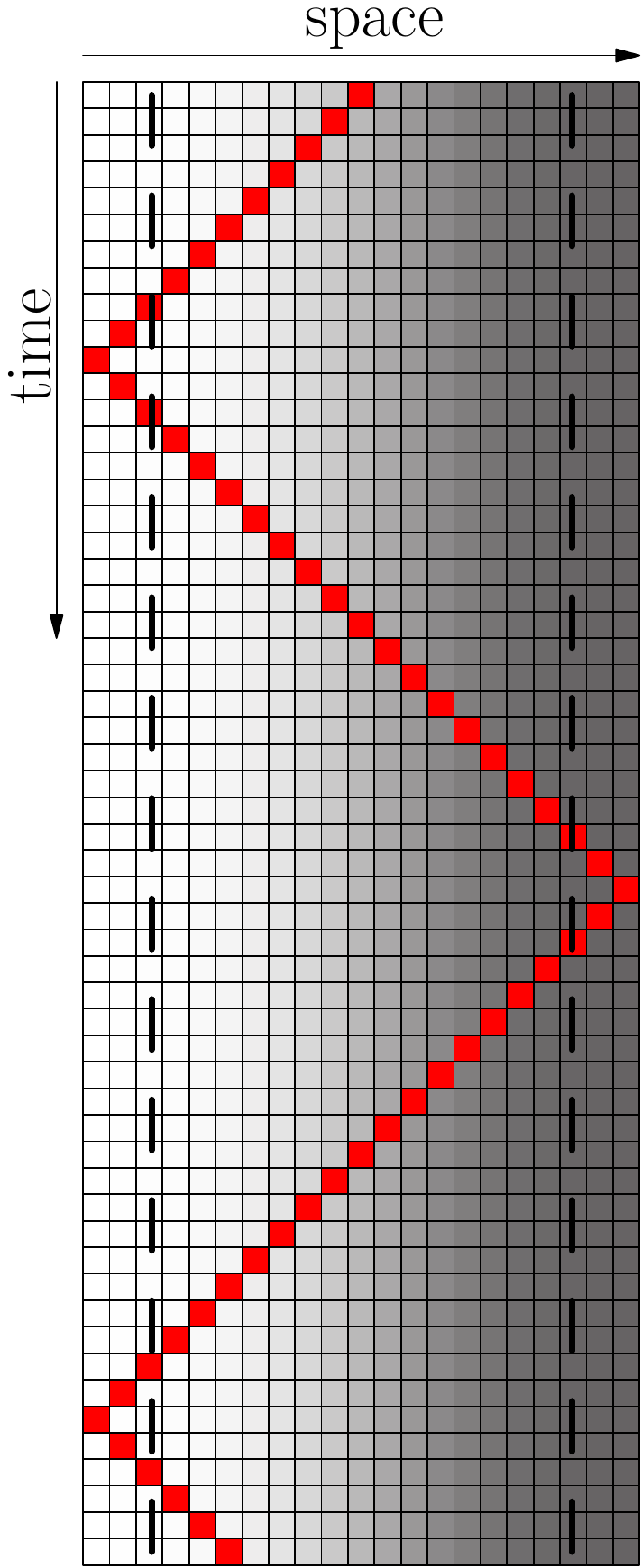}
\subcaption{Example hand-coded ANN, environmental setting~1}\label{fig:evolvehandcoded:hct_st1}
\end{minipage}
\hspace*{.01\linewidth}%
\begin{minipage}[b]{.18\linewidth}
\includegraphics[width=1.0\textwidth]{cellular_side2_handcoded-eps-converted-to.pdf}
\subcaption{Example hand-coded ANN, environmental setting~2}\label{fig:evolvehandcoded:hct_st2}
\end{minipage}\\ 
\\
\begin{minipage}[b]{.49\linewidth}
\includegraphics[width=0.9\textwidth]{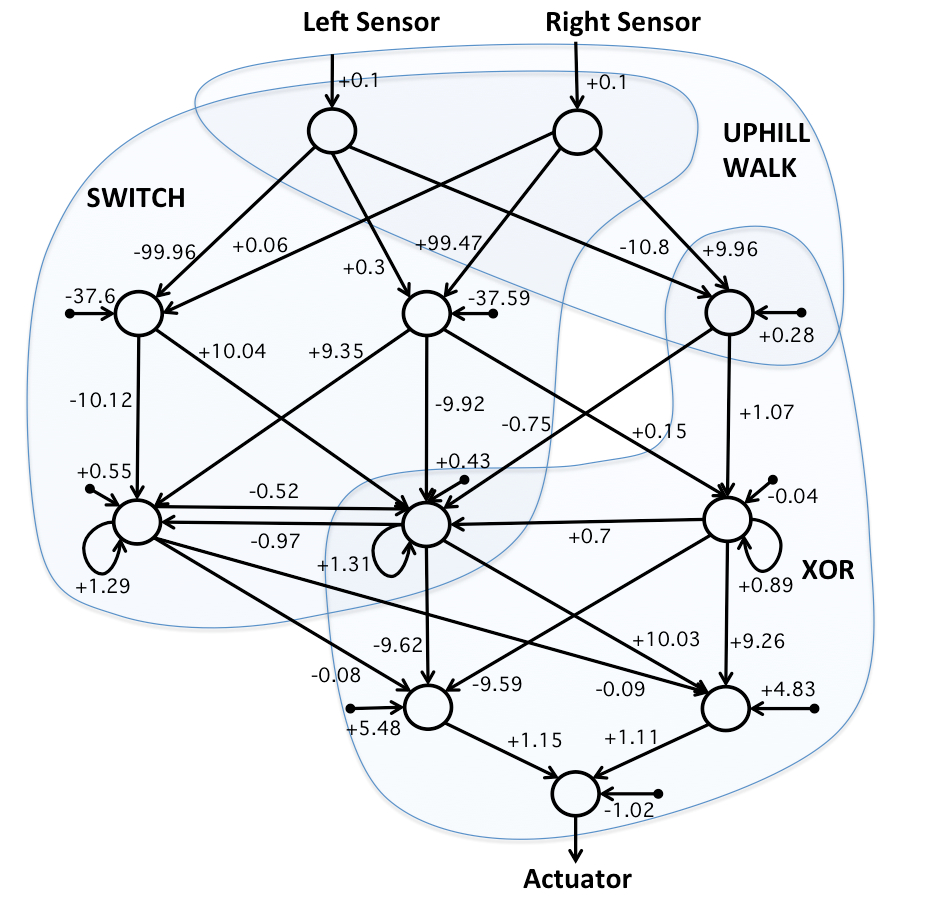}
\subcaption{Schematic drawing of the evolved hand-coded ANN}\label{fig:evolvehandcoded:evolved}
\end{minipage}
\hspace*{.05\linewidth}%
\begin{minipage}[b]{.18\linewidth}
\includegraphics[width=1.0\textwidth]{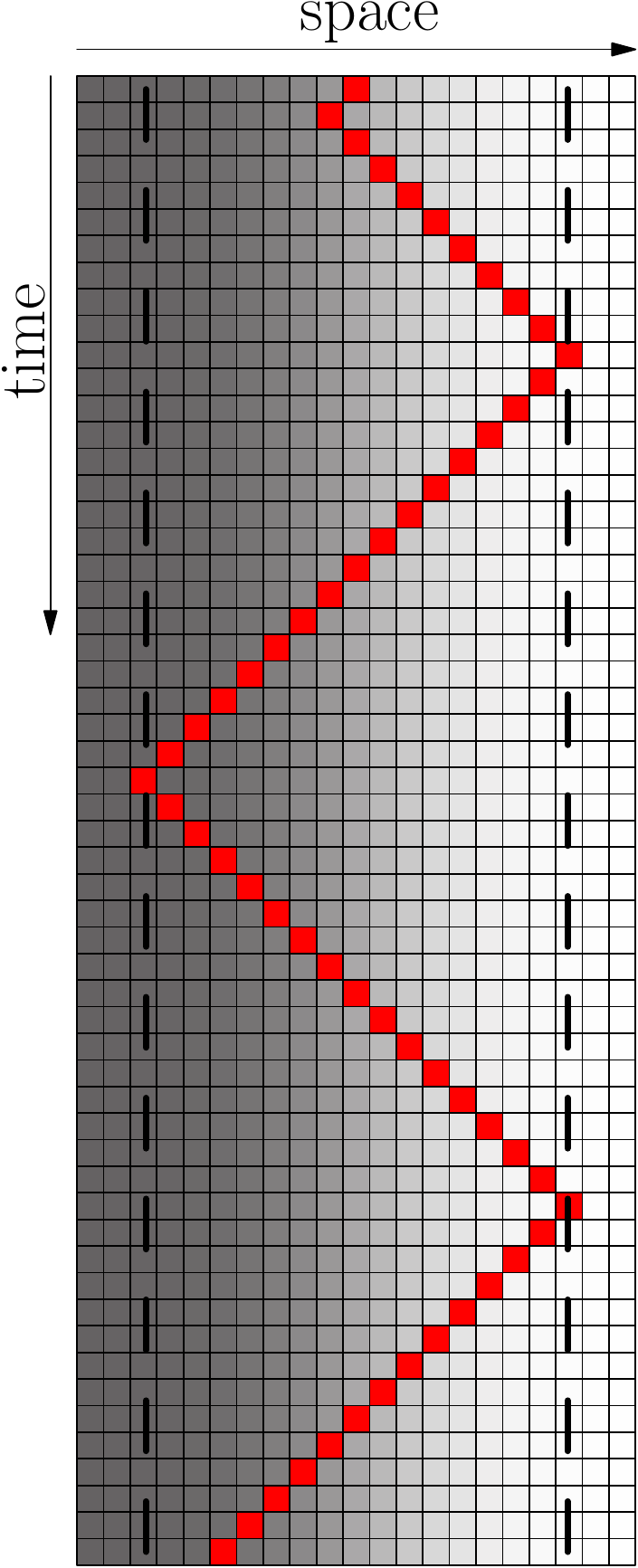}
\subcaption{Evolved ANN, environmental setting~1}\label{fig:evolvehandcoded:ehc_st1}
\end{minipage}
\hspace*{.01\linewidth}%
\begin{minipage}[b]{.18\linewidth}
\includegraphics[width=1.0\textwidth]{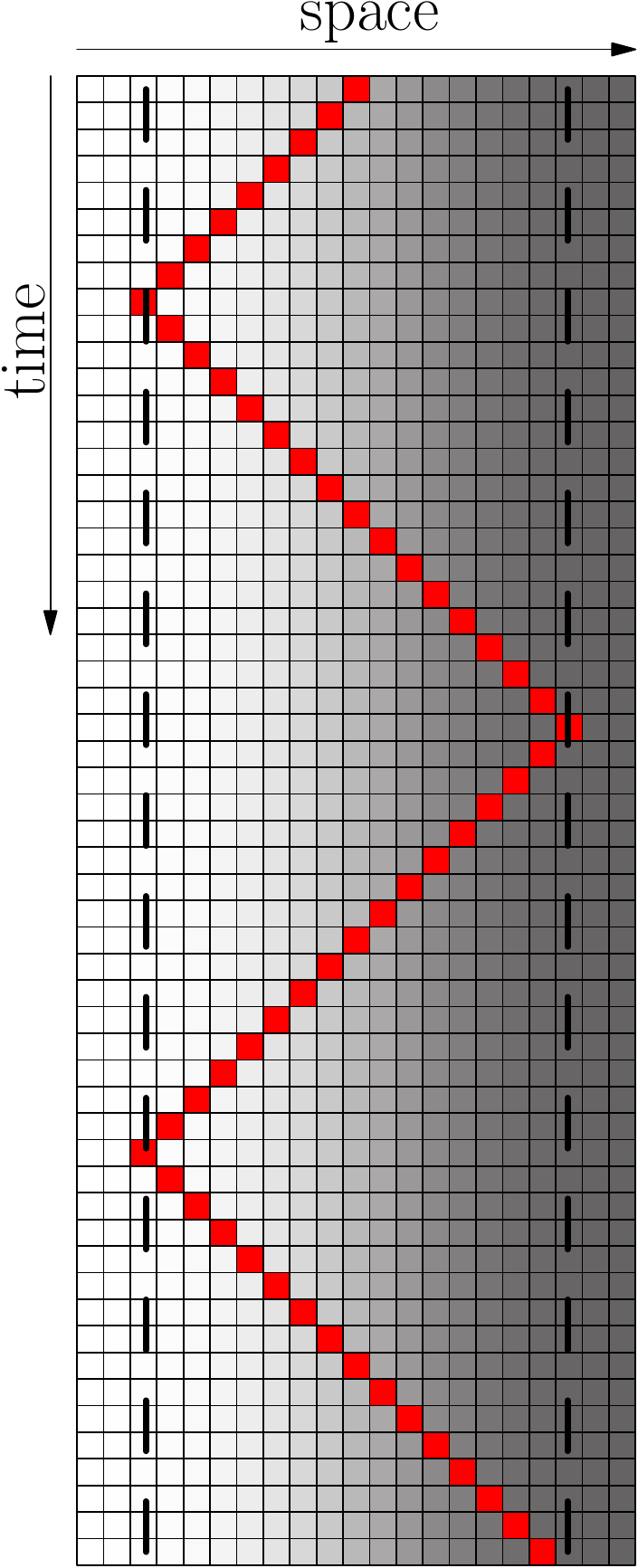}
\subcaption{Evolved ANN, environmental setting~2}\label{fig:evolvehandcoded:ehc_st2}
\end{minipage}
\caption{An example hand-coded ANN with sub-optimal fitness (\ref{fig:evolvehandcoded:base}) is evolved to reach an ANN with the maximum fitness (\ref{fig:evolvehandcoded:evolved}).}\label{fig:evolvehandcoded}
\end{figure}

\begin{figure}[h!]
\begin{minipage}[b]{.49\linewidth}
\includegraphics[width=0.9\textwidth]{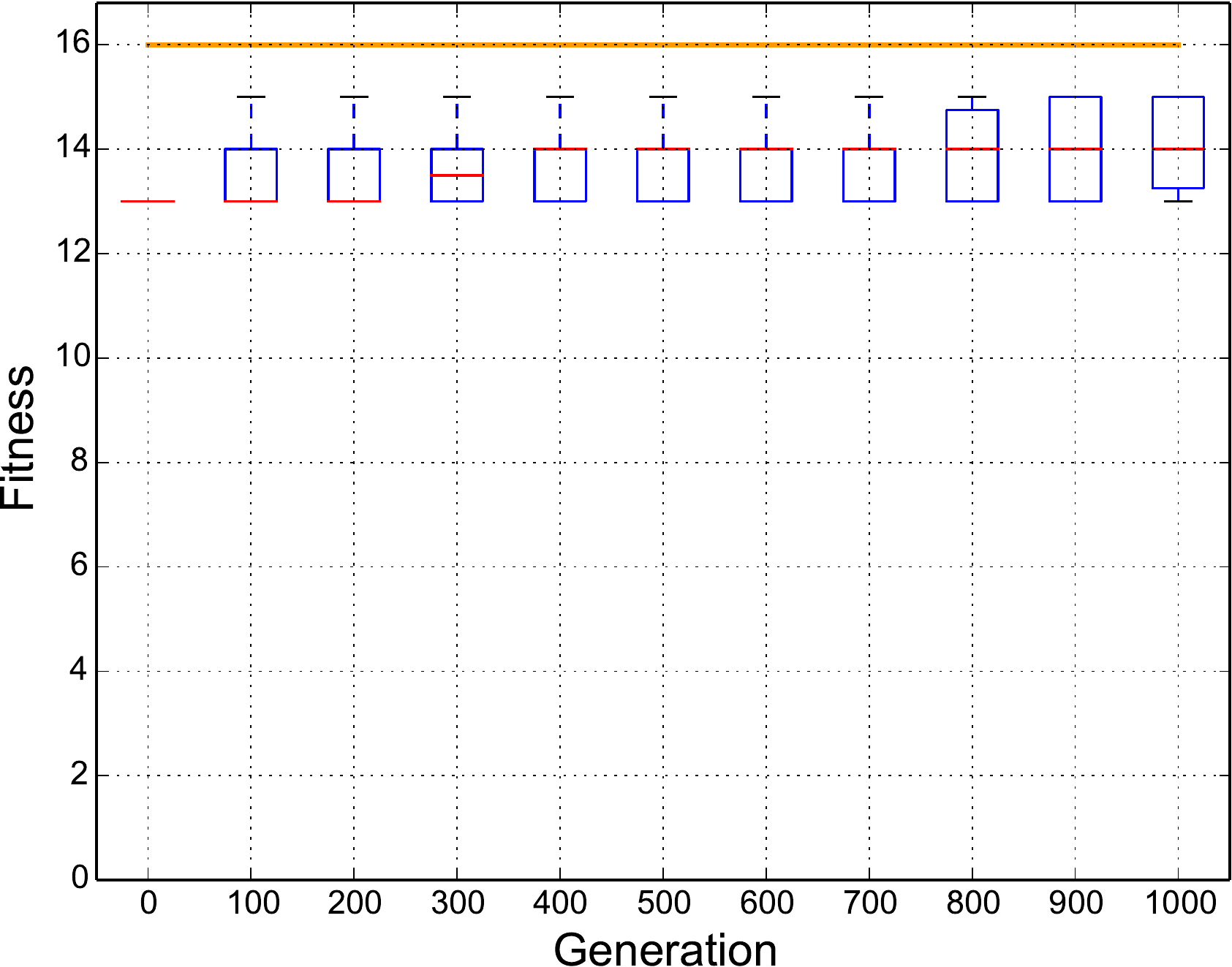}
\subcaption{fitness dynamics.}\label{fig:results:evolvingHC:swtr}
\end{minipage}%
\hspace*{.05\linewidth}%
\begin{minipage}[b]{.18\linewidth}
\includegraphics[width=1.0\textwidth]{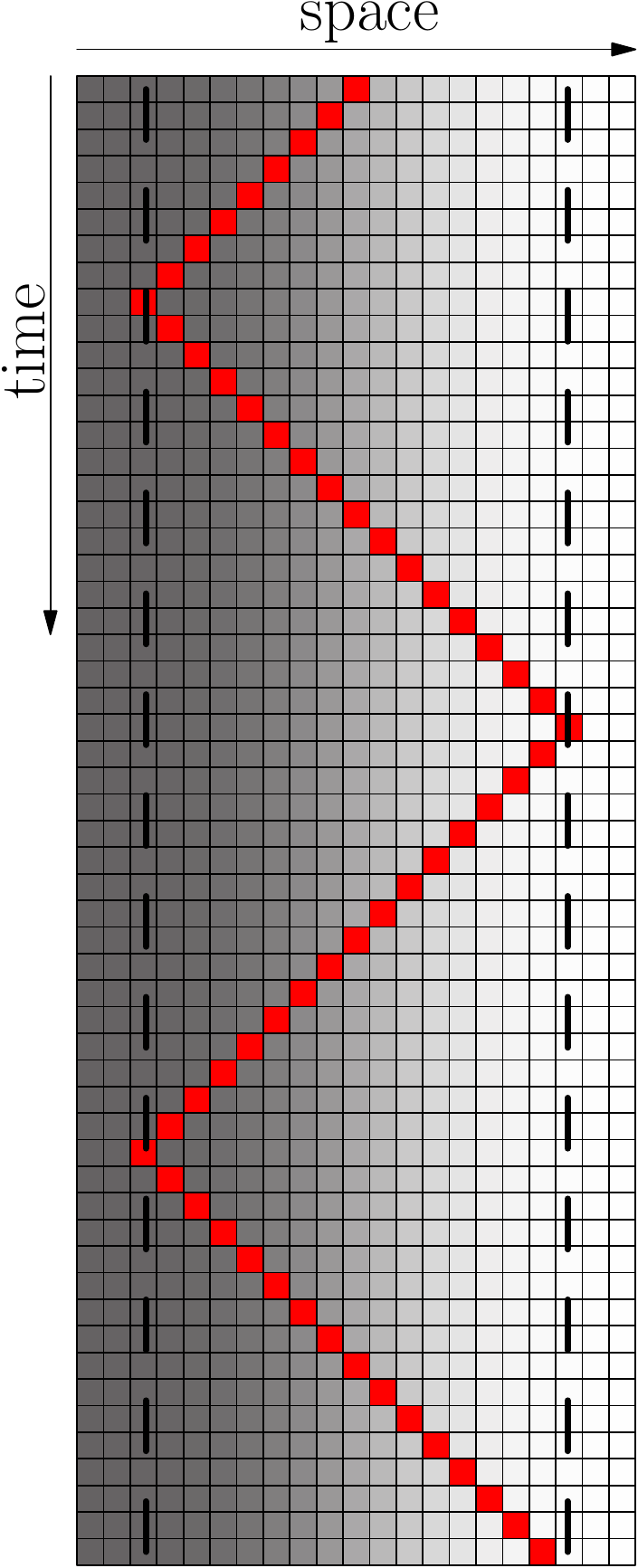}
\subcaption{setting~1}\label{fig:results:evolvingHC:swst1}
\end{minipage}
\hspace*{.01\linewidth}%
\begin{minipage}[b]{.18\linewidth}
\includegraphics[width=1.0\textwidth]{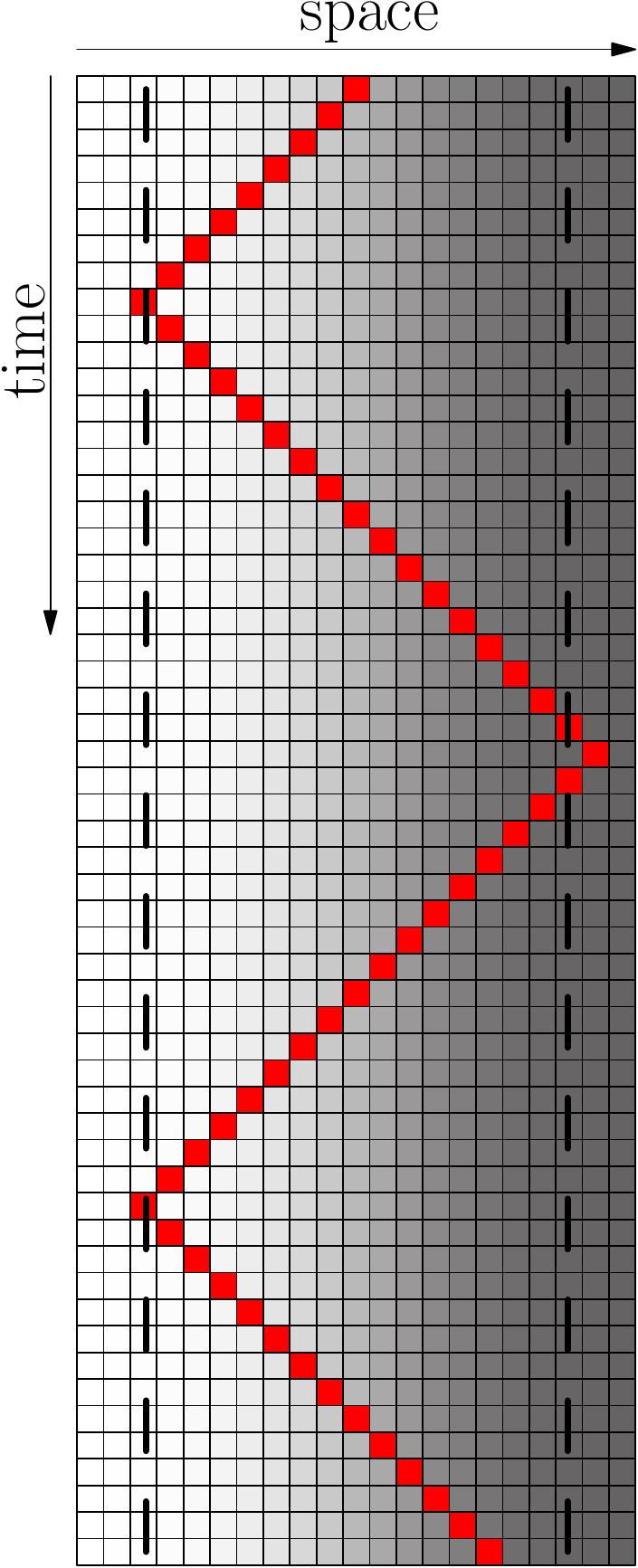}
\subcaption{setting~2}\label{fig:results:evolvingHC:swst2}
\end{minipage}
\caption{Evolutionary results of the \emph{Switch} fitness regime starting with the hand-coded ANN (Fig.~\ref{fig:evolvehandcoded:base}): $\text{fitness}
  = F(1,0,0)$, the minimum of both evaluations was used as the fitness value for each genome.
\ref{fig:results:evolvingHC:swtr}) Fitness dynamics of the evolutionary population. The horizontal orange line indicates the maximum fitness achievable by a reactive Wankelmut controller. \ref{fig:results:evolvingHC:swst1}, \ref{fig:results:evolvingHC:swst2}) trajectory of the best performing genome in both environmental settings.  
  }\label{fig:results:evolvingHCsw}
\end{figure}

\begin{figure}[h!]
\begin{minipage}[b]{.49\linewidth}
\includegraphics[width=0.9\textwidth]{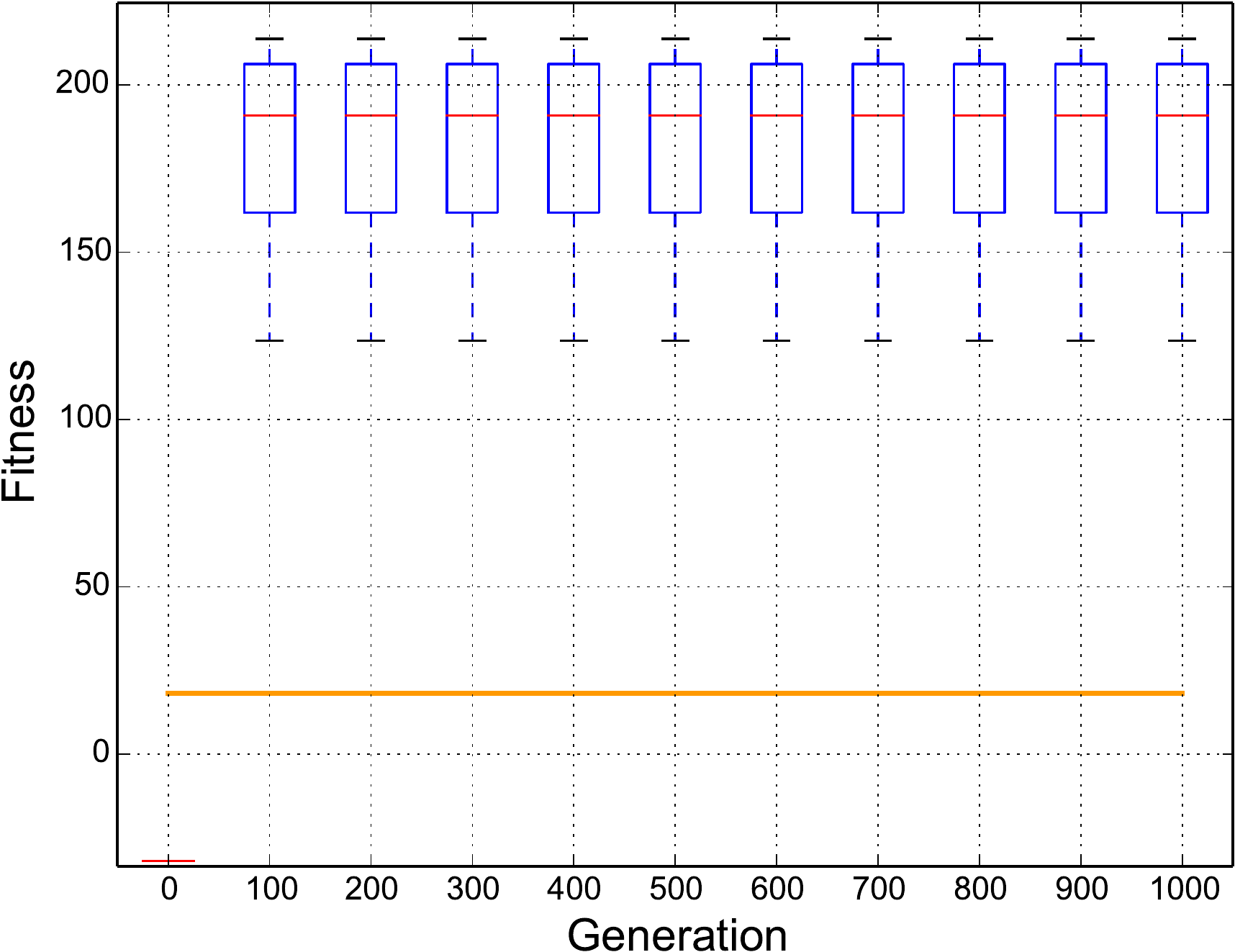}
\subcaption{fitness dynamics.}\label{fig:results:evolvingHC:cumtr}
\end{minipage}%
\hspace*{.05\linewidth}%
\begin{minipage}[b]{.18\linewidth}
\includegraphics[width=1.0\textwidth]{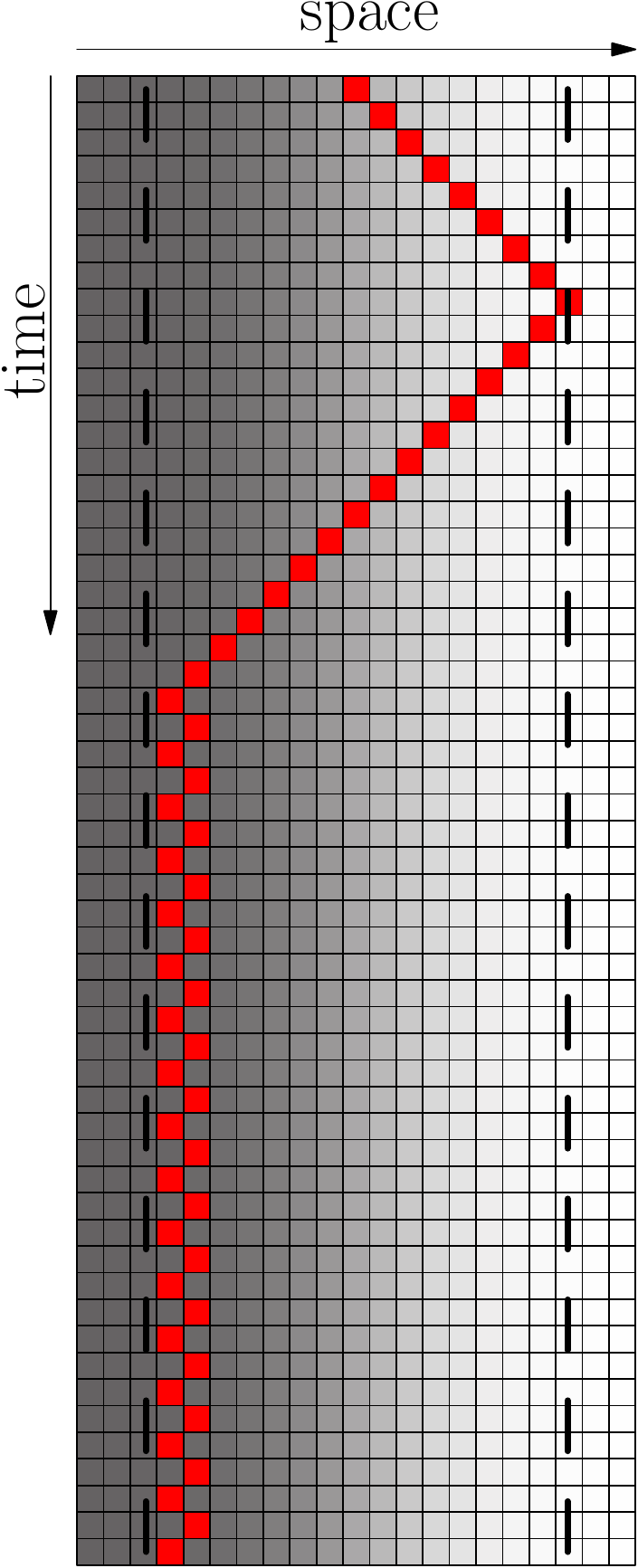}
\subcaption{setting~1}\label{fig:results:evolvingHC:cumst1}
\end{minipage}
\hspace*{.01\linewidth}%
\begin{minipage}[b]{.18\linewidth}
\includegraphics[width=1.0\textwidth]{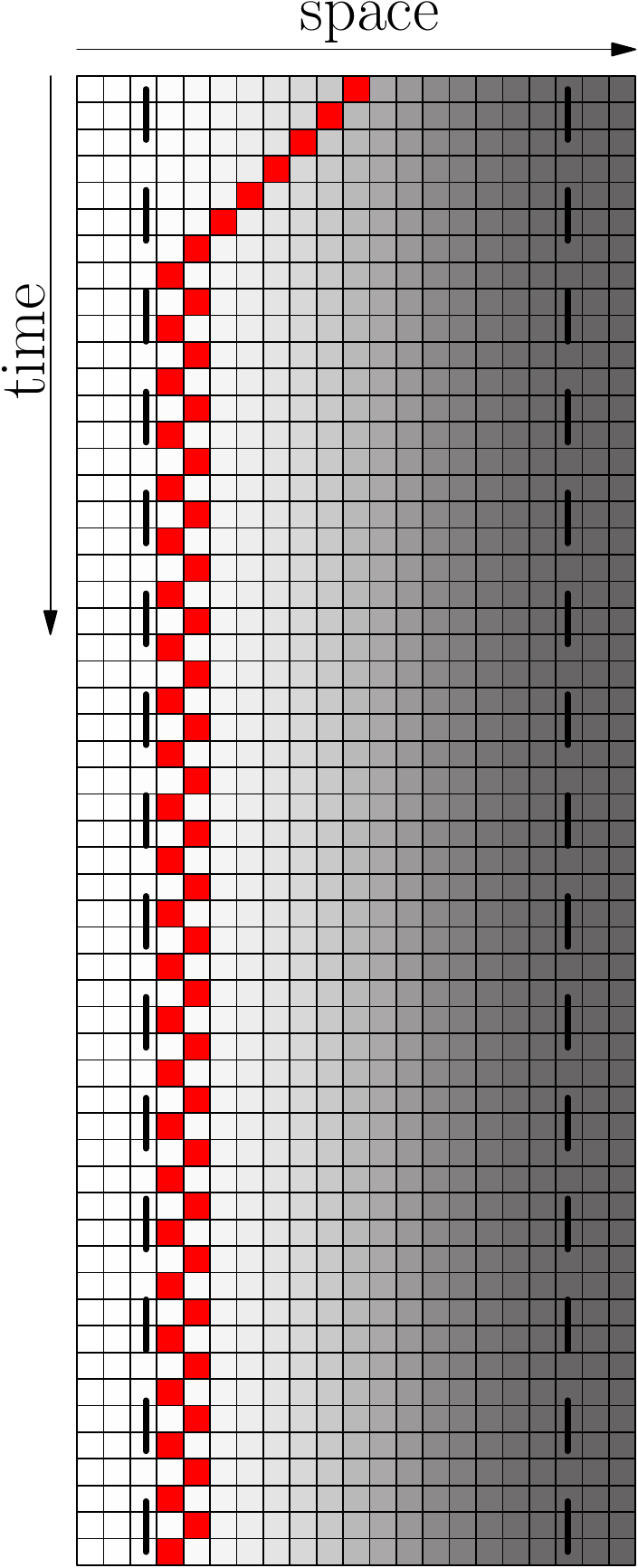}
\subcaption{setting~2}\label{fig:results:evolvingHC:cumst2}
\end{minipage}
\caption{Evolutionary results of the \emph{Cummulative} fitness regime starting with the hand-coded ANN (Fig.~\ref{fig:evolvehandcoded:base}): $\text{fitness}
  = F(0,0.01,0)$, the minimum of both evaluations was used as the fitness value for each genome.
\ref{fig:results:evolvingHC:cumtr}) Fitness dynamics of the evolutionary population. The horizontal orange line indicates the maximum fitness achievable by a reactive Wankelmut controller. \ref{fig:results:evolvingHC:cumst1}, \ref{fig:results:evolvingHC:cumst2}) trajectory of the best performing genome in both environmental settings.  
  }\label{fig:results:evolvingHCcum}
\end{figure}

\begin{figure}[h!]
\begin{minipage}[b]{.49\linewidth}
\includegraphics[width=0.9\textwidth]{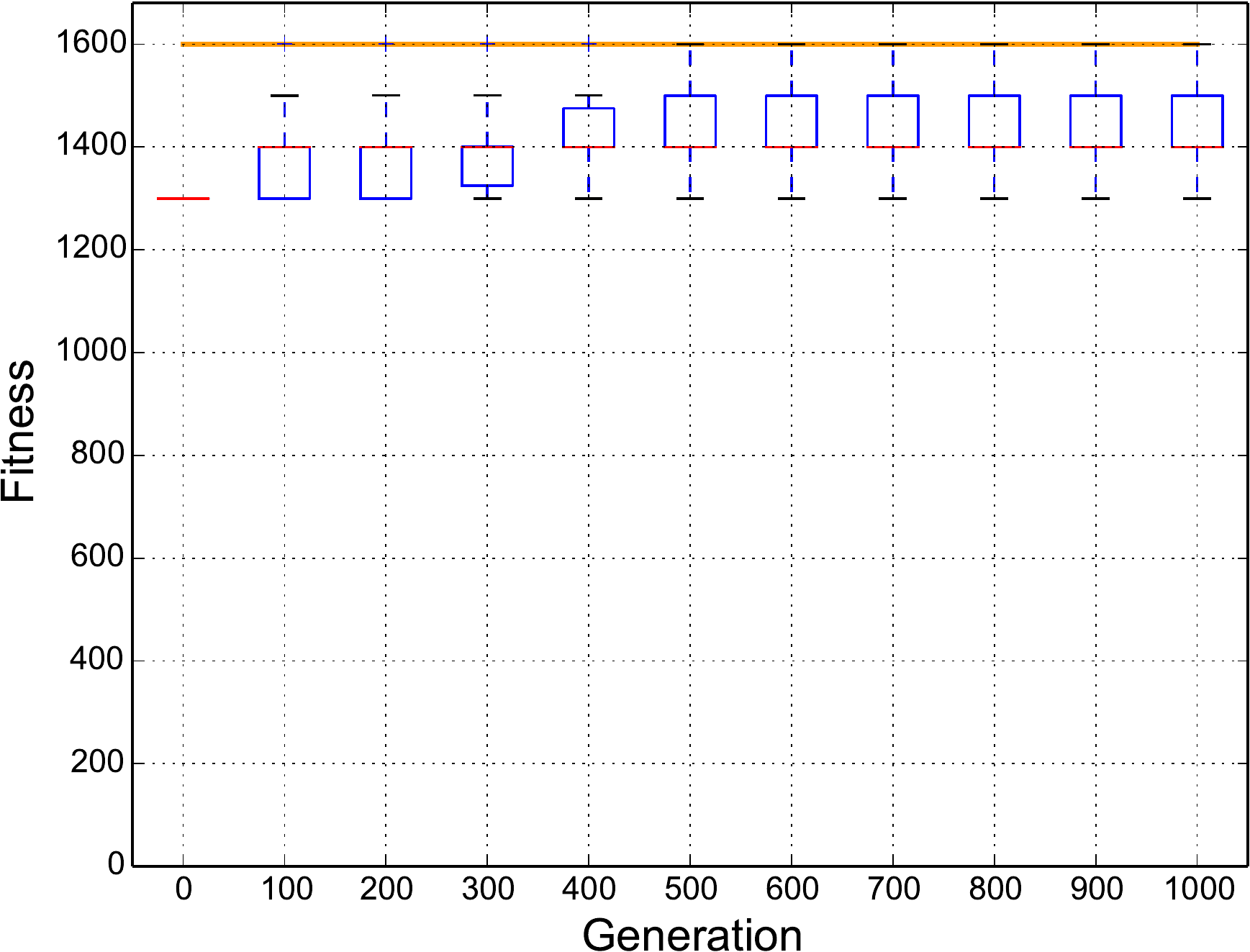}
\subcaption{fitness dynamics.}\label{fig:results:evolvingHC:swcumtr}
\end{minipage}%
\hspace*{.05\linewidth}%
\begin{minipage}[b]{.18\linewidth}
\includegraphics[width=1.0\textwidth]{acellular_side1_sw_cum_ANN}
\subcaption{setting~1}\label{fig:results:evolvingHC:swcumst1}
\end{minipage}
\hspace*{.01\linewidth}%
\begin{minipage}[b]{.18\linewidth}
\includegraphics[width=1.0\textwidth]{acellular_side2_sw_cum_ANN}
\subcaption{setting~2}\label{fig:results:evolvingHC:swcumst2}
\end{minipage}
\caption{Evolutionary results of the \emph{Cummulative+Switch} fitness regime starting with the hand-coded ANN (Fig.~\ref{fig:evolvehandcoded:base}): $\text{fitness}
  = F(100,0.01,0)$, the minimum of both evaluations was used as the fitness value for each genome.
\ref{fig:results:evolvingHC:swcumtr}) Fitness dynamics of the evolutionary population. The horizontal orange line indicates the maximum fitness achievable by a reactive Wankelmut controller. \ref{fig:results:evolvingHC:swcumst1}, \ref{fig:results:evolvingHC:swcumst2}) trajectory of the best performing genome in both environmental settings.  
  }\label{fig:results:evolvingHCswcum}
\end{figure}

\begin{figure}[h!]
\begin{minipage}[b]{.49\linewidth}
\includegraphics[width=0.9\textwidth]{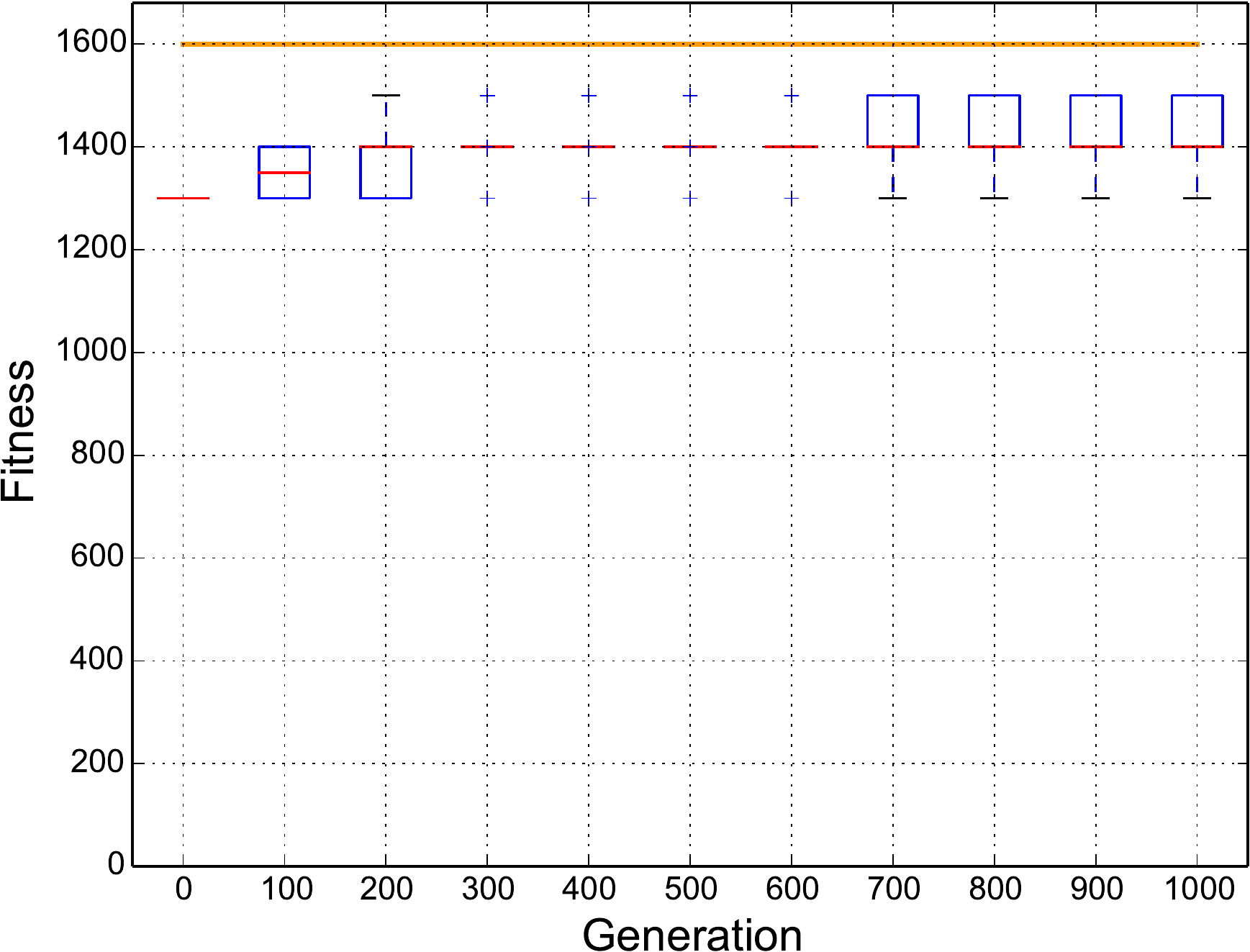}
\subcaption{fitness dynamics.}\label{fig:results:evolvingHC:swinstr}
\end{minipage}%
\hspace*{.05\linewidth}%
\begin{minipage}[b]{.18\linewidth}
\includegraphics[width=1.0\textwidth]{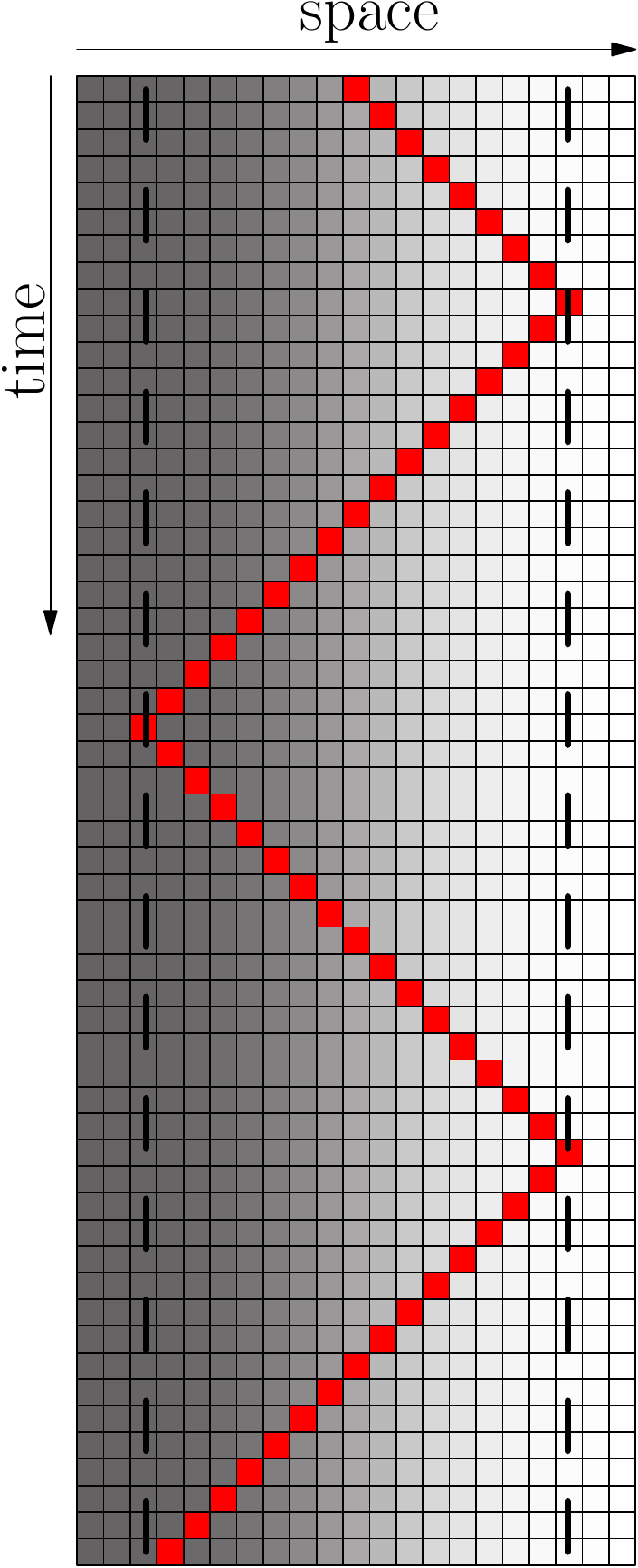}
\subcaption{setting~1}\label{fig:results:evolvingHC:swinsst1}
\end{minipage}
\hspace*{.01\linewidth}%
\begin{minipage}[b]{.18\linewidth}
\includegraphics[width=1.0\textwidth]{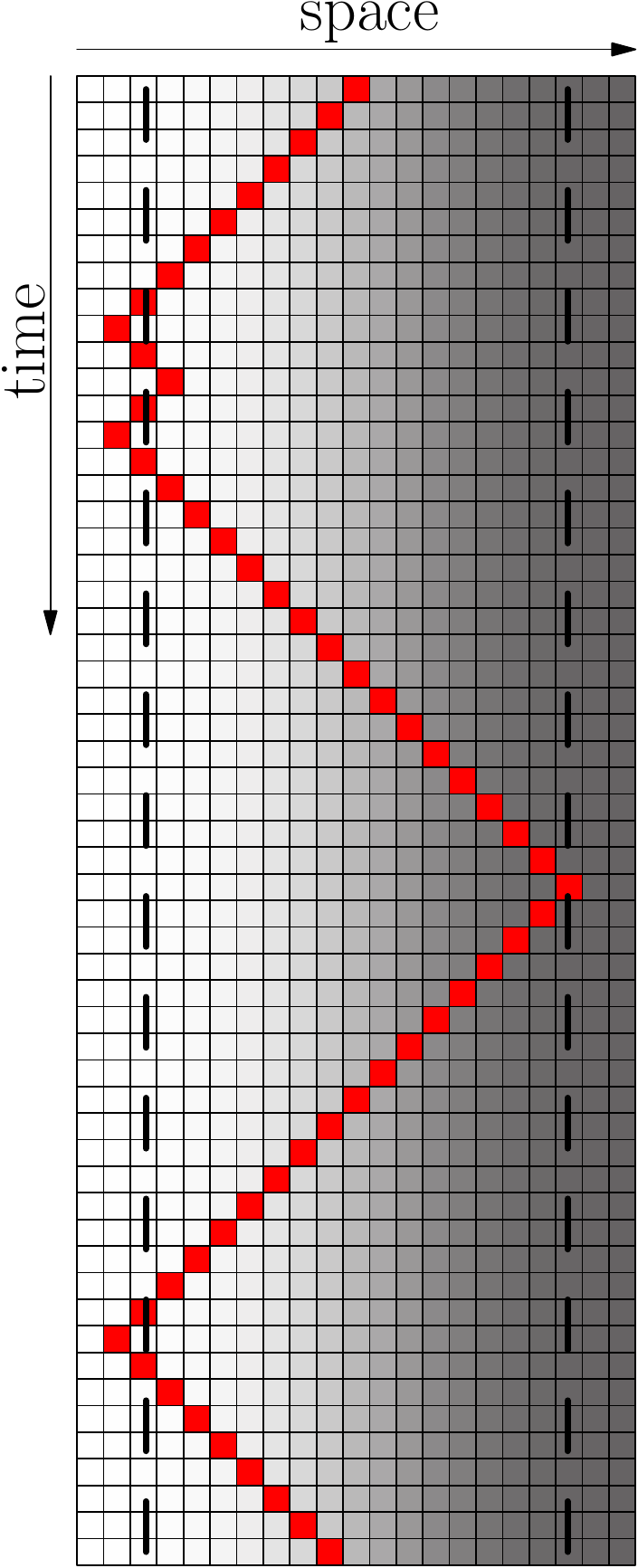}
\subcaption{setting~2}\label{fig:results:evolvingHC:swinsst2}
\end{minipage}
\caption{Evolutionary results of the \emph{Instant+Switch} fitness regime starting with the hand-coded ANN (Fig.~\ref{fig:evolvehandcoded:base}): $\text{fitness}
  = F(100,0,0.01)$, the minimum of both evaluations was used as the fitness value for each genome.
\ref{fig:results:evolvingHC:swinstr}) Fitness dynamics of the evolutionary population. The horizontal orange line indicates the maximum fitness achievable by a reactive Wankelmut controller. \ref{fig:results:evolvingHC:swinsst1}, \ref{fig:results:evolvingHC:swinsst2}) trajectory of the best performing genome in both environmental settings.  
  }\label{fig:results:evolvingHCswins}
\end{figure}

One of the hand-coded ANNs along with its behavior in the two environments is represented in Fig.~\ref{fig:evolvehandcoded:base}~-~\ref{fig:evolvehandcoded:hct_st2}. Although the agent is reactive and switches after passing the thresholds in each side, the behavior is not optimal: the agent had a delay in switching when it passes the thresholds at the environment represented in Fig.~\ref{fig:evolvehandcoded:hct_st2}. In an attempt to get a perfect behavior, we initialized an ANN described in section~\ref{sec:ann} with the weights of the hand-coded ANN and used the fitness functions to evolve the network. 
Fig.~\ref{fig:results:evolvingHCsw}~-~\ref{fig:results:evolvingHCswins} represent the results of the evolution for our different fitness regimes.
Fig.~\ref{fig:evolvehandcoded:evolved}~-~\ref{fig:evolvehandcoded:ehc_st2} represent the best evolved network and its behaviors in both environments. As it is shown in the figures, evolution has flattened the network by giving values to additional links between the nodes and consequently, the evolved ANN demonstrates an optimal behavior.

\section{Discussion and Conclusion}

We have proposed the Wankelmut task which is a very simple
task.
In the Wankelmut task, we evolve a controller that switches between two alternative conflicting tasks. Yet, there is no prioritizing between the two subtasks and therefore it is not of the type of a subsumption architecture. It is also not repeatedly alternating between two subtasks since the switching between the two tasks is based on the environmental clues.
The simplicity of the pseudo-code shown in Fig.~\ref{fig:pseudoCode}
clearly demonstrates the simplicity of the required controller that
implements an optimal, flexible, and reactive Wankelmut agent.

Our results indicate that plain-vanilla evolving ANNs easily can
evolve the reactive uphill walk (see
Fig.~\ref{fig:results:single-sw:c}). However switching to the
opposite behavior (downhill motion) and then flipping back was not
found by any of our approaches, regardless what fitness regime we used
and regardless what substrate we used for evolution (ANN or CTRNN). In
all tested fitness regimes the desired behavior did not evolve. In 
one of these fitness regimes (\emph{Cummulative} regime) evolution found a
``cheap trick'' to maximize the fitness with a surprising behavior,
however, the desired reactive Wankelmut behavior did not evolve there.

In order to prove that the task is solvable by the encoding that we have used here, we designed two hand-coded ANNs that solve the task. However, the hand-coded networks demonstrated a behavior that is quite good but not optimal. To further improve this, we then evolved one of the hand-coded ANNs and achieved the optimal behavior showing that the encoding covers the solution and the evolution can evolve the behavior when it is searching in the vicinity of the solution (Fig.\ref{fig:evolvehandcoded}).

We speculate that learning the downhill behavior ``destroys'' the already
learned network structure for performing the uphill walk behavior.
Thus, to evolve a combination of both would require a specific
evolutionary framework that is designed to generate modules, store
(freeze) useful modules and to combine them. However, such a
functionality seems not to evolve from itself in our system in an emergent way.

Given that the hand-coded solution is very simple, a tree-based approach
with exhaustive search or a genetic programming (GP)
approach~\cite{koza92} is expected to be able to find the desired behavior. 
However we would expect also here to hit the same "wall of complexity", as
controllers that require a bit more complexity overwhelm the
exhaustive tree-search, while the existing local optima, that already
fooled the ANN+Evo approach and the CTRNN+Evo approach, will also fool
the stochastic GP search in a similar way. This remains to be tested
in future experiments.

We point out that the target of evolving controllers for the Wankelmut
task from scratch requires to evolve a behavioral switch. However,
evolution of such a switch represents a hen-egg problem. The switch
is useless without the two motion-modules (subnets) for uphill and
downhill motion, while those sub-modules are useless without the
switch.

In addition, we want to stress the fact that a similar behavior could be
exerted by just switching the environment in an oscillatory way. This
is not the same as our envisioned Wankelmut behavior, as our behavior
is intrinsically switching its behavioral pattern in reaction to a
stable environment. So it is intrinsic dynamics exhibited in a global
stable environment and not a fixed (stable) behavioral pattern in a
globally dynamic environment.

The Wankelmut task might also prove to be an interesting benchmark for
more sophisticated methods that push towards modularization. Examples
are Artificial Epigenetic Networks as reported by
\cite{turner16}: They used the coupled inverted pendulums
benchmark~\cite{hamann11a} which requires the concurrent evolution of
several behaviors similar to the Wankelmut task. However, the control
of coupled inverted pendulums is more complex than the simple world of
Wankelmut. Other methods alternate between different tasks on
evolutionary time-scales~\cite{Kashtan2005} and there are methods
that, in addition, also impose costs on
connections between nodes in neural networks~\cite{Clune2012evolutionaryorigins}. 
Other interesting approaches either pre-determine or push towards
modularity~\cite{Nolfi96usingemergent}. The approach of HyperNEAT was
tested for its capability to generate modular networks with a negative
result in the sense that modularity was not automatically
generated~\cite{CluneBMO10}. \cite{Bongard2011gecco} reports
modularity by imposing different selection pressures to different
parts of the network.

In fact we propose here two challenges for the scientific
  community at once:
\begin{enumerate}
\item The first challenge is to find the most simple uninformed
  evolutionary computation algorithm that can solve the Wankelmut task
  presented here. Then the community can search for the next simple
  task that shows to be unsolvable for this new algorithm.  This will
  yield iterative progress in the field.
\item The Wankelmut task is the simplest task found so far
  (concerning required memory size, number of modules, dimensionality
  of the world it operates in, etc.). Still, there might be even
  simpler tasks that already break plain-vanilla uninformed
  evolutionary computation, so we also pose the challenge to search
  for such simpler benchmark tasks.
\end{enumerate}
It should be noted that there is no evidence that such walls of
complexity are consistent in a way that breaking one wall may cause a new wall at a different position. This is
quite similar to the reasoning behind the ``no free lunch''
theorem. Theoretically, an optimization algorithm cannot be optimal
for all tasks. However, neither natural nor artificial evolution achieve in general
optimal solutions. Instead, especially natural, evolution has proven to be a
great heuristic for which the walls might be fixed in a certain place
but definitely far away from the wall for state-of-the-art artificial evolution. Hence,
we should try to search for that one heuristic that allows us to push
all walls of many different tasks as far as possible forward (while
still accepting the implications of no free lunch).

We think that either dismissing all pre-informed methods or finding minimally pre-informed methods to solve the Wankelmut task is important. Natural evolution has produced billions
of billions of reactive and adaptive behaviors of organisms much more complex than the Wankelmut task
and it has achieved this without any information that promoted
self-complexification and self-modularization. In contrast, the evolutionary process started
from scratch and developed all of that due to evolution-intrinsic forces. We think that this can be
a lesson for evolutionary computation: studying how
an uninformed process that is neither pre-fabricated towards complexification or modularization
and which is not specifically rewarded for complexification or modularization
can still yield complex solutions. Nature has shown it and evolutionary computation and biologists 
together should find out how this was achieved. Maybe this would then be not ``evolutionary computation" anymore
but rather ``artificial evolution" a real valid, yet still simple model of natural evolution.

\section*{Author Contributions}

T.S., P.Z., H.H. contributed to the writing of the paper in equal
parts. T.S. had the initial idea of the problem study, defined the
main problem statement, programmed the initial cellular simulator used
in this study, and produced the schematic drawing
(Fig.~\ref{fig:task}). P.Z. programmed the CTRNN code, hand-coded ANNs, parts of the
analysis scripts, and scripts for the box plots. H.H. programmed the
ANN code and the evolutionary code, parts of the analysis scripts, and
the `asymptote' script used to produce the trajectory figures. The
numerical analysis and interpretation of results was a joint effort of
all authors.

\section*{Funding}
T.S. was supported by the EU FP7-FET PROACTIVE grant \#601074
(ASSISI$_{bf}$). P.Z. and H.H. were supported by the EU H2020-FET
PROACTIVE grant \#640959 ({\it flora robotica}).

\section*{Acknowledgments}
We thank Dr. J{\"u}rgen Stradner for his early investigations of the
problem (data not used here) and Dr. Ronald Thenius for his input of
Artificial Neural Networks.

\bibliographystyle{unsrt}
\bibliography{./bibliography}
\newpage
\section*{Figures}
\setcounter{subfigure}{0}

\setcounter{subfigure}{0}

\setcounter{subfigure}{0}

\setcounter{subfigure}{0}

\setcounter{subfigure}{0}

\setcounter{subfigure}{0}
\setcounter{subfigure}{0}
\setcounter{subfigure}{0}

\setcounter{subfigure}{0}

\setcounter{subfigure}{0}

\setcounter{subfigure}{0}

\setcounter{subfigure}{0}

\end{document}